%% file: nuq_arxiv.tex
\documentclass[10pt,twocolumn,letterpaper]{article}
\usepackage{iccv}
\usepackage{times}
\usepackage{epsfig}
\usepackage{graphicx}
\usepackage{amsmath}
\usepackage{amssymb}

\input{math_commands}

\usepackage{url}
\usepackage{amsthm}
\usepackage[utf8]{inputenc} % allow utf-8 input
\usepackage[T1]{fontenc}    % use 8-bit T1 fonts
\usepackage{booktabs}       % professional-quality tables
\usepackage{amsfonts}       % blackboard math symbols
\usepackage{nicefrac}       % compact symbols for 1/2, etc.
\usepackage{microtype}      % microtypography
\usepackage{bm}
\usepackage{color}
\usepackage{extarrows}
\usepackage{subfigure}
\usepackage{multirow}
\usepackage{mathtools}
\input{definitions}

\usepackage[pagebackref=true,breaklinks=true,colorlinks,bookmarks=false]{hyperref}

\iccvfinalcopy % *** Uncomment this line for the final submission

\def\iccvPaperID{10802} % *** Enter the ICCV Paper ID here

\ificcvfinal\fi

\begin{document}

%%%%%%%%% TITLE
%\title{Modeling Neural Uncertainty for Improved Stochastic Video Prediction}
\title{A Hierarchical Variational Neural Uncertainty Model \\for Stochastic Video Prediction}
\author{Moitreya Chatterjee$^{1*}$ \qquad Narendra Ahuja$^{1}$ \qquad Anoop Cherian$^{2}\thanks{Equal contribution.}$\\
$^1$University of Illinois at Urbana-Champaign
Champaign, IL 61820, USA\\
$^2$Mitsubishi Electric Research Laboratories,
Cambridge, MA 02139\\
\texttt{metro.smiles@gmail.com\quad n-ahuja@illinois.edu\quad cherian@merl.com}
}

\maketitle

%%%%%%%%% ABSTRACT
%\begin{abstract}
%   The ABSTRACT is to be in fully-justified italicized text, at the top
%   of the left-hand column, below the author and affiliation
%   information. Use the word ``Abstract'' as the title, in 12-point
%   Times, boldface type, centered relative to the column, initially
%   capitalized. The abstract is to be in 10-point, single-spaced type.
%   Leave two blank lines after the Abstract, then begin the main text.
%   Look at previous CVPR abstracts to get a feel for style and length.
%\end{abstract}

\begin{abstract}
    Predicting the future frames of a video is a challenging task, in part due to the underlying stochastic real-world phenomena. Prior approaches to solve this task typically estimate a latent prior characterizing this stochasticity, however do not account for the predictive uncertainty of the (deep learning) model. Such approaches often derive the training signal from the mean-squared error (MSE) between the generated frame and the ground truth, which can lead to sub-optimal training, especially when the predictive uncertainty is high. Towards this end, we introduce \emph{Neural Uncertainty Quantifier} (NUQ) - a stochastic quantification of the model's predictive uncertainty, and use it to weigh the MSE loss. We propose a hierarchical, variational framework to derive NUQ in a principled manner using a deep, Bayesian graphical model. Our experiments on four benchmark stochastic video prediction datasets show that our proposed framework trains more effectively compared to the state-of-the-art models (especially when the training sets are small), while demonstrating better video generation quality and diversity against several evaluation metrics.
\end{abstract}

\section{Introduction}
\label{sec:intro}

Extrapolating the present into the future is a task essential to predictive reasoning and planning. When artificial intelligence systems are deployed to work side-by-side with humans, it is critical that they reason about their visual context and generate plausible futures so that they can anticipate the potential needs of humans or catastrophic risks and be better equipped. Such a visual future generation framework could also benefit applications such as video surveillance~\cite{wang2018background}, human action recognition and forecasting~\cite{sun2019relational,vuppalapati2020human} as well as simulation of real-world scenarios to train robot learning algorithms, including autonomous driving~\cite{jain2015car}. However, such applications have a high element of stochasticity, which makes this prediction task challenging.

\begin{figure}[t]
    \centering
    \includegraphics[width=6.9cm,trim={0cm 0.5cm 0cm 0cm},clip]{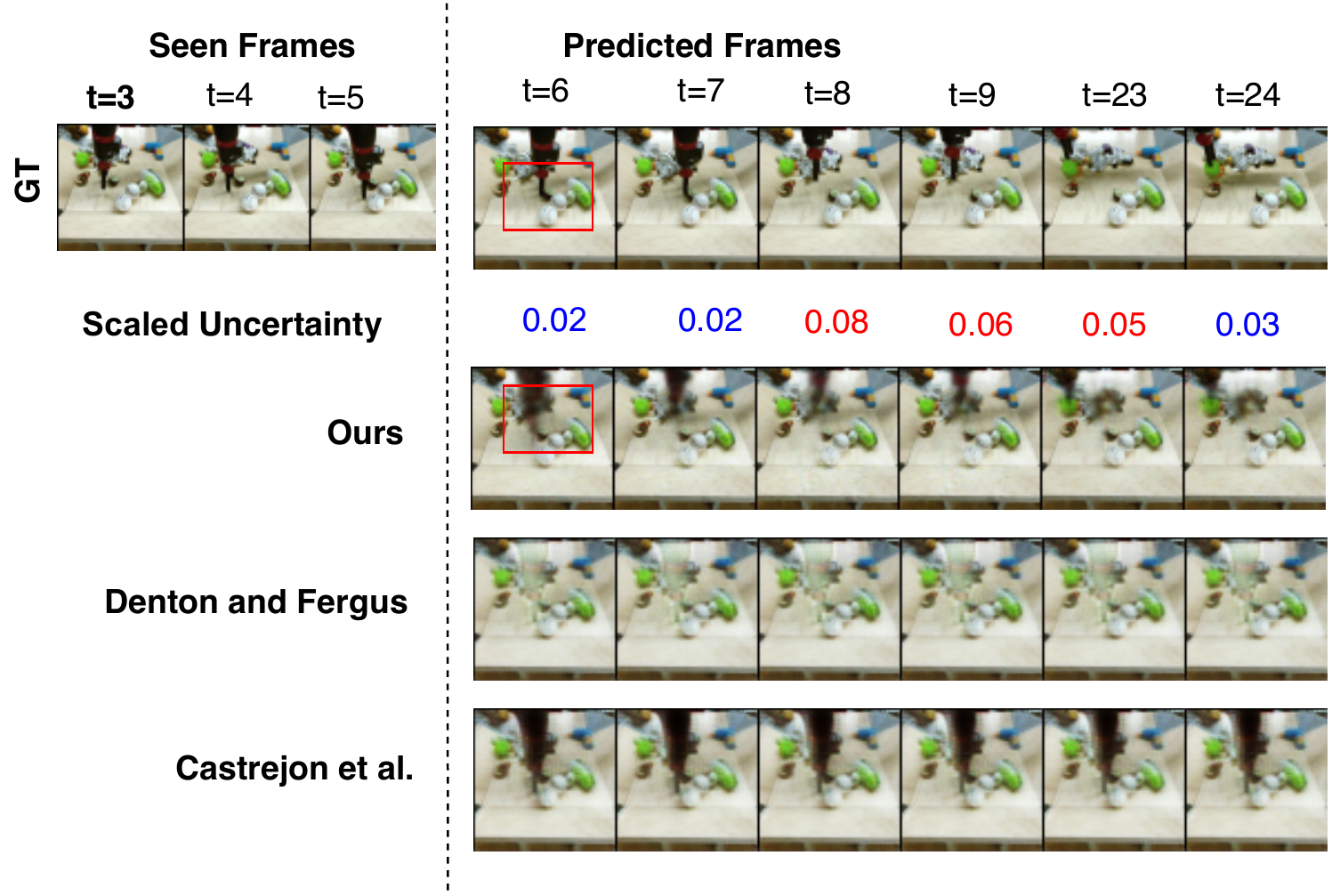}
   \caption{Qualitative results vis-\'a-vis state-of-the-art video prediction baselines using the proposed NUQ framework on the BAIR Push dataset~\cite{ebert2017self}, trained using only 2,000 samples (rather than the full 40K samples). Regions with high motion are shown by a red box. Also shown is an estimate of the per-frame scaled uncertainty estimated by our model. Note that the robotic arm changes direction at $t=8$, which is reflected in the predicted uncertainty. }
    \label{fig:bair_sota}
    \vspace*{-0.6cm}
\end{figure}

The resurgence of deep neural networks, especially the advent of generative adversarial networks~\cite{goodfellow2014generative}, has enabled significant progress in the development of frameworks for generating visual data, such as images~\cite{karras2019style}. While, temporally-evolving extensions of such image generation techniques have shown benefits in artificially producing video sequences for deterministic visual contexts~\cite{vondrick2016generating,walker2017pose,fragkiadaki2015recurrent,kwon2019predicting,jin2020exploring}, they usually fail to model real-world sequences that are often highly stochastic. 

Several recent works in video generation, thus design modules to factor in data stochasticity while making predictions ~\cite{liang2017dual,babaeizadeh2017stochastic,denton2018stochastic,castrejon2019improved}. Specifically, such methods assume a latent stochastic prior, from which random samples are drawn, in order to generate future frames. In Babaeizadeh \etal~\cite{babaeizadeh2017stochastic}, this stochastic prior is assumed to follow a fixed normal distribution, which is sampled at every time step, while Denton and Fergus~\cite{denton2018stochastic}, learn this prior from data. The latter's key insight is to use a variational posterior to guide the learning of the prior to produce the sufficient statistics of the normal distribution governing the prior. Such stochastic methods typically employ a deterministic decoder (a neural network) that combines an embedding of the visual context and a random sample from the stochastic prior to generate a future video frame. The variance in this prior accounts for the stochasticity underlying the data. To train such models, the mean-squared error (MSE) is then minimized by comparing the predictions against the true video frames. 

Nonetheless existing stochastic methods have largely ignored the predictive uncertainty (aleatoric uncertainty)~\cite{kendall2017uncertainties} of the models, which might adversarially impact downstream tasks that leverage these predictions. From a machine learning stand point, ignoring the predictive uncertainty might lead to the model being unnecessarily penalized (via the MSE), even if it makes a very uncertain prediction that ends up being different from the ground-truth. This can destabilize the training of the underlying neural networks, leading to slower convergence or requiring larger training data. This is of importance because such data might be expensive or sometimes even difficult to collect (e.g., predicting the next human actions in instruction videos, or a rare traffic incident), and thus effective training with limited data is essential. 

In this work, we rise up to these challenges by quantifying the predictive uncertainty of a stochastic frame prediction model and using it to calibrate its training objective. In particular a stochastic estimate of the predictive uncertainty, derived from the latent space of the model, is used to weigh the MSE. That is, when the uncertainty is high, the MSE is down-weighted proportionately, and vice versa; thereby regularizing the backpropagation gradients to train the frame generation module. Moreover, this uncertainty estimate can be used for downstream tasks, such as for example, regulating the manuevers in autonomous driving~\cite{jain2015car,vuppalapati2020human}. We call our scheme, \emph{Neural Uncertainty Quantifier} (NUQ).

We observe that the weight on the MSE that NUQ introduces, basically amounts to the variance of the normal distribution governing the generated output. Thus, an obvious consideration would be to estimate the variance directly from the output. However, this may be cumbersome due to the very high dimensionality of the output space (order of the number of pixels).  We instead, choose to derive it from the variance of the latent space prior, which has far fewer dimensions. Specifically, NUQ leverages a variational, deep, hierarchical, graphical model to bridge the variance of the latent space prior and that of the output. Our framework is trained end-to-end. Sample generations by our framework is shown in Figure~\ref{fig:bair_sota}. In addition, inspired by the recent successes of generative adversarial networks~\cite{goodfellow2014generative,kwon2019predicting,liang2017dual}, we propose a variant of our framework that uses a novel sequence discriminator, in an adversarial setting. This discriminator module helps to constrain the space of possible output frames, while enforcing motion regularities in the generated videos.

To empirically verify our intuitions, we present experiments on a synthetic (Stochastic Moving MNIST~\cite{denton2018stochastic}) and three challenging real world datasets: KTH-Action ~\cite{schuldt2004recognizing}, BAIR push ~\cite{ebert2017self}, and UCF-101~\cite{ucf2012doc} for the task of future frame generation. Our results show that our framework converges faster than prior stochastic video generation methods, and leads to state-of-the-art video generation quality, even when the dataset size is small, while exhibiting generative diversity in the predicted frames. 

Below, we summarize the main contributions of this paper:
\begin{enumerate}
    \item We present \emph{Neural Uncertainty Quantifier} (NUQ), a deep, Bayesian network that learns to estimate the predictive uncertainty of stochastic frame generation models, which can be leveraged to control the training updates, for faster and improved convergence of predictive models.
    \item We propose a novel, hierarchical, variational training scheme that allows for incorporating problem-specific knowledge into the predictions via hyperpriors on the uncertainty estimate.
    \item Experimental results demonstrate our framework's better video generation and faster training capabilities, even with small training sets compared to recent state-of-the-art methods on stochastic video generation tasks, across multiple datasets.  
\end{enumerate}

\section{Related Work}
\label{sec:related_work}
 Early works in video frame prediction mostly resorted to end-to-end deterministic architectures~\cite{ranzato2014video,srivastava2015unsupervised,finn2016unsupervised}. Ranzato \etal~\cite{ranzato2014video} proposed to divide frames into patches and extrapolate their evolution in time. Srivastava \etal~\cite{srivastava2015unsupervised} use image encoders with pre-trained weights to encode the frames. ContextVP~\cite{byeon2018contextvp} and PredNet~\cite{lotter2016deep} leverage Convolutional LSTMs~\cite{xingjian2015convolutional} for video prediction. Fragkiadaki \etal~\cite{fragkiadaki2015recurrent} proposes pose extrapolation using LSTMs. More recent approaches~\cite{kwon2019predicting,yu2019efficient} seek to predict frames bidirectionally (future and past), during training. However, the inherent deterministic nature of such models~\cite{jin2020exploring} often becomes a bottleneck to their performance. Instead, we seek to investigate approaches that allow modeling of the underlying stochasticity in the data while generating an assessment of the model's predictive uncertainty. 

Stochastic approaches constitute a recently emerging and one of the most promising classes of video prediction methods~\cite{liang2017dual,babaeizadeh2017stochastic,denton2018stochastic}. These approaches model the data stochasticity using a latent prior distribution and are thus readily generalizable to real-world scenarios. Popular among them are STORNs ~\cite{bayer2014learning}, VRNNs~\cite{chung2015recurrent}, SRNNs~\cite{fraccaro2016sequential}, and DMMs~\cite{krishnan2017structured}.  SV2P~\cite{babaeizadeh2017stochastic} is a more recent method that uses a single set of stochastic latent variables that are assumed to follow a fixed prior distribution. Denton and Fergus~\cite{denton2018stochastic} improve upon SV2P~\cite{babaeizadeh2017stochastic} by allowing the prior distribution to be adapted at every time step by casting the prior as a trainable neural network. Their method is shown to achieve superior empirical performance, thus underlining the importance of learning to model data stochasticity. We also note that generative models have recently been adapted to incorporate stochastic information through a hierarchical latent space~\cite{tomczak2018vae,vahdat2020nvae}. Such networks have also been applied to frame prediction tasks~\cite{castrejon2019improved}. None of these approaches however, explore the effectiveness of modeling the predictive uncertainty. While, technically it might be possible that the stochastic modules in these prior approaches can learn to quantify this uncertainty implicitly, it may need longer training periods or larger datasets. Instead we show that explicitly incorporating the predictive uncertainty into the learning objective, via a hierarchical, variational framework improves training and inference.

 Another line of work in frame prediction seeks to decouple the video into static and moving components~\cite{villegas2017decomposing,denton2017unsupervised,liu2018future,ho2019sme,wu2020future,guen2020disentangling}. Some of these approaches are deterministic, others stochastic. Denton \etal~\cite{denton2017unsupervised} extracts content and pose information for this purpose. Villegas \etal~\cite{villegas2017decomposing} adopt a multiscale approach towards frame prediction which works by building a model of object motion, however they require supervisory information, such as annotated pose, during training. Ye \etal~\cite{ye2019compositional} propose a compositional approach to video prediction by stitching the motion of individual entities. While promising, their approach relies on auxiliary information such as spatial locations of the entities, and as a result, is difficult to generalize. Jin \etal~\cite{jin2020exploring} investigates decoupling in the frequency space, however they do not model the data stochasticity explicitly. Hsieh \etal~\cite{hsieh2018learning} describes a similar approach by modeling the motion and appearance of each object in the video, but without requiring any auxiliary information. Different from these set of approaches, our proposed framework models frames holistically and is thus agnostic to the video content. 

Modeling the predictive uncertainty in deep networks has garnered significant attention lately~\cite{blundell2015weight,clements2019estimating,kumar2020luvli,malinin2018predictive,tagasovska2019single}. Some of these works~\cite{ardizzone2020training,kumar2020luvli,malinin2018predictive} investigate it in a classification setting, while some others~\cite{blundell2015weight,harakeh2020bayesod} in the context of regression. Uncertainty has also recently been explored in the context of generative models~\cite{lee2020estimation,mcallister2019robustness,welling2011bayesian}. However, predictive uncertainty modeling in the context of frame prediction has remained largely unexplored. NUQ attempts to fill this gap.

\section{Background}
\label{sec:background}
Suppose $\vx_{1:T} := \seq{\vx_1, \vx_2, \cdots, \vx_{T}}$ denotes a sequence of random variables, each $\vx_t$ representing a video frame at time step $t$. Assuming we have access to a few initial frames $x_{1:F}$, to set the visual context (where $1\leq F<T$), our goal is to generate the rest of the frames $\vx_{F+1}$  onwards autoregressively, i.e., conditioned on the seen frames and what has been generated hitherto. This task amounts to finding a prediction model $p_{\theta}(\cdot)$, parameterized by $\theta$, that minimizes the expected negative log-likelihood.

When unknown factors of variation are involved in the data generation process, a determinstic predictive model is insufficient. A standard way to incorporate stochasticity is by assuming the generated frames are in turn conditioned on a latent prior model $\p(\vz_t)$; i.e., $\vz_t\sim \p(\vz_t), \vx_t \sim p_{\theta}(\vx_{t}| \vx_{1:t-1},\vz_t)$. Specifically, the stochasticity in the generative process is characterized by the variance in $\p(\vz_t)$, that produces diversity in $\vz_t\sim\p(\vz_t)$. Diversity among predicted frames emerges as a result of this variance. 

A well-known problem with the use of such latent stochastic priors is the intractability that it brings into the estimation of the \emph{evidence} or the log-partition function: $\p(\vx_t|\vx_{1:t-1})=\int_{\vz_t}\p_{\theta}(\vx_t|\vx_{1:t-1},\vz_t) \p(\vz_t) d\vz_t$. This problem is typically avoided by casting this estimation in an encoder-decoder setup, where the encoder embeds $\vx_{1:t}$ as $\vz_t \sim \p(\vz_t|\vx_{1:t})$, while the decoder outputs $\vx_t \sim \p_{\theta}(\vx_t|\vx_{1:t-1},\vz_t)$. In order to train efficiently, access to a variational posterior $\q_{\phi}(\vz_t|\vx_{1:t})$ -- that approximates the true posterior $\p(\vz_t|\vx_{1:t})$ of the encoder -- is assumed. Using this approximate posterior, learning the model parameters $\theta$ and $\phi$ amounts to maximizing the variational lower bound, $\loss_{\theta,\phi}$ ~\cite{kingma2013auto}:
\begin{equation}
\small
\begin{split}
     \log \p(\vx_t|\vx_{1:t-1}) \geq \loss_{\theta,\phi}, \text{ where } \\ \loss_{\theta,\phi} := \int_{\vz_{t}} \q_{\phi}(\vz_{t}|\vx_{1:t}) \log \prob{\theta}(\vx_t | \vx_{1:t-1}, \vz_{t}) d\vz_t \\  -  \int_{\vz_{t}} \q_{\phi}(\vz_{t}|\vx_{1:t}) \log\frac{\q_{\phi}(\vz_{t}|\vx_{1:t})}{ \p(\vz_t)} d\vz_t 
\end{split}
\label{eq:pre_elbo}
\normalsize
\end{equation}
From the definition, this amounts to:
\begin{equation}
\small
\begin{multlined}
    \loss_{\theta,\phi} = \expect{\q_{\phi}(\vz_{t}|\vx_{1:t})} \log \prob{\theta}(\vx_t | \vx_{1:t-1},\vz_{t}) - \\ \kl\left(\q_{\phi}(\vz_{t}|\vx_{1:t}) \| \p(\vz_t)\right), \text{for } t>F.
\end{multlined}
\normalsize
\label{eq:elbo}
\end{equation}
Leveraging the re-parametrization trick~(\cite{kingma2013auto}) allows efficient optimization of the likelihood loss in Eq.~\ref{eq:elbo}, permitting us to learn the parameters $\theta$ and $\phi$. Note that the expectation term in Eq.~\ref{eq:elbo} boils down to a standard MSE over all predicted frames $\vx_{F+1:T}$ in the training set when the $\prob{\theta}(\cdot)$ term is assumed to follow a Gaussian distribution with an isotropic constant variance. In this setting, the KL divergence in Eq.~\ref{eq:elbo} acts as a regularizer on $\q_{\phi}(\cdot)$ so that this posterior does not just copy an encoding of $\vx_t$ available to it as $\vz_t$, instead captures the density of a latent distribution that is useful to the prediction model in maximizing the first term in Eq.~\ref{eq:elbo}. 

In conditional variational autoencoders~\cite{babaeizadeh2017stochastic,kingma2013auto}, the latent prior $\p(\vz_t)$ is typically assumed to be $\normal(0,1)$ - a choice that can be sub-optimal. A better approach is perhaps to learn this prior so that the stochasticity of the future frame can be guided by the data itself. To this end, Denton and Fergus~\cite{denton2018stochastic} suggests a learned stochastic prior model $\p(\vz_t) = \p_{\psi}(\vz_t) := \p_{\psi}(\vz_t|\vx_{1:t-1})$, parametrized by $\psi$, which is learned by minimizing its divergence from the variational posterior $\q_{\phi}(\cdot)$ through the KL-term in Eq.~\ref{eq:elbo}. As the posterior $q_\phi(\cdot)$ has access to the current input sample $\vx_t$, it can guide the prior (which does not have access to $\vx_t$, but only to $\vx_{1:t-1}$) to produce a distribution on $\vz_t$ that mimics the posterior (and hence we can discard the posterior at test time). Thus, the training-time sampling pipeline is given by: $\vz_t\sim \q_{\phi}(\vz_t|\vx_{1:t}), \vx_t\sim \p_{\theta}(\vx_t|\vx_{1:t-1},\vz_{t})$, and $\p_{\psi}\xleftrightarrow[]{d} \q_{\phi}$ (matching in distribution). 

While learning the stochastic prior $\p_{\psi}(\cdot)$ allows for characterizing the data stochasticity, the model's predictive uncertainty remains unaccounted for. Our hierarchical framework for estimating this uncertainty, follows a two-step process. The first is the estimation of the learned prior, $\p_{\psi}(\cdot)$. The key idea in the second step is to leverage the variance in the learned prior $\p_{\psi}(\cdot)$ to estimate this uncertainty. Since the prior estimation network, $\p_{\psi}(\cdot)$, is retained both during training and inference (unlike the posterior), this permits its usage for downstream tasks, during inference.

\section{Proposed Method}
\begin{figure*}[t]
    \vspace{-0.8cm}
    \centering
    \includegraphics[width=12cm]{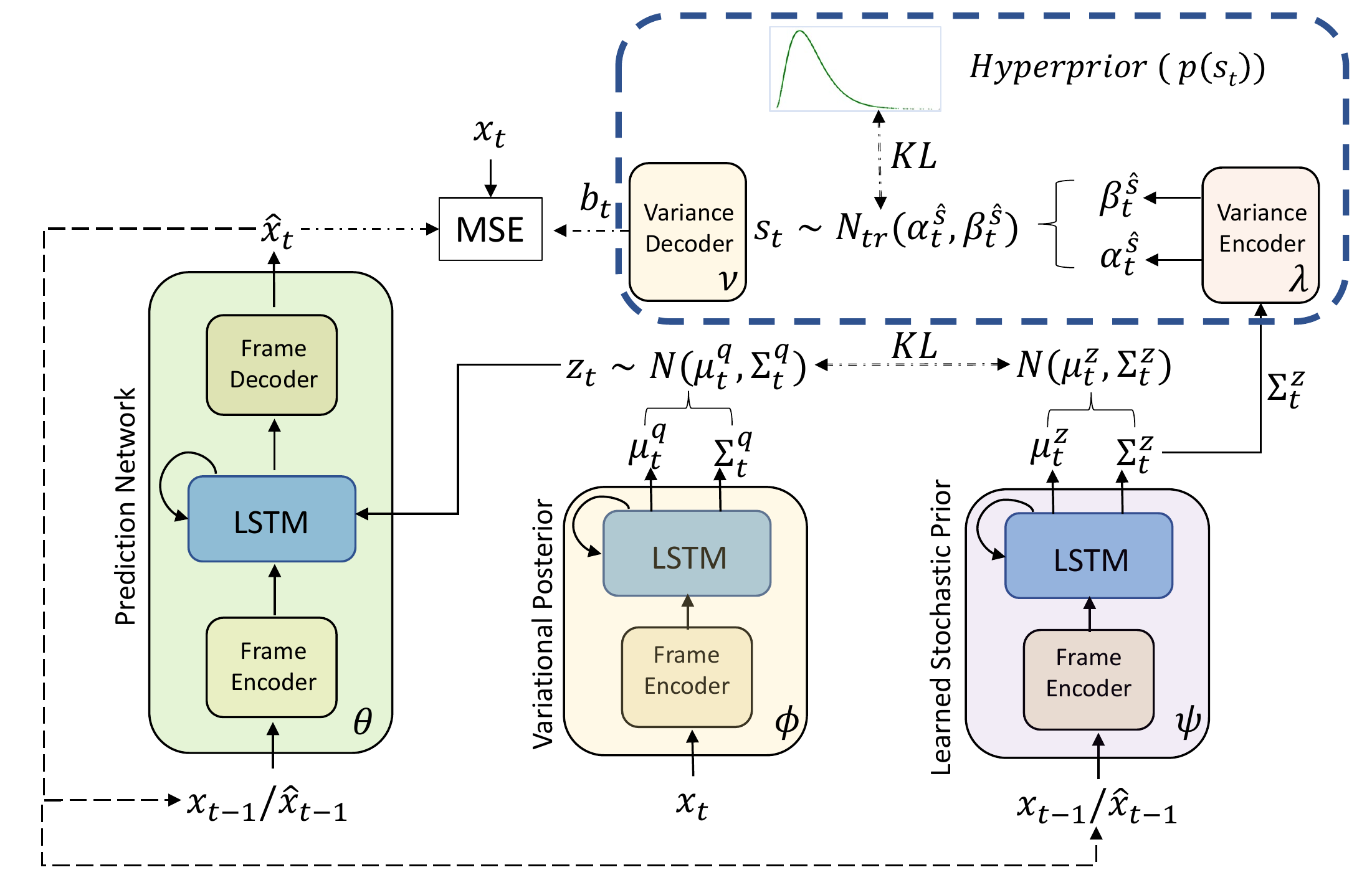}
    \caption{Overview of our approach. } % See text for details.
    \label{fig:overview}
    \vspace{-0.6cm}
\end{figure*}

As alluded to above, we seek to control the prediction model using the uncertainty estimated directly from the stochastic prior. Subsequently, we assume the prediction model consists of an LSTM, $\f_{\theta}$, with weights $\theta$ such that:
\begin{equation}
 \small
    \hat{\vx}_t = \f_{\theta}(\vx_{1:t-1}, \vz_{t}) := \f_{\theta}(\vx_{t-1},\vz_t; \vh^{\theta}_{t-1}),
    \label{eq:3}
 \normalsize
\end{equation}
%$\p(\vx_t|\f_\theta)$
where $\hat{\vx}_t$ denotes the $t^{th}$ generated frame and $h^{\theta}_{t-1}$ captures the internal states of the LSTM. The generated frame $\hat{\vx}_t$ is then sent through the likelihood model to compute the MSE. With this setup, we are now ready to introduce our neural uncertainty quantifier (NUQ). Figure~\ref{fig:overview} provides an overview of our framework. 

\subsection{Neural Uncertainty Quantifier}
\label{sec:nup}

As is standard practice, let us assume the data likelihood model $\prob{\theta}(\vx_t | \vx_{1:t-1},\vz_{t}) \sim \normal(\vx_t, \frac{1}{b_t})$, where $b_t$ denotes the precision (inverse variance) of our prediction model, where $b_t > 0$, $b_t \in \mathbb{R}$. Denton and Fergus~\cite{denton2018stochastic} assumes $b_t$ to be an isotropic constant, such that the negative log-likelihood of the predicted frame $\hat{\vx}_t$ boils down to computing the $\ell_2$-loss. This reduces Eq.~\ref{eq:elbo} to become the evidence lower bound (ELBO)~\cite{kingma2013auto}:
\begin{equation}
\begin{multlined}
\small %\footnotesize
    \loss_{\theta,\phi,\psi} := \sum_{t=F+1}^T \frac{1}{2}\enorm{\hat{\vx}_t-\vx_t}^2 + \\ \kl\left(\q_{\phi}(\vz_t|\vx_{1:t})\ \|\  \p_{\psi}(\vz_t|\vx_{1:t-1})\right).
\normalsize
\end{multlined}
\label{eq:elbo_reduced}
\end{equation}

Our key insight to the proposed uncertainty measure arises from the observation that the $\ell_2$-norm term in~\eqref{eq:elbo_reduced} does not include any dependency on the uncertainty associated with the prediction of $\hat{\vx}_t$. Note that there are two extreme situations when this loss is large that impacts effective training: (i) when there is no uncertainty associated with the generation of $\hat{\vx}_t$, however the prediction model is not trained well, such that $\hat{\vx}_t$ does not match $\vx_t$, and (ii) when there is uncertainty involved in the generative model such that the generated $\hat{\vx}_t$, while plausible given the context, is different from $\vx_t$. Thus, the key research question for effective model training becomes: how can we equip the prediction model to differentiate these situations? Our solution is to directly condition the prediction model with the uncertainty derived from the prior $\p_{\psi}(\vz_t|\vx_{1:t-1})$, so that when the stochasticity is high for the generated frames, the $\ell_2$-loss term is weighed down such that the gradients computed on this term will have  a lesser impact in updating the weights of the neural network; thereby stabilizing the training. 

Suppose our prior $\p_{\psi}(\vz_t|\vx_{1:t-1})$  is a normal distribution $\normal(\vmu_t^\vz, \mSigma_t^\vz)$, with parameters $\vmu_t^\vz$, the mean, and $\mSigma_t^\vz$ the covariance matrix - capturing the predictive uncertainty, in the latent space. For better characterization of this prior model, we assume it to be implemented as an LSTM $\g_{\psi}$  with weights $\psi$ such that $(\vmu_t^{\vz},\mSigma_t^{\vz}) = g_{\psi}(\vx_{t-1}; \vh^{\psi}_{t-1})$, where $\vh^{\psi}_{t-1}$ denotes the hidden state. Similarly, we assume the posterior $\q_{\phi}(\vz_t|\vx_{1:t})$ is  normally-distributed: $\normal(\vmu^q_t, \mSigma^q_t)$, and is implemented using an LSTM $l_{\phi}(\vx_{t}; \vh^{\phi}_{t})$ with weights $\phi$ and hidden state $\vh^{\phi}_t$. This leads us to the following sampling pipeline:
\begin{align}
\small %\footnotesize
    \hat{\vx}_t = \f_{\theta}(\vx_{t-1},\vz_{t}; \vh^{\theta}_{t-1}) &\sim \normal(\vx_t, \frac{1}{b_t}),\\
    \label{eq_prior}\vz_t|\vx_{1:t-1} \sim \normal(\mu^\vz_t, \mSigma^\vz_t)\ &;\  (\mu^\vz_t, \mSigma^\vz_t) = g_{\psi}(\vx_{t-1}; \vh^{\psi}_{t-1}),\\
    \label{eq_post}\vz_t|\vx_{1:t} \sim \normal(\mu^{q}_t, \mSigma^q_t) \ &; \ (\mu^{q}_t, \mSigma^q_t) = l_{\phi}(\vx_{t}; \vh^{\psi}_{t}),
\normalsize
\end{align}

 Using this setup, we are now ready to present our hierarchical, generative, variational  model for uncertainty estimation, an overview of which is shown in Figure~\ref{fig:overview}. To set the stage, let us assume the precision is sampled from the distribution $\p(b_t|\vx_{1:t-1})$. Then, we can rewrite the log-likelihood in Eq.~\ref{eq:pre_elbo} by including the precision distribution as: 
\begin{align}
\label{eq:overall_likelihood}
\int_{b_t,z_t}\log p(x_t|b_t, z_t, x_{1:t-1})&+\log p(z_t|x_{1:t-1})\\
&\qquad+\log p(b_t| x_{1:t-1})\ db_t\ d z_t\notag
\end{align}

The above integral is intractable. Hence, we approximate it by sampling $b_t$ and $z_t$. Note that the first two terms taken together is essentially the left-hand side of Eq.~\ref{eq:pre_elbo}, except with the additional conditioning on $b_t$. Using the variational lower bound~\cite{kingma2013auto}, like in Eq.~\ref{eq:pre_elbo}, we have :
\begin{equation}
\small
\begin{multlined}
\log \p(\vx_t|b_t, \vx_{1:t-1}) \geq \expect{\q_{\phi}(\vz_{t}|\vx_{1:t})} \log \prob{\theta}(\vx_t | \vx_{1:t-1},\vz_{t}, b_t) - \\ \kl\left(\q_{\phi}(\vz_{t}|\vx_{1:t}) \| \p_{\psi}(\vz_t|\vx_{1:t-1}\right), \text{for } t>F.
\end{multlined}
\normalsize
\label{eq:overall_1}
\end{equation}
Please see the supplementary for the derivation. 

As stated before, we seek to connect the uncertainty $\mSigma^\vz_t$ in the latent prior $p_{\psi}(z_t|x_{1:t-1})$ to the precision $b_t$. We accomplish this via a variational encoder-decoder network. Such a formulation permits the flexibility of introducing customized prior distributions on its latent space. During training, the encoder component of this network, $\zeta_{\lambda}(\cdot)$, with parameters $\lambda$, takes as input $\mSigma^\vz_t$, and produces the sufficient statistics of the posterior distribution  $\q_{\lambda}(\cdot)$ governing the latent space of this network, while the decoder $\tau_{\nu}(\cdot)$, with parameters $\nu$ draws a sample $s_t$, from this distribution, and decodes it to generate $b_t$, with a distribution on $b_t$ denoted by $\prob{\nu}(\cdot)$. This sampling scheme is described as follows:
 \begin{equation}
 \small
 \begin{multlined}
  b_t \sim \prob{\nu}(b_t | s_{t}, \vx_{1:t-1}) = p_\nu(b_t|s_t, \Sigma^z_t),\\  \quad s_t \sim \q_{\lambda}(s_{t}|\vx_{1:t-1}) = \q_{\lambda}(s_{t}|\Sigma^z_t)%\qquad
 \end{multlined}
 \label{eq:b_enc_dec}
 \normalsize
 \end{equation}

 In order to provide appropriate regularization for the latent space distribution, $\q_{\lambda}(\cdot)$, we assume a manually-defined hyper-prior distribution governing the latent space of this module denoted $\p(s_t)$. Let the hyper prior $\p(s_t) \sim D_{\gamma}(\alpha_t^s,\beta_t^s)$, with parameters $\alpha_t^s,\beta_t^s$ chosen by the user and let $\q_{\lambda}(s_{t}|\Sigma^z_t) \sim D_{\lambda}(\alpha_t^{\hat{s}},\beta_t^{\hat{s}})$, where the parameters $\alpha_t^{\hat{s}},\beta_t^{\hat{s}}$ are estimated by the encoder network $\zeta_{\lambda}(\cdot)$.
 With this setup, analogous to Eq.~\ref{eq:elbo}, we obtain the following variational lower bound on the likelihood of $b_t$:
\begin{equation}
\small
\begin{multlined}
\log \p(b_t|\vx_{1:t-1}) \geq \expect{\q_{\lambda}(s_{t}|\Sigma^z_t)} \log \prob{\nu}(b_t | s_{t}, \Sigma^z_t) - \\ \kl\left(\q_{\lambda}(s_{t}|\Sigma^z_t)\ \|\  \p(s_t)\right), \text{for } t>F
\end{multlined}
\normalsize
\label{eq:overall_2}
\end{equation}
Please see the supplementary for the derivation. Plugging Eq.~\ref{eq:overall_1} and Eq.~\ref{eq:overall_2} in Eq.~\ref{eq:overall_likelihood}, we have the following:
\begin{equation}
\small
\begin{split}
& \expect{\q_{\phi}(\vz_{t}|\vx_{1:t})} \log \prob{\theta}(\vx_t | \vx_{1:t-1},\vz_{t}, b_t) - \\ &\qquad\qquad\qquad\qquad\qquad\qquad\kl\left(\q_{\phi}(\vz_{t}|\vx_{1:t}) \| \p_{\psi}(\vz_t|\vx_{1:t-1})\right) + \\ & \expect{\q_{\lambda}(s_{t}|\vx_{1:t-1})} \log \prob{\nu}(b_t | s_{t}, \Sigma^z_t) - \kl\left(\q_{\lambda}(s_{t}|\Sigma^z_t) \| \p(s_t)\right)
\end{split}
\normalsize
\end{equation}

 Assuming that $\prob{\theta}(\vx_t | \vx_{1:t-1},\vz_{t}, b_t)$ follows a Gaussian distribution $\normal(\vx_t, \frac{1}{b_t})$, along the lines of Eq.~\ref{eq:elbo_reduced}, leads us to our final ELBO objective, which we minimize:
\begin{equation}
\small
\begin{split}
 & \loss^P_{\theta,\phi,\psi,\lambda} = \sum_{t=F+1}^T\frac{1}{2} \left[b_t \enorm{\hat{\vx}_t - \vx_t}^2 - \log b_t\right]- \\  &
\expect{\q_{\lambda}(s_{t}|\Sigma^z_t)} \log \prob{\nu}(b_t | s_{t}, \Sigma^z_t) + \\ 
& \eta_1 \kl(\q_\phi(\vz_t|\vx_{1:t}) \| \p_{\psi}(\vz_t | \vx_{1:t-1})) + \eta_2 \kl\left(\q_{\lambda}(s_{t}|\Sigma^z_t) \| \p(s_t)\right)
\end{split}
\label{eq:final_obj}
\normalsize
\end{equation}
where $\eta_1,\eta_2\geq 0$ are regularization constants (as suggested in Higgins \etal~\cite{higgins2017beta}).

\begin{figure}[t]
    \vspace{-0.2cm}
     \centering 
     \subfigure[Uncertainty and learned prior]{\includegraphics[scale=0.25]{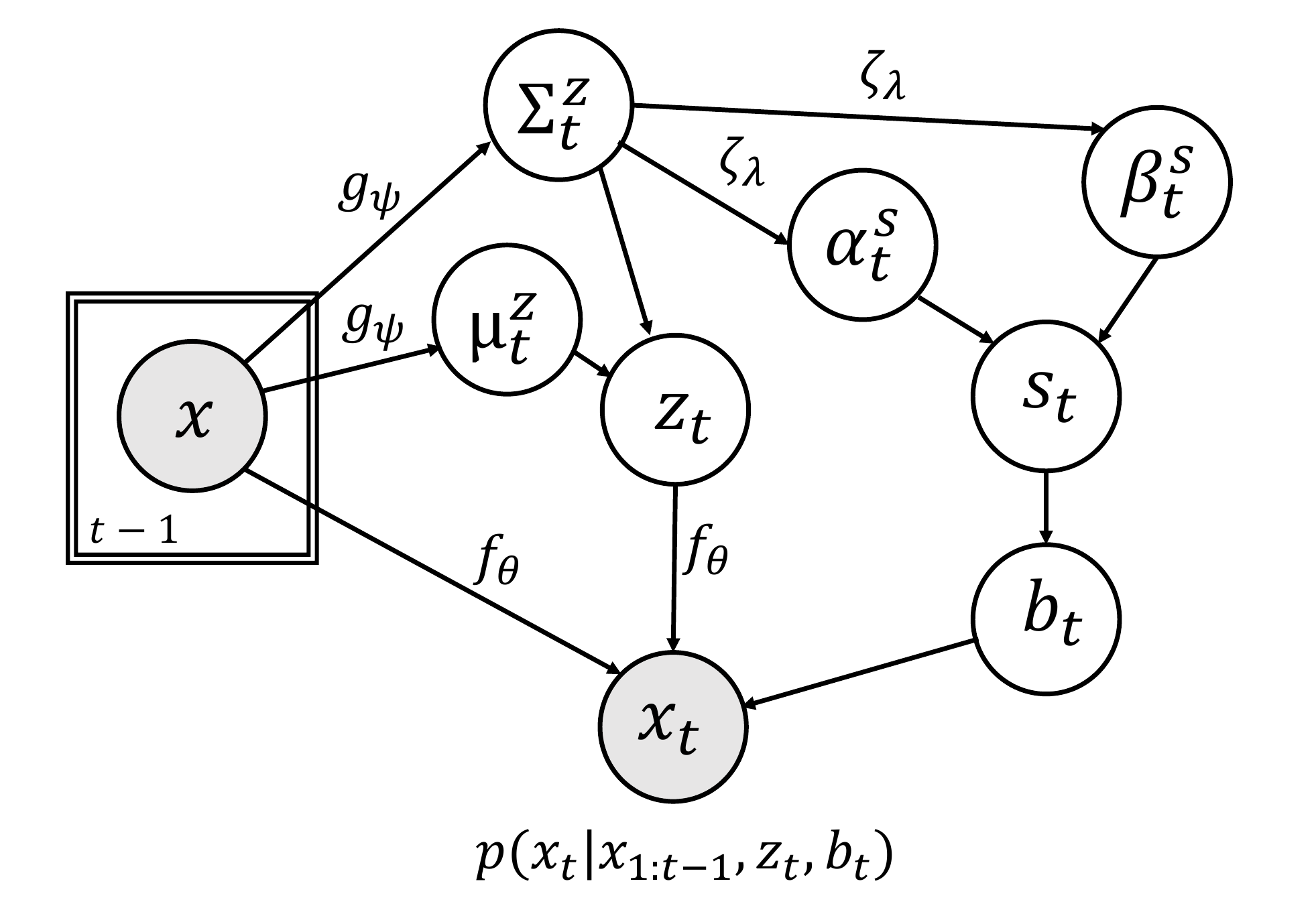}}
     \subfigure[Posterior]{\includegraphics[scale=0.25]{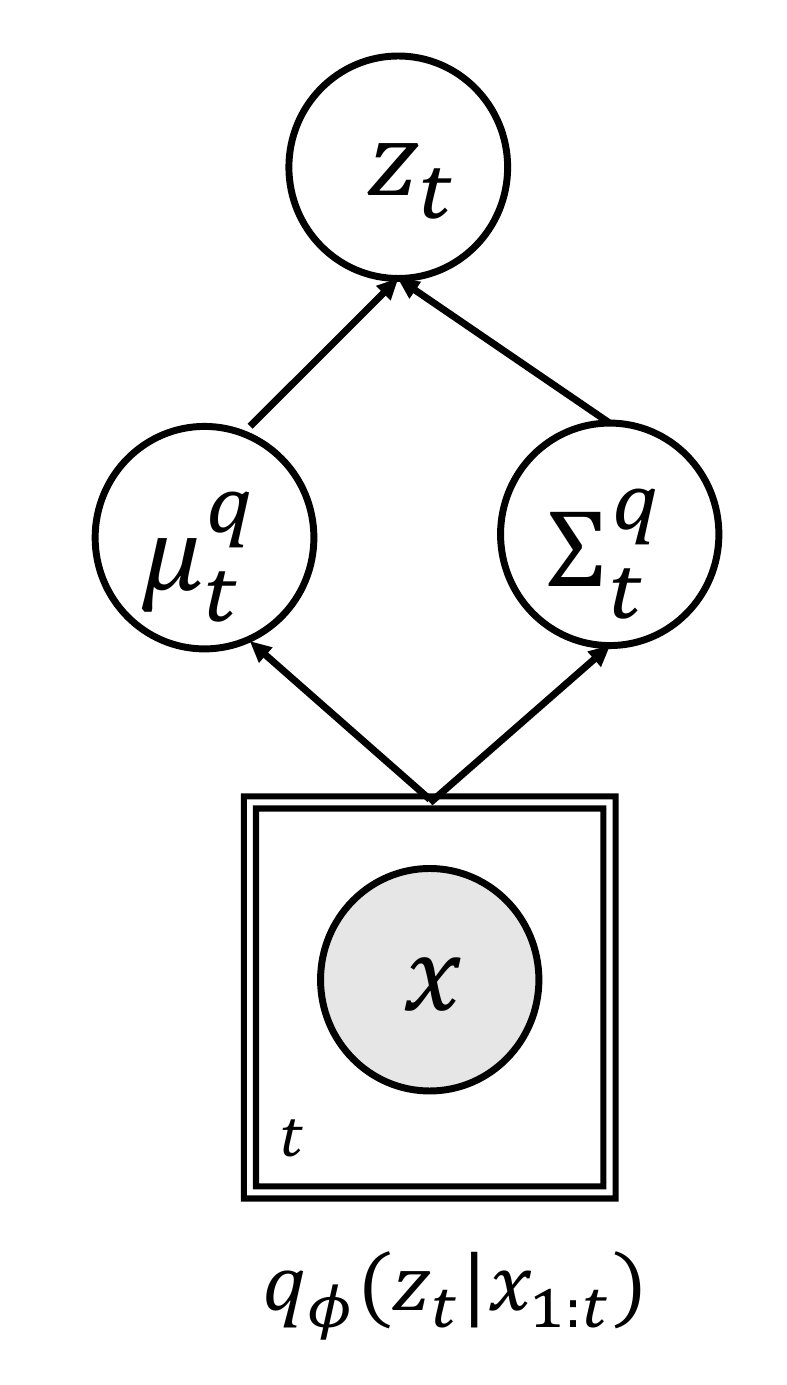}}
     \caption{Plate diagrams depicting the graphical model of our NUQ-framework. (a) shows the sampling dependencies between the learned prior and the uncertainty prediction modules, while (b) shows our posterior sampling framework. The plates denote, for example, $t-1$ repetitions of the random variable $\vx$ in (a).}
     \vspace{-0.5cm}
     \label{fig:plates}
 \end{figure}

Given our setup, a natural choice in a hierarchical Bayesian conjugate sense is to assume $\p(s_t) \sim \Gamma(\alpha_t^s,\beta_t^s)$, the gamma distribution, which is a conjugate prior for the precision. Unfortunately, however using the gamma distribution for the posterior does not permit the reparametrization trick~\cite{knowles2015stochastic,kingma2013auto}, which is essential for making sampling-based networks differentiable. While one may resort to approximations to the gamma prior such as using implicit gradients~\cite{figurnov2018implicit} or generalized re-parameterization techniques~\cite{ruiz2016generalized}, these approaches can be computationally expensive. Instead, we propose to approximate it by a truncated normal distribution $\normal_{tr}(\alpha^{\hat{s}}_t, \beta^{\hat{s}}_t)$ (which is amenable to re-parametrization), where now $\alpha^{\hat{s}}_t (\geq 0$) and $\beta^{\hat{s}}_t$ correspond to the mean and the standard deviation of the truncated normal, respectively and are estimated by the encoder network  $\zeta_{\lambda}(\cdot)$. In practice, the truncation is effected through rejection sampling~\cite{robert1995simulation}; i.e., we sample from a normal distribution, and reject samples if they are negative. Empirically, we find this choice of the hyper prior (being a gamma distribution) and our truncated-normal posterior combination to be beneficial. Thus, the KL divergence in~\eqref{eq:overall_2} will eventually promote the network $\zeta_{\lambda}(\cdot)$ to produce the sufficient statistics of a truncated-normal distribution which will closely approximate the true gamma hyper-prior governing $\p(s_t)$. Additionally, since $s_t$ is a scalar ($\in \mathbb{R}$), thus $b_t$ ($=\frac{1}{s_t}$) is directly sampled from $q_{\lambda}(\cdot)$ rather than through the decoder network, $\tau_{\nu}(\cdot)$ . Figure~\ref{fig:plates} presents a plate diagram of our proposed hierarchical graphical model.

\subsection{Sequence Discriminator}
\label{sec:discriminator}
Inspired by the success of generative adversarial networks in generating realistic images and realistic object motions~\cite{goodfellow2014generative,goodfellow2014generative,gulrajani2017improved,barsoum2018hp,liang2017dual,kwon2019predicting}, as well as the synergy that GANs bring about in improving the quality of other generative models~\cite{cherian2020sound2sight,gur2020hierarchical,bao2017cvae}, we propose to integrate NUQ with a sequence discriminator, where the generated frame sequences are input to a discriminator that checks for their realism and motion coherence. Different from prior approaches that employ GANs for future frame prediction~\cite{liang2017dual,kwon2019predicting}, our discriminator ($\disc_{\vw}$) is a recurrent neural network with weights $\vw$. It takes as input $k$ contiguous frames with image dimensions $\delta_h\times \delta_w$, and produces a non-negative score, denoting the probability of that sequence being real or fake. Thus, $\disc_{\vw}:\reals{\delta_h\times \delta_w\times k}\to \reals{}_+$. Suppose, $\vy_{1:k}\subset\vx_{1:T}$ represents $k$ contiguous frames starting at a random time step from video sequences $\vx_{1:T}$ in the training dataset $\dataset$. If $\hat{\vx}_{t-k+1:t}$ represents a sequence of $k$ generated frames, then our discriminator loss is given by: 

\begin{align}
\label{eq:seq_disc}
\small
    &\loss_{\vw}^{\disc} := -\!\!\sum_{t=F+1}^T\!\!\!\expect{\vy_{1:k}\sim\dataset}\log\left[\disc_{\vw}(\vy_{1:k})\right] +       \\
    &\expect{\hat{\vx}_{t-k+1:t}\sim\p(\hat{\vx}_{t-k+1:t}|\vx_{1:t-k},\vz_{1:t})} \log \left[1-\disc_{\vw}(\hat{\vx}_{t-k+1:t})\right],\notag
\normalsize
\end{align}

\noindent where, the discriminator is trained to distinguish between the generated (with label zero) and real (with label one) input sequences, by minimizing $\loss_{\vw}^{\disc}$, while the generator tries to maximize it. 
Combining the ELBO loss in~\eqref{eq:final_obj} with the generator loss, we have our modified training loss for this variant, given by $\loss = \loss^P_{\theta,\phi,\psi,\lambda} - \gamma \loss_{\vw}^{\disc}$, where $\gamma > 0$ is a small regularization constant. For both variants (Eq.~\ref{eq:final_obj} or Eq.~\ref{eq:seq_disc}), we optimize the final objective using ADAM~\cite{kingma2014adam}.

\section{Experiments}
\label{sec:expts}
 \begin{table*}[t]
 \vspace{-0.2cm}
 \small
 \centering
 \caption{SSIM, PSNR, and LPIPS scores on the  test set for different datasets after @1, @5, and @Convergence (C) (upto 150 epochs) epochs of training with varying training set sizes. [Key: Best results in \textbf{bold} and second-best in \textcolor{blue}{blue}.] }\label{tab:smmnist_bair_kth}
 \begin{tabular}{l|c|c|c||c|c|c||c|c|c}
\hline \hline
\multirow{2}{*}{\textbf{Dataset: SMMNIST}} & \multicolumn{3}{c}{\textbf{SSIM $\uparrow$}} & \multicolumn{3}{c}{\textbf{PSNR $\uparrow$}} & \multicolumn{3}{c}{\textbf{LPIPS $\downarrow$}} \\ \cline{2-10}
& @1 & @5 &	@C & @1 & @5 &	@C & @1 & @5 &	@C \\ \hline
\multicolumn{10}{c}{\textit{Number of training samples - 2,000}} \\ \hline
%& @1 & @5 &	@ C & @1 & @5 &	@ C & @1 & @5 &	@C \\ \hline
 Ours & \textbf{0.8686}	& \textbf{0.8638} &	\textbf{0.8948} &	\textcolor{blue}{17.76} &	\textbf{18.13} &	\textcolor{blue}{18.14} & \textbf{0.3087} & \textbf{0.2836} & \textbf{ 0.1803} \\ \hline
 Ours (w/o discriminator) & \textcolor{blue}{0.8599}	& \textcolor{blue}{0.8825} &	\textcolor{blue}{0.8929} &	 \textbf{17.82} &	\textcolor{blue}{18.07} &	\textbf{18.48}  & \textcolor{blue}{0.3283} &\textcolor{blue}{0.3158} & \textcolor{blue}{0.1967} \\ \hline
 Denton and Fergus~\cite{denton2018stochastic} & 0.8145 &	0.8650 &	0.8696 &	17.07 &	{18.05} &	{18.13} & 0.3429 & 0.3345  & 0.2321 \\
 Castrejon \etal~\cite{castrejon2019improved} & 0.8564 &	{0.8748} &	{0.8868} &	{17.36} &	17.98 &	18.12 & 0.3392 & 0.3432  & 0.2262 \\
 Hsieh \etal~\cite{hsieh2018learning}  & 0.4538 &	 0.8419 &	0.8569 &	11.27 &	16.40 &	16.70 & 0.4370 & 0.3696 & 0.2842 \\ \hline \hline
 \multicolumn{10}{c}{\textit{Number of training samples - 8,000 (full training set)}} \\ \hline
& @1 & @5 &	@ C & @1 & @5 &	@ C & @1 & @5 &	@C \\ \hline
 Ours & \textbf{0.8524}	& \textbf{0.8610} &	\textbf{ 0.9088} &	\textbf{17.93} &	\textcolor{blue}{18.14} &	\textbf{19.07} & \textbf{0.3787} & \textbf{0.3013} & \textcolor{blue}{0.1149} \\ \hline
 Denton and Fergus~\cite{denton2018stochastic} & 0.8154 &	0.8607 &	0.8819 &	\textcolor{blue}{17.49} &	\textbf{18.22} &	\textcolor{blue}{18.30}  & 0.4061 & 0.3626 & 0.2813 \\
 Castrejon \etal~\cite{castrejon2019improved} & \textcolor{blue}{0.8640} &	\textcolor{blue}{0.8708} &	\textcolor{blue}{0.8868} &	17.23 &	18.06 &	18.27 & \textcolor{blue}{0.3939} & \textcolor{blue}{0.3316}  & \textbf{0.1040}  \\
 Hsieh \etal~\cite{hsieh2018learning}  & 0.5328 &	 0.8374 &	0.8801 &	11.46 &	16.65 &	16.70  & 0.4217 & 0.4039 & 0.2747 \\ \hline \hline
 \end{tabular}

\begin{tabular}{l|c|c|c||c|c|c||c|c|c}
\hline \hline
\multirow{2}{*}{\textbf{Dataset: BAIR Push}} & \multicolumn{3}{c}{\textbf{SSIM  $\uparrow$}} & \multicolumn{3}{c}{\textbf{PSNR $\uparrow$}} & \multicolumn{3}{c}{\textbf{LPIPS $\downarrow$}} \\ \cline{2-10}
& @1 & @5 &	@C & @1 & @5 &	@C & @1 & @5 &	@C \\ \hline
 \multicolumn{10}{c}{\textit{Number of training samples - 2,000}} \\ \hline
%& @1 & @5 &	@ C & @1 & @5 &	@ C \\ \hline
 Ours & \textbf{0.7709}	& \textbf{0.8230} &	\textbf{0.8314} &	\textbf{18.40} &	\textbf{19.15} &	\textbf{19.26} & \textbf{0.3394} & \textbf{0.2014}& \textbf{0.1574} \\ \hline
 Denton and Fergus~\cite{denton2018stochastic} & \textcolor{blue}{0.7351} &	0.7853 &	0.8196 &	\textcolor{blue}{17.32} &	17.44 &	18.49 & 0.3531& 0.3197 & \textcolor{blue}{0.1725} \\
 Castrejon \etal~\cite{castrejon2019improved} & 0.7094 &	\textcolor{blue}{0.7961} &	\textcolor{blue}{0.8221} &	17.19 &	\textcolor{blue}{17.92} &	\textcolor{blue}{18.79} & \textcolor{blue}{0.3433} & \textcolor{blue}{0.2560} & 0.1742 \\
 Hsieh \etal~\cite{hsieh2018learning}  & 0.4979 &	 0.7901 &	0.7989 &	11.32 &	15.28 &	16.00 & 0.4159& 0.3899 & 0.1891 \\ \hline \hline
 \multicolumn{10}{c}{\textit{Number of training samples - 43,264 (full training set)}} \\ \hline
& @1 & @5 &	@ C & @1 & @5 &	@ C & @1 & @5 &	@C \\ \hline
 Ours & \textbf{0.8135}	& \textbf{0.8336} &	\textbf{0.8460} &	\textbf{19.03} &	\textbf{19.14} &	\textcolor{blue}{19.31} & \textbf{0.1656} & \textbf{0.1470} & \textcolor{blue}{0.1296} \\ \hline
 Denton and Fergus~\cite{denton2018stochastic} & 0.7782 &	0.8198 &	0.8328 &	\textcolor{blue}{18.30} &	18.38 &	18.81 & 0.2119  & 0.1843 & 0.1499 \\
 Castrejon \etal~\cite{castrejon2019improved} & \textcolor{blue}{0.7816} &	\textcolor{blue}{0.8309} &	\textcolor{blue}{0.8437} &	18.29 &	\textcolor{blue}{18.56} &	\textbf{19.59} & \textcolor{blue}{0.1878} & \textcolor{blue}{0.1720} & \textbf{0.1181} \\
 Hsieh \etal~\cite{hsieh2018learning}  & 0.7507 &	 0.8123 &	0.8323 &	16.52 &	16.61 &	16.61 & 0.2140 & 0.1829 & 0.1713 \\ \hline \hline
 \end{tabular}

\begin{tabular}{l|c|c|c||c|c|c||c|c|c}
\hline \hline
\multirow{2}{*}{\textbf{Dataset: KTH Action}} & \multicolumn{3}{c}{\textbf{SSIM $\uparrow$}} & \multicolumn{3}{c}{\textbf{PSNR $\uparrow$}} & \multicolumn{3}{c}{\textbf{LPIPS $\downarrow$}} \\ \cline{2-10}
& @1 & @5 &	@C & @1 & @5 &	@C & @1 & @5 &	@C \\ \hline
 \multicolumn{10}{c}{\textit{Number of training samples - 1,911  (full training set)}} \\ \hline
 Ours & \textbf{0.7990}	& \textbf{0.8192} &	\textcolor{blue}{0.8448} &	\textbf{22.62} &	\textcolor{blue}{22.89} &	\textbf{24.02} & \textbf{0.4309} & \textbf{ 0.3390} & \textbf{0.2238} \\ \hline
 Denton and Fergus~\cite{denton2018stochastic} & \textcolor{blue}{0.7028} &	\textcolor{blue}{0.8056} &	0.8374 &	21.29 &	\textbf{22.93} &	24.73 & 0.4621 & 0.3580 & {0.2497} \\
 Castrejon \etal~\cite{castrejon2019improved} & 0.6345 &	0.8054 &	\textbf{0.8510} &	\textcolor{blue}{21.31} &	21.12 &	\textcolor{blue}{24.82} & \textcolor{blue}{0.4513} & \textcolor{blue}{0.3471} & \textcolor{blue}{0.2395} \\
 Hsieh \etal~\cite{hsieh2018learning}  &  0.4647 &	0.5335  & 0.7057 & 11.25 &	12.32 &	16.44 &0.5189 & 0.3939 & 0.2771 \\ \hline \hline
\end{tabular}

\begin{tabular}{l|c|c|c||c|c|c||c|c|c}
\hline \hline
\multirow{2}{*}{\textbf{Dataset: UCF-101}} & \multicolumn{3}{c}{\textbf{SSIM $\uparrow$}} & \multicolumn{3}{c}{\textbf{PSNR $\uparrow$}} & \multicolumn{3}{c}{\textbf{LPIPS $\downarrow$}} \\ \cline{2-10}
& @1 & @5 &	@C & @1 & @5 &	@C & @1 & @5 &	@C \\ \hline
 \multicolumn{10}{c}{\textit{Number of training samples - 11,425  (full training set)}} \\ \hline
Ours & \textbf{0.7359}	& \textbf{0.7636} &	\textcolor{blue}{0.7729} &	\textbf{21.25} &	\textbf{21.98} &	\textbf{22.73} & \textbf{0.3914} & \textbf{0.2865} & \textbf{0.0836} \\ \hline
Denton and Fergus ~\cite{denton2018stochastic} & {0.6253} &	{0.7540} &	0.7603 &	19.35 &	{20.60} &	21.64 & 0.3507 & 0.3006 & 0.1259 \\
Castrejon \etal ~\cite{castrejon2019improved} & \textcolor{blue}{0.6712} &	\textcolor{blue}{0.7555} &	\textbf{0.7756} &	\textcolor{blue}{20.58} &	\textcolor{blue}{20.58} &	\textcolor{blue}{22.53} & \textcolor{blue}{0.3414} & \textcolor{blue}{0.2965} & \textcolor{blue}{0.1036} \\
Hsieh \etal ~\cite{hsieh2018learning}  &  0.6199 &	0.6800  & 0.7103 & 16.65 &	17.18 &	18.41 &0.3989 & 0.3239 & 0.1771 \\ \hline 
\end{tabular}

\normalsize
\end{table*}

In this section, we empirically validate the efficacy of NUQ on challenging real-world and synthetic datasets.

\noindent \textbf{Datasets:} We conduct experiments on four standard stochastic video prediction datasets, namely (i) Stochastic Moving MNIST (SMMNIST)~\cite{denton2018stochastic,cherian2020sound2sight}, (ii) BAIR Robot Push~\cite{ebert2017self}, (iii) KTH-Action~\cite{schuldt2004recognizing}, and (iv) UCF-101~\cite{ucf2012doc}. In SMMNIST, a hand-written digit moves in rectilinear paths within a $48\times 48$ pixel box, bouncing off its walls, where the post bounce movement directions are stochastic. The dataset has a test set size of 1,000 videos~\cite{cherian2020sound2sight}.  The BAIR Push Dataset~\cite{ebert2017self} consists of $64 \times 64$ pixel videos featuring highly stochastic motions of a Sawyer robotic arm pushing objects on a table. This dataset has 257 test samples using the split of Denton and Fergus~\cite{denton2018stochastic}. The KTH Action Dataset~\cite{schuldt2004recognizing} is a small dataset of $64 \times 64$ pixel videos containing a human performing various actions (walking, jogging, etc.), captured in a controlled setting with a static camera. The test set for this dataset consists of 476 videos. Finally, the UCF-101 Dataset~\cite{ucf2012doc} is a dataset of videos, resized to $64 \times 64$ containing 101 common human action categories (such as pushups, cricket shot, etc.), spanning both indoor and outdoor activities. The test set for this dataset consists of 1,895 videos. We hypothesize that by incorporating the predictive uncertainty, NUQ undergoes more efficient training updates and can thus train with fewer samples, efficiently. We therefore conduct experiments with varying training set sizes for some of these datasets.

\noindent \textbf{Baselines and Evaluation Metrics: } To carefully evaluate the performance improvement brought about by incorporating our uncertainty estimation method into a stochastic video generation framework, we choose three competitive and closely-related state-of-the-art methods within the stochastic video prediction realm as baselines, namely: (i) Denton and Fergus~\cite{denton2018stochastic}, (ii) Castrejon \etal~\cite{castrejon2019improved},  and (ii) Hsieh \etal~\cite{hsieh2018learning}. At test time, we follow the standard protocol of generating 100 sequences for all models and report performances on sequences that matches best with the ground truth~\cite{denton2018stochastic}. To quantify the generation quality, we use standard evaluation metrics: (i) per-frame Structural Similarity (SSIM)~(\cite{channappayya2008rate}), (ii) Peak Signal to Noise Ratio (PSNR), and (iii) Learned Perceptual Image Patch Similarity (LPIPS)~\cite{zhang2018unreasonable} - with a VGG backbone. We report the average scores on these metrics across all predicted frames.

\begin{figure}[t]
\vspace{-0.5cm}
   \centering
    \subfigure[SSIM Measure]{\includegraphics[width=3.0cm,trim={9.4cm 0cm 10cm 0cm},clip]{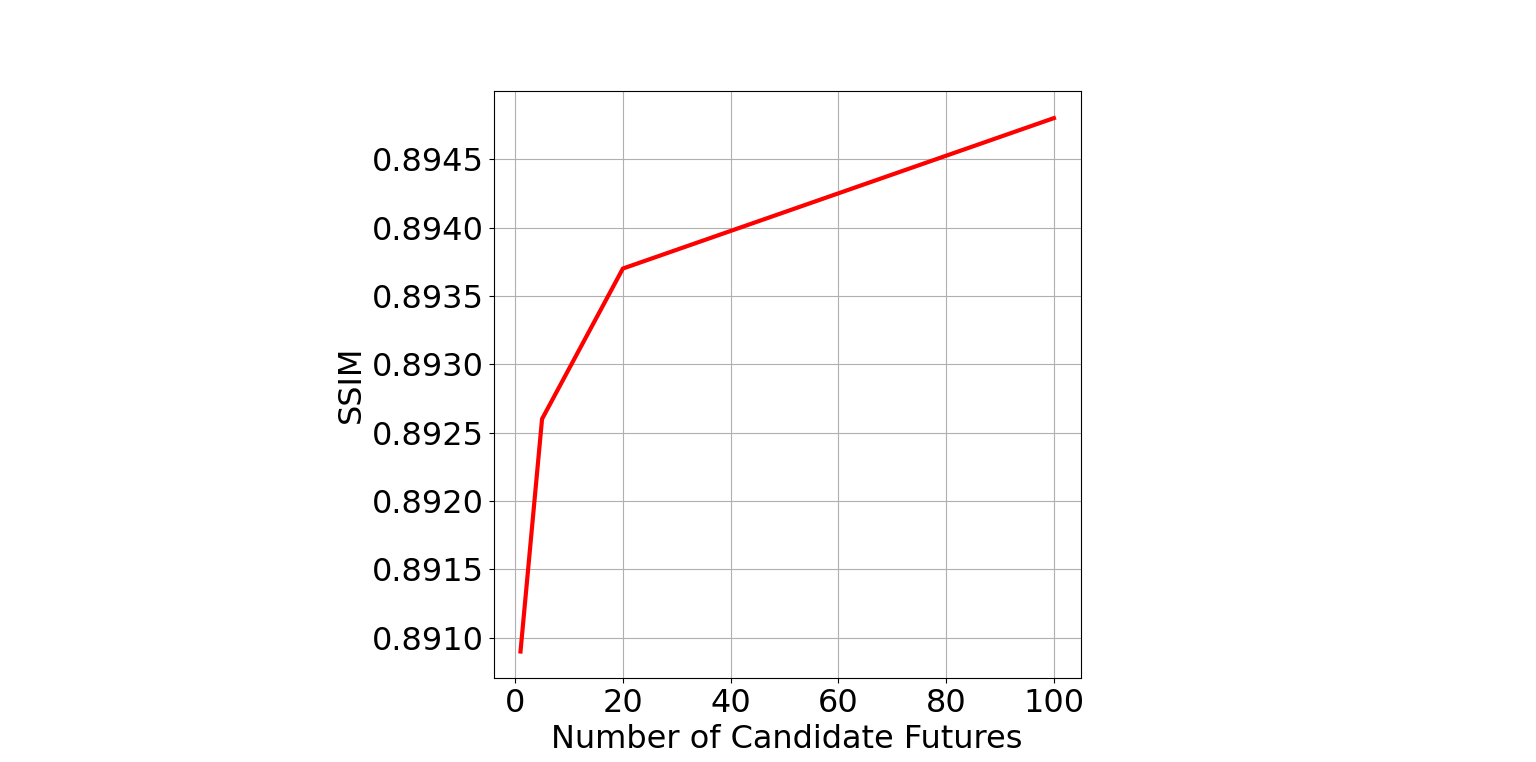}  }
    %\hspace{-0.5cm}
    \subfigure[Qualitative]{\includegraphics[width=5.0cm,trim={0cm 0cm 7cm 0cm},clip]{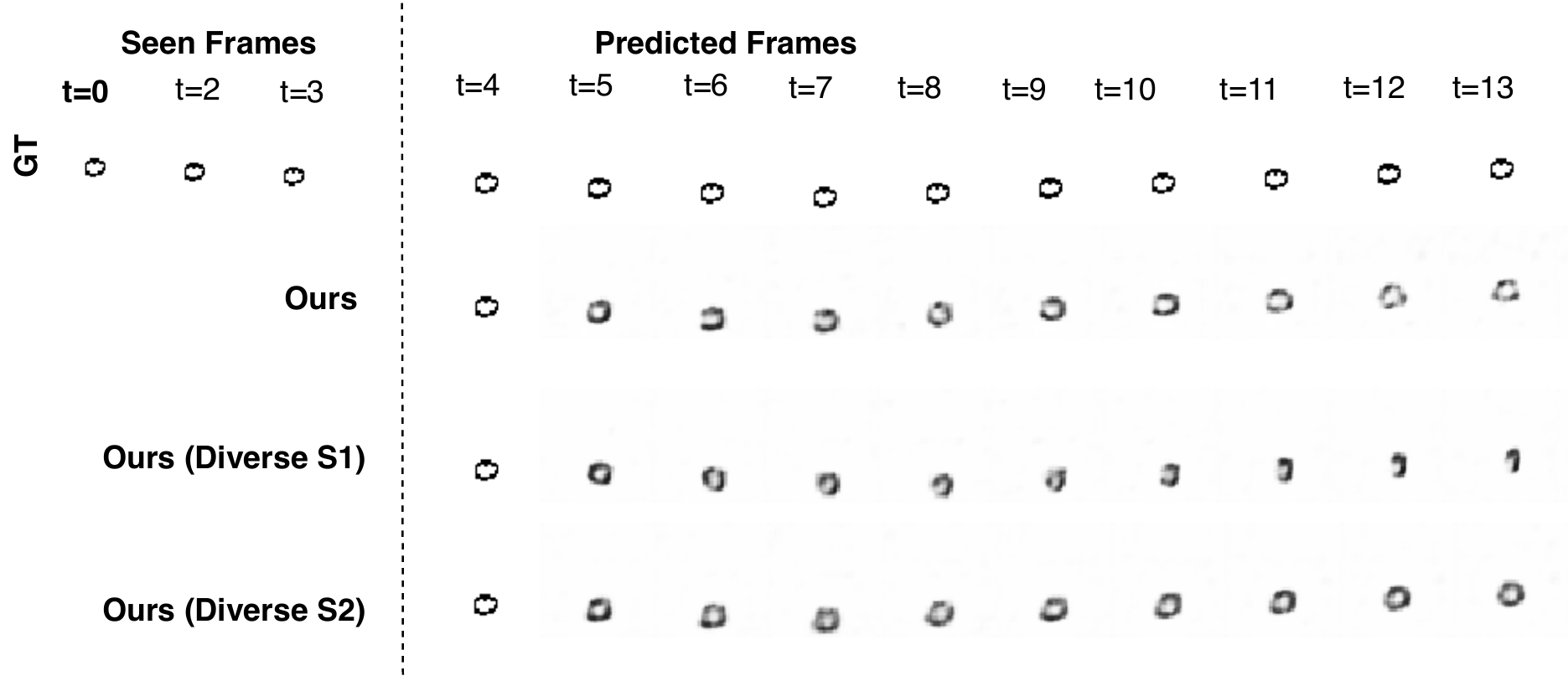} }
  \caption{ Diversity Results: SSIM score with increasing number of generated futures, per time step, on the SMMNIST test set, as a quantification of output diversity (left) and qualitative generation results (right) using NUQ trained with 2000 training samples.}
    \label{fig:SMMNIST_div}
    \vspace*{-0.5cm}
\end{figure}

\begin{table*}[th]
\small
\vspace*{-0.2cm}
\centering
\caption{SSIM, PSNR, and LPIPS scores on the SMMNIST test set after @1, @5, and @Convergence (C) (upto 150 epochs) epochs of training with alternative formulations of our model using 2,000 training samples. [Key: Best results in \textbf{bold}].}\label{tab:smmnist_ablation}
\begin{tabular}{l|c|c|c||c|c|c||c|c|c}
\hline \hline
\multirow{2}{*}{\textbf{Dataset: SMMNIST}} & \multicolumn{3}{c}{\textbf{SSIM $\uparrow$}} & \multicolumn{3}{c}{\textbf{PSNR $\uparrow$}} & \multicolumn{3}{c}{\textbf{LPIPS $\downarrow$}} \\ \cline{2-10}
& @1 & @5 &	@C & @1 & @5 &	@C & @1 & @5 &	@C \\ \hline
$\p(s_t) \sim \text{Uniform}[0, 1]$  & 0.8173  &	0.8374  & 0.8523	 & 17.6	 & 17.95	 & 18.06 & 0.3442 & 0.3038 & 0.198	 \\
Estimate $b_t$ from the decoder $p_{\theta}(\cdot)$  & {0.7627}	& 0.7628 &{0.7828} &	 {17.54} &	17.55 &	{17.55} & 0.3463 & 0.3259 & 0.2225 \\ 
Estimate $b_t$ w/o variance encoder-decoder  	& 0.7450 & 0.7454	  & 0.7648	 & 16.22	 & 16.53	 &	16.78 & 0.3469 & 0.3263 & 0.2328 \\ \hline
NUQ (Ours) & \textbf{0.8686}	& \textbf{0.8638} &	\textbf{0.8948} &	\textbf{17.76} &	\textbf{18.13} &	\textbf{18.14} & \textbf{0.3087} & \textbf{0.2836} & \textbf{ 0.1803} \\  \hline \hline
\end{tabular}
\normalsize
\vspace*{-0.4cm}
\end{table*}

\begin{figure}[t]
    \centering
    \includegraphics[width=6.9cm,trim={0cm 0cm 7cm 0cm},clip]{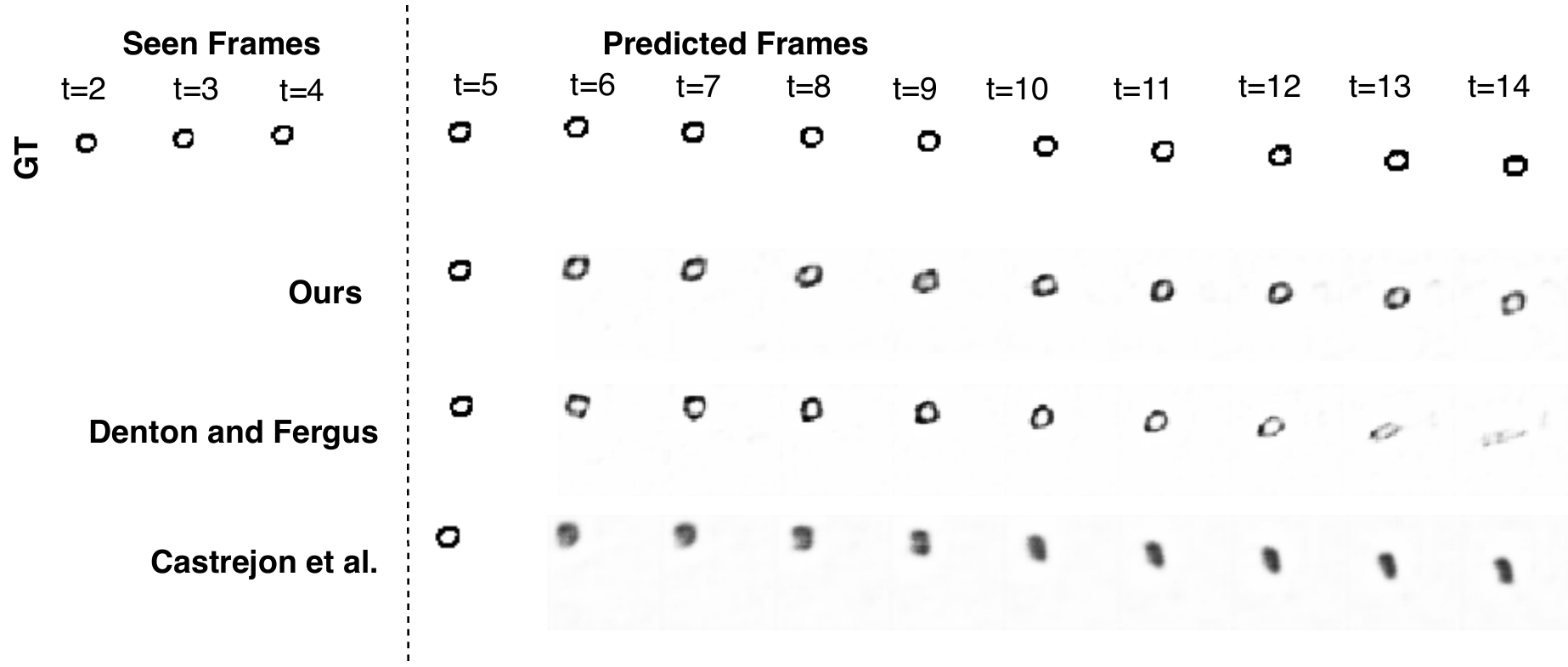}\qquad
    \includegraphics[width=4.8cm,trim={1cm 1cm 0cm 0cm},clip]{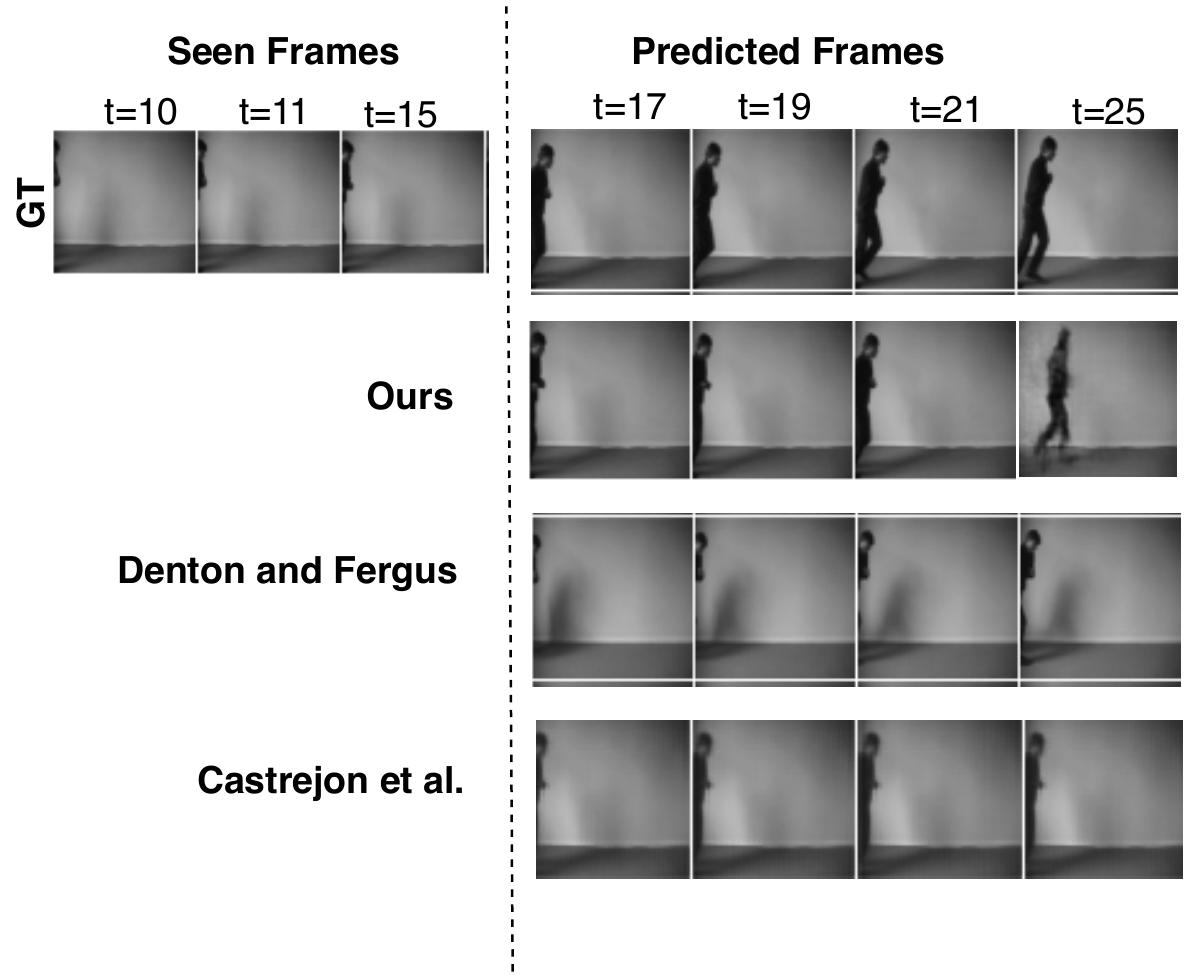}
    %\vspace{-0.3cm} 
   \caption{Qualitative results of NUQ against competing baselines on SMMNIST (top) and on KTH-Action (bottom).}
    \label{fig:smmnist_kth_comp}
    \vspace*{-0.4cm}
\end{figure}

\begin{figure}[t]

    \centering
    \includegraphics[width=8cm,trim={0cm 8cm 0cm 0cm},clip]{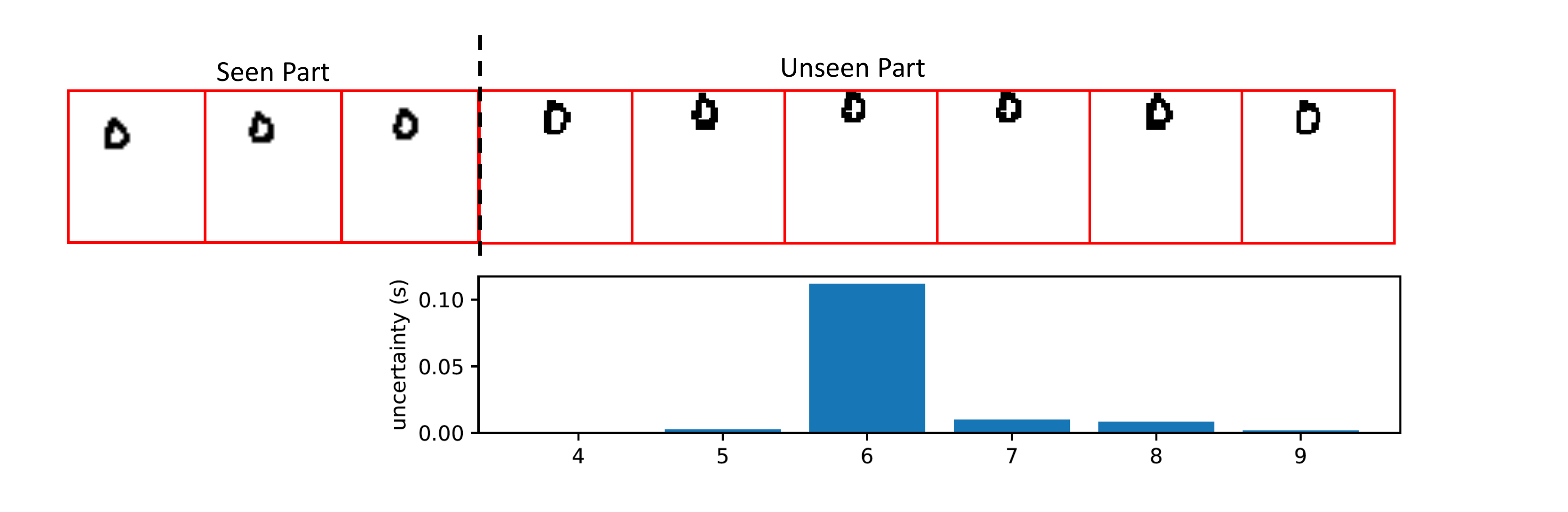}
   \caption{Evolution of scaled uncertainty on the SMMNIST dataset (trained with 2,000 training samples) against time-steps. }
    \label{fig:smmnist_prob}
    \vspace*{-0.2cm}
\end{figure}

\begin{table}[th]
%\scriptsize
\vspace*{-0.1cm}
\centering
\caption{Human preference scores for samples generated using NUQ versus competing baselines across different datasets, }\label{tab:human_pref} 

\begin{tabular}{l||c}
\hline \hline
\textbf{Dataset (\# Training Samples)} & \textbf{Prefer: Ours/~\cite{castrejon2019improved}}/~\cite{denton2018stochastic} \\ \hline
\textbf{SMMNIST (2000 samples)} & \textbf{89\%} / 11\% / 0 \%	 \\
\textbf{BAIR Push (2000 samples)} & \textbf{78\%} / 22\% / 0 \%	 \\
\textbf{KTH-Action (1911 samples)} & \textbf{78\%} / 11\% / 11 \%	 \\ \hline
\end{tabular}
\vspace*{-0.6cm}
\end{table}

\noindent\textbf{Experimental Setup:} 
For SMMNIST, BAIR Push, and KTH Action, we train all methods with 5 seen and 15 unseen frames, while at test time 20 frames are predicted after the first 5 seen ones.  When training with UCF-101, 15 seen and 10 unseen frames are used, while at test time the number of unseen frames is set to 15. For the baseline methods~\cite{denton2018stochastic,castrejon2019improved,hsieh2018learning}, we use the publicly available implementations from the authors.  To ensure our proposed NUQ-framework is similar in learning capacity, we use the ``DCGAN encoder-decoder'' architecture for the frames, as in Denton and Fergus~\cite{denton2018stochastic}, for all datasets. Our variance encoder network $\zeta_{\lambda}(\cdot)$ is a multi-layer perceptron with 2 layers, and our sequence discriminator is an LSTM with a single $256$-d hidden layer. More architectural details are in the Supplementary Materials. We set $k=3$ in ~\eqref{eq:seq_disc}, and $\gamma$ to be 0.00001. We use $\eta_1=0.0001$ and $\eta_2=0.001$ in~\eqref{eq:final_obj}. Learning rate is set to 0.002 and no scheduling is used. The hyper-parameters for the baseline methods are chosen using a small validation set ($\sim 5\%$ of training data). 

\noindent \textbf{State-of-the-Art Comparisons:}
 In Table~\ref{tab:smmnist_bair_kth}, we report quantitative evaluations of our model versus competing baselines across the four datasets. We observe that both variants of the NUQ outperform recent competitive baselines by wide margins on all measures (upto 10\% in SSIM), with the one with the discriminator being slightly better - underscoring the benefits of adversarial training. While noticeable gains are obtained across the board, NUQ really shines under limited training set sizes. We surmise that this gain is attributable to the failure of the baseline methods in incorporating predictive uncertainty explicitly into the learning objective. 
 
 From the table, we also see that NUQ converges faster than other methods both for small and large training set sizes.  Figures~\ref{fig:bair_sota} and~\ref{fig:smmnist_kth_comp} show sample generation results from SMMNIST, BAIR Push, and KTH-Action datasets versus competing baselines, trained with 2,000 and 1,911 samples, respectively. From these figures, we see that compared to baseline methods, frames generated by our method are superior at capturing both the appearance and the motion of the object (i.e. digit/robot arm/human) even under limited training data. Human annotators, when presented with a few random sample generations by NUQ versus competing methods, overwhelmingly choose NUQ samples for their realism, as shown in Table~\ref{tab:human_pref}. Figure~\ref{fig:smmnist_prob} shows the evolution of a scaled uncertainty estimate derived from $\frac{1}{b_t}$ over different frames. The plot shows the increase in uncertainty co-occurs with the bounce of the digit against the boundary, suggesting that the uncertainty is well grounded in the data. 
 
\noindent\textbf{Alternative Formulations:} Next, we discuss the results of some alternative formulations of our model. We consider: (i) replacing the  gamma hyperprior on $\p(s_t)$ with $\uniform(0,1)$ distribution, (ii) estimating $b_t$ from the frame decoder $p_{\theta}(\cdot)$ by producing a diagonal covariance matrix, and (iii) assuming a deterministic mapping from $\mSigma_t^\vz$ to $b_t$ through a multi-layer perceptron.  Table~\ref{tab:smmnist_ablation} presents the performance of these alternatives on the SMMNIST dataset. From the first row of the table, we see that choosing the $\uniform(0,1)$  as priors results in suboptimal variants of NUQ. Further, the results also show that either estimating $b_t$ directly from the decoder $p_{\theta}(\cdot)$, or computing it deterministically from $\mSigma^{\vz}_t$ performs poorly, suggesting that such estimation techniques are not ideal. % for $b_t$

\noindent\textbf{Diversity:} In Figure~\ref{fig:SMMNIST_div} (a), we plot the average SSIM of NUQ for a set of futures, with increasing cardinality of this set. Our plot shows that the SSIM increases with larger number of futures, suggesting that the possibility of matching with a ground truth future increases with more futures, implying better diversity of our model. Figure~\ref{fig:SMMNIST_div}(b) presents diverse generation results on the SMMNIST dataset by NUQ.

\section{Conclusions}
\label{sec:conclude}
Recent approaches have demonstrated the need for modeling data uncertainty in video prediction models. However, in this paper we show that the state of the art in such stochastic methods do not leverage the model's predictive uncertainty to the fullest extent. Indeed, we show that explicitly incorporating this uncertainty into the learning objective via our proposed Neural Uncertainty Quantifier (NUQ) framework, can lead to faster and more effective model training even with fewer training samples, as validated by our experiments.
%\input{ethics}

%%%%%%%%% BODY TEXT

{\small
\bibliographystyle{ieee_fullname}
\bibliography{stochastic}
}
\appendix
\section{Uncertainty Visualizations}
\label{sec:uncertainty}

\begin{figure}[t]
    \centering
    \includegraphics[width=6.9cm,trim={0cm 8cm 0cm 0cm},clip]{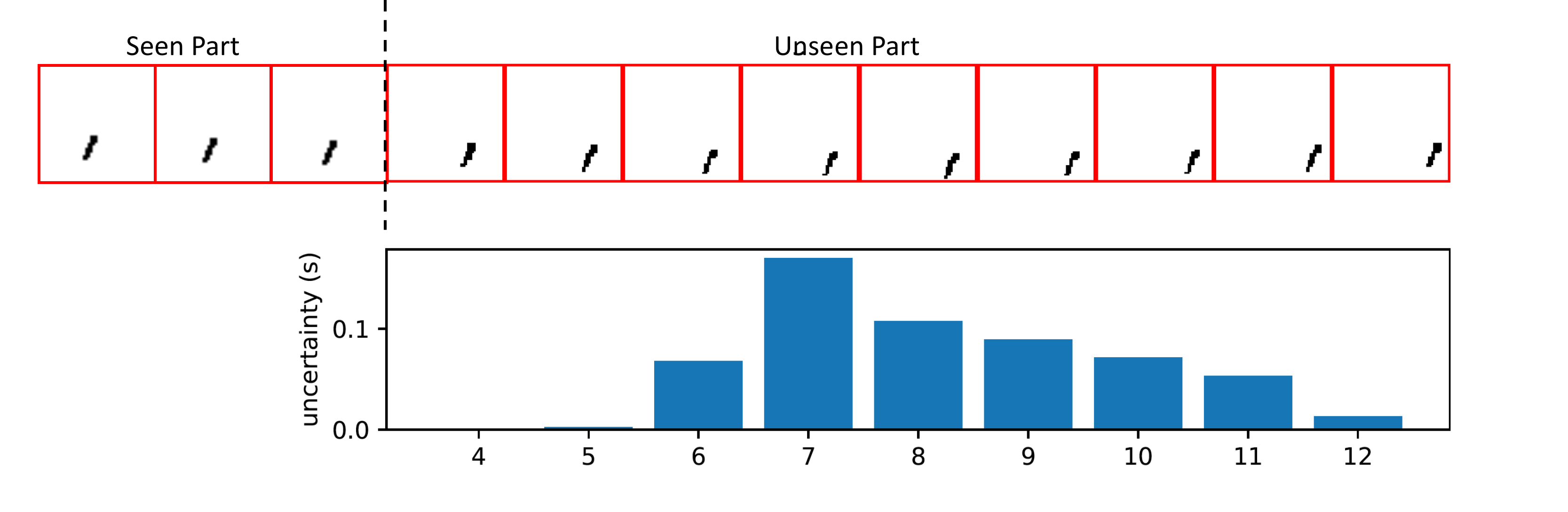} % [width=15cm,trim={0cm 8cm 0cm 0cm},clip]
    \includegraphics[width=6.9cm,trim={0cm 8cm 0cm 0cm},clip]{figs/supplementary_figs/prob_mnist_result2.pdf} % [width=14cm,trim={0cm 8cm 0cm 0cm},clip]
    \includegraphics[width=6.9cm,trim={0cm 6cm 0cm 0cm},clip]{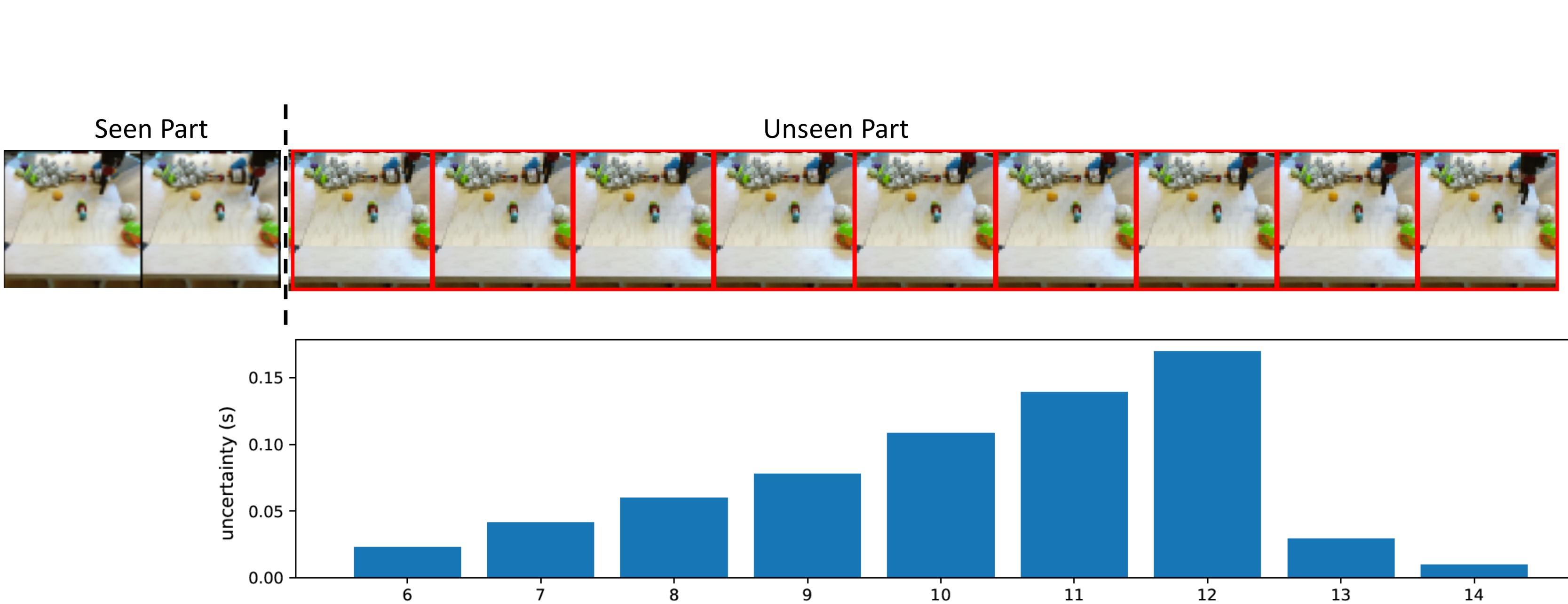} % [width=14cm,trim={0cm 6cm 0cm 0cm},clip]
    \includegraphics[width=6.9cm,trim={0cm 6cm 0cm 0cm},clip]{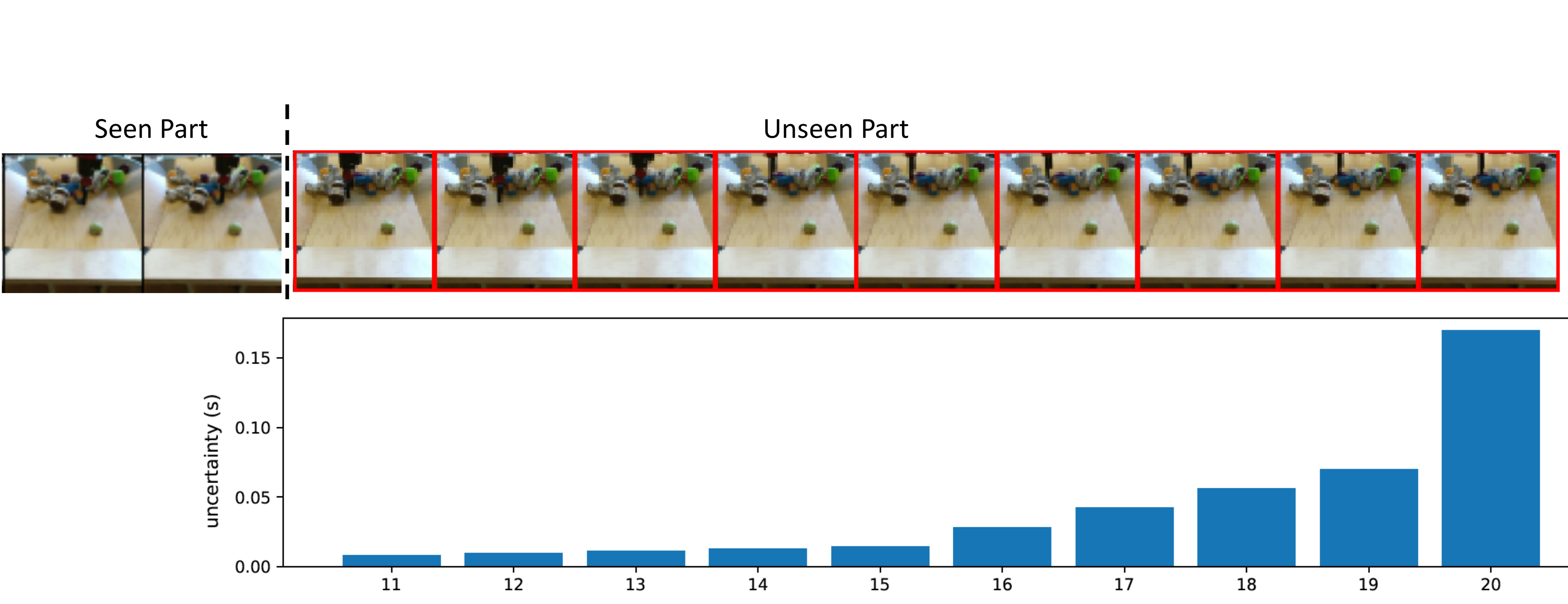} 
   \caption{Top two rows: Evolution of scaled uncertainty on the SMMNIST dataset (with NUQ trained with 2,000 training samples on the SMMNIST Dataset) against time-steps. The plot shows the increase in uncertainty co-occurs with the bounce of the digit against the boundary. Bottom two rows: Scaled Uncertainty against time-steps on BAIR Push (with NUQ trained with 2000 training samples on the BAIR Push Dataset), showing that uncertainty co-occurs with the occlusion of the robot arm.}
    \label{fig:smmnist_bair_prob}
    %\vspace*{-0.5cm}
\end{figure}
Figure~\ref{fig:smmnist_bair_prob} visualizes the scaled uncertainty values against the visual frames, across the SMMNIST and BAIR Push datasets, each trained with 2000 samples. See the caption of the figure for more details.

\section{Alternative Formulations}
\label{sec:alternatives}
Is NUQ the best formulation that one could have for quantifying uncertainty within a stochastic prediction model? In this section, we propose several alternatives and empirically evaluate them against the results we obtained using our formulation of NUQ, as an answer to this interesting research question.
\subsection{Alternatives}

\noindent\textbf{Alternative Priors:} In this variant, we replace our empirical gamma hyperprior, $\p(s_t)$, with a half-normal distribution with location set to $0$ and scale set to $1$, and in another variant we use the $\uniform(0,1)$ distribution as the hyperprior. 

\begin{figure}[th]
    \centering
    %\hspace{-2.5cm}
    \includegraphics[width=6.9cm]{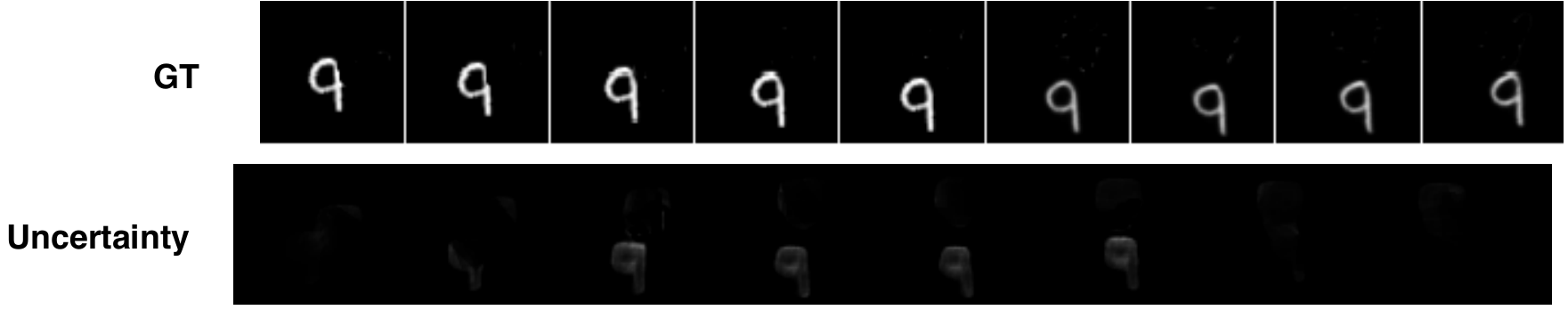} % 
   \caption{Visualization of pixel-wise uncertainty obtained by estimating the variance of the output, directly from the decoder for the SMMNIST Dataset, trained with 2,000 training samples.}
    \label{fig:smmnist_pix_unc}
    %\vspace*{-0.5cm}
\end{figure}

\noindent\textbf{Using Mahalanobis Distance:} In this variant, we use the frame decoder, $p_{\theta}(\cdot)$, to produce a diagonal covariance instead of producing the parameters of the gamma prior. Specifically, the output layer of the decoder now predicts both the future frame and the diagonal elements of $\mSigma_t^{\vz}$, where we assume $\mSigma_t^{\vz}$ is $n\times n$, decoded frames are size $d\times d$, and $D=d^2$. Generating an estimate of the output covariance matrix thus implies predicting the $D$ terms along the diagonal of this matrix. We then reshape these $D$ terms to $d\times d$ to match with the pixel resolution of the frames. We then use this uncertainty (precision) to weigh the MSE at a pixel level (instead of the precision $b_t$). This uncertainty is visualized for a sequence in Figure~\ref{fig:smmnist_pix_unc}.

\noindent\textbf{Directly Estimate precision from latent prior:} In this formulation, we repurpose the variance encoder $\zeta_{\lambda}$, to emit the variance to the MSE, $b_t$, directly. This is in contrast with the architecture of the variance encoder in NUQ, where it is used to estimate the sufficient statistics of the truncated normal distribution, $\alpha^{\ts}_t$ and $\beta^{\ts}_t$. We essentially replace the final hidden layer of the network with a single neuron with sigmoid activation in order to realize this setting.

\subsection{Alternatives -- Results}
In Table~\ref{tab:smmnist_hyperprior}, we provide comparisons of the above alternative formulations on the SMMNIST dataset, trained with 2,000 training samples. From the first row, we see that the $\text{Uniform}[0, 1]$ distribution variant under-performs compared to using the gamma distribution as a hyper-prior, as in NUQ. We surmise that this is due to these distributions being more spread out over the probability space, as a result of which they often sample $s_t$'s which do not match the true underlying distribution. This results in the MSE term in the loss function, being overly weighed when it should not  be and vice-versa.

Results for our other alternative, to compute the Mahalanobis-type precision matrix directly from the frame decoder, is provided in the second row in Table~\ref{tab:smmnist_hyperprior}; its performance is similar to the other alternatives. We also attempted to directly estimate $s_t$ from the variance of the latent space prior $\mSigma^{\vz}_t$. The results for this setting are shown in the third row in Table~\ref{tab:smmnist_hyperprior}. However, this setting performs poorly suggesting that a deterministic mapping of the $\mSigma_t^{\vz}$ to $s_t$ is not ideal, perhaps because the difference in the uncertainty distribution in the latent space and in the output is not accurately modeled this way.  Overall, the results in the table clearly show that our proposed formulation of the model yields the best empirical performance, nonetheless some other formulations to our model seem promising. % it weakens the MSE significantly

\begin{table*}[th]
%\scriptsize
\centering
\caption{SSIM, PSNR, and LPIPS scores on the SMMNIST test set after @1, @5, and @Convergence (C) (upto 150) epochs of training with alternative formulations of our model using 2,000 training samples. [Key: Best results in \textbf{bold}].}\label{tab:smmnist_hyperprior}
\begin{tabular}{l|c|c|c||c|c|c||c|c|c}
\hline \hline
\multirow{2}{*}{\textbf{Dataset: SMMNIST}} & \multicolumn{3}{c}{\textbf{SSIM $\uparrow$}} & \multicolumn{3}{c}{\textbf{PSNR $\uparrow$}} & \multicolumn{3}{c}{\textbf{LPIPS $\downarrow$}} \\ \cline{2-10}
& @1 & @5 &	@C & @1 & @5 &	@C & @1 & @5 &	@C \\ \hline
%\multicolumn{7}{c}{\textit{Number of training samples - 2000}} \\ \hline
$\p(s_t) \sim \text{Uniform}[0, 1]$  & 0.8173  &	0.8374  & 0.8523	 & 17.6	 & 17.95	 & 18.06 & 0.3442 & 0.3038 & 0.198	 \\
%%Estimate $s_t$ directly from prior  	& 0.7450 & 0.7454	  & 0.7848	 & 15.22	 & 15.23	 &	16.48 \\
Estimate $b_t$ from the decoder $p_{\theta}(\cdot)$  & {0.7627}	& 0.7628 &{0.7828} &	 {17.54} &	17.55 &	{17.55} & 0.3463 & 0.3259 & 0.2225 \\ 
Estimate $b_t$ w/o variance enc-dec  	& 0.7450 & 0.7454	  & 0.7648	 & 16.22	 & 16.53	 &	16.78 & 0.3469 & 0.3263 & 0.2328 \\ \hline
%NUP (ours) & \textbf{0.8706}	& \textbf{0.8932} &	\textbf{0.8948} &	\textbf{17.76} &	\textbf{18.13} &	\textbf{18.42}  \\ \hline
NUQ (Ours) & \textbf{0.8686}	& \textbf{0.8638} &	\textbf{0.8948} &	\textbf{17.76} &	\textbf{18.13} &	\textbf{18.14} & \textbf{0.3087} & \textbf{0.2836} & \textbf{ 0.1803} \\  \hline \hline
\hline
\end{tabular}
\end{table*}

\section{Architectural Details}
\label{sec:arch_det}
In this section, we elaborate on some of the architectural design choices that we made while implementing NUQ. Our primary objective while designing the architectural framework of NUQ was to ensure that our network's generation capacity remained similar to the state-of-the-art baselines, such as Denton and Fergus~\cite{denton2018stochastic}, such that all gains obtained by our framework, could be attributed to modeling the prediction uncertainty.

\subsection{Frame Encoder}
Our frame encoder consists of a hierarchical stack of 2d-convolution filters. For $48 \times 48$ inputs, we design a 4-layer network. The first layer consists of 64, $4 \times 4$ 2d-convolutional filters with stride 2 and padding 1, which are followed by 2d-BatchNorm and LeakyReLU non-linearity. In every subsequent layers, we keep doubling the number of filters. For $64 \times 64$ inputs, we adapt this network to make it a 5-layer one.

\subsection{LSTMs}
All LSTM modules in our framework, including the sequence discriminator, have a single hidden layer with 256-d hidden states, except for the LSTM in the frame decoder $p_{\theta}(\cdot)$, which has 2 hidden layers, each of 256-d.

\subsection{Frame Decoder}
We design the frame decoder in congruence with the frame encoder, so as to permit skip connections between them, in a U-Net style network~\cite{ronneberger2015u}. Therefore, our frame decoder obeys a similar architecture akin to the frame encoder, except the 2d-convolution filters are now replaced with 2d-deconvolution filters and the number of filters in each layer is doubled (in order to accommodate the skip connection).

\subsection{Variance Encoder}
Our variance encoder, $\zeta_{\lambda}(\cdot)$, is a 2-layer multi-layer perceptron, with LeakyReLU activations, which ultimately produces the sufficient statistics of the truncated normal distribution governing the posterior in the latent space.

\section{Derivations}
\label{sec:derivation}
In this section, we present the derivations of Eq. 9 and Eq. 11 in the paper.
We derive Eq. 9, along the lines of the variational lower bound derivation in Kingma and Welling~\cite{kingma2013auto}:
\begin{equation}
\small
\begin{multlined}
    \log \p(\vx_t|b_t, \vx_{1:t-1}) = \log \p(\vx_t|b_t, \vx_{1:t-1}) . \int_{\vz_{t}} \q_{\phi}(\vz_{t}|\vx_{1:t})  d\vz_t \\
    = \int_{\vz_{t}} \q_{\phi}(\vz_{t}|\vx_{1:t}) \log \frac{\p(\vx_t, \vz_{t}|b_t, \vx_{1:t-1})}{\p(\vz_{t}|\vx_{1:t})} d\vz_t \\
    = \int_{\vz_{t}} \q_{\phi}(\vz_{t}|\vx_{1:t}) \log \frac{\p(\vx_t, \vz_{t}|b_t, \vx_{1:t-1})}{\q_{\phi}(\vz_{t}|\vx_{1:t})} d\vz_t \\
    + \int_{\vz_{t}} \q_{\phi}(\vz_{t}|\vx_{1:t}) \log \frac{\q_{\phi}(\vz_{t}|\vx_{1:t})}{\p(\vz_{t}|\vx_{1:t})} d\vz_t \\
\end{multlined}
\normalsize
\end{equation}

The second term in the above equation is essentially a KL-Divergence, which is non-negative. We therefore have: 
\begin{equation}
\small
\begin{multlined}
    \log \p(\vx_t|b_t, \vx_{1:t-1}) \geq \int_{\vz_{t}} \q_{\phi}(\vz_{t}|\vx_{1:t}) \log \frac{\p(\vx_t, \vz_{t}|b_t, \vx_{1:t-1})}{\q_{\phi}(\vz_{t}|\vx_{1:t})} d\vz_t \\
    = \int_{\vz_{t}} \q_{\phi}(\vz_{t}|\vx_{1:t}) \log \frac{\p(\vx_t | \vx_{1:t-1}, \vz_{t}, b_t) \p(\vz_{t}|\vx_{1:t-1})}{\q_{\phi}(\vz_{t}|\vx_{1:t})} d\vz_t 
\end{multlined}
\normalsize
\end{equation}
This yields Eq. 9, when the expression inside the log is split into two, with the first term amounting to the expectation term in Eq. 9, while the second one resulting in the KL-term.

Our NUQ framework is a essentially, a hierarchical variational encoder-decoder network, where the second level of the hierarchy is described by Eq. 11. Derivation for Eq. 11, thus proceeds analogously to Eq. 9, as follows:
\begin{equation}
\small
\begin{multlined}
    \log \p(b_t| \vx_{1:t-1}) = \log \p(b_t | \vx_{1:t-1}) . \int_{s_{t}} \q_{\lambda}(s_{t}|\vx_{1:t-1})  ds_t \\
    = \int_{s_{t}} \q_{\lambda}(s_{t}|\vx_{1:t-1}) \log \frac{\p(b_t, s_{t}| \vx_{1:t-1})}{\p(s_{t}|b_t, \vx_{1:t-1})} ds_t \\
    = \int_{s_{t}} \q_{\phi}(s_{t}|\vx_{1:t-1}) \log \frac{\p(b_t, s_{t}| \vx_{1:t-1})}{\q_{\lambda}(s_{t}|\vx_{1:t-1})} ds_t \\
    + \int_{s_{t}} \q_{\lambda}(s_{t}|\vx_{1:t-1}) \log \frac{\q_{\lambda}(s_{t}|\vx_{1:t-1})}{\p(s_{t}|b_t, \vx_{1:t-1})} ds_t \\
\end{multlined}
\normalsize
\end{equation}

Like before, the second term in the above equation is essentially a KL-Divergence, which is non-negative. We therefore have: 
\begin{equation}
\small
\begin{multlined}
    \log \p(b_t| \vx_{1:t-1}) \geq \int_{s_{t}} \q_{\lambda}(s_{t}|\vx_{1:t-1}) \log \frac{\p(b_t, s_{t}|\vx_{1:t-1})}{\q_{\lambda}(s_{t}|\vx_{1:t-1})} ds_t \\
    = \int_{s_{t}} \q_{\lambda}(s_{t}|\vx_{1:t-1}) \log \frac{\p(b_t | \vx_{1:t-1}, s_{t}) \p(s_{t})}{\q_{\lambda}(s_{t}|\vx_{1:t-1})} ds_t 
\end{multlined}
\normalsize
\end{equation}
When the expression inside the log is split into two, the first term results in the expectation term in Eq. 11, while the second one amounts to the KL-term.

In this section, we present model performances of NUQ versus competing baselines on the UCF-101 dataset~\cite{ucf2012doc} -- a dataset of videos, resized to $64 \times 64$ containing 101 common human action categories (such as pushups, cricket shot, etc.), spanning both indoor and outdoor activities. The test set for this dataset consists of 1,895 videos. In order to conduct experiments in this setting, we train all models by showing them 5 context frames and task them to predict the next 15. The results showcase the dominance of NUQ over competing methods on this challenging dataset as well.

\section{Quantitative Evaluation of Diversity} %{Diversity of Generated Samples}
\label{sec:diversity}
\begin{figure}[th]
   \centering
    \subfigure[SMMNIST - SSIM]{\label{fig:smmnist_ssim_div}\includegraphics[width=6.8cm,trim={0cm 0cm 4cm 0cm},clip]{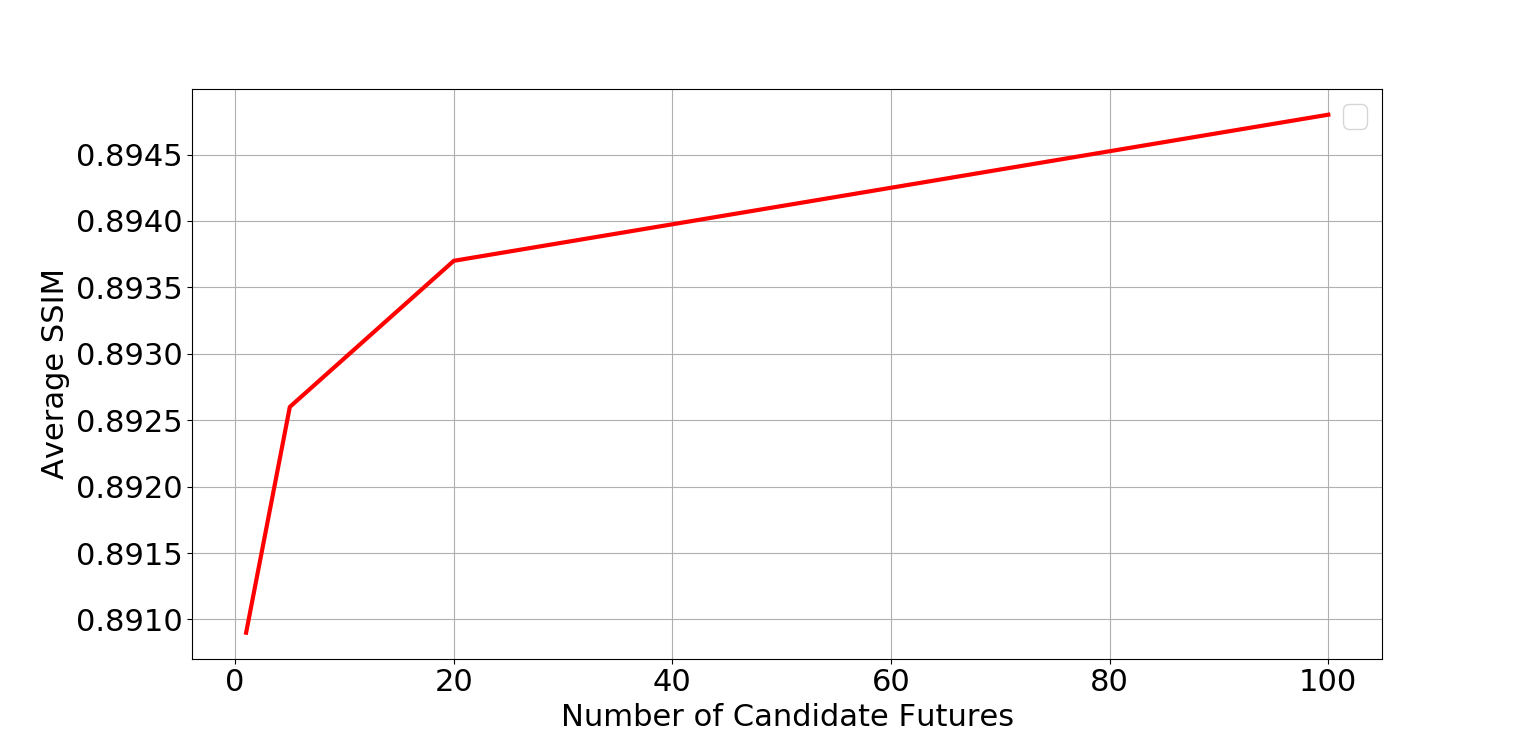}  }
    \subfigure[BAIR Push - SSIM]{\label{fig:bair_ssim_div}\includegraphics[width=6.8cm,trim={0cm 0cm 4cm 0cm},clip]{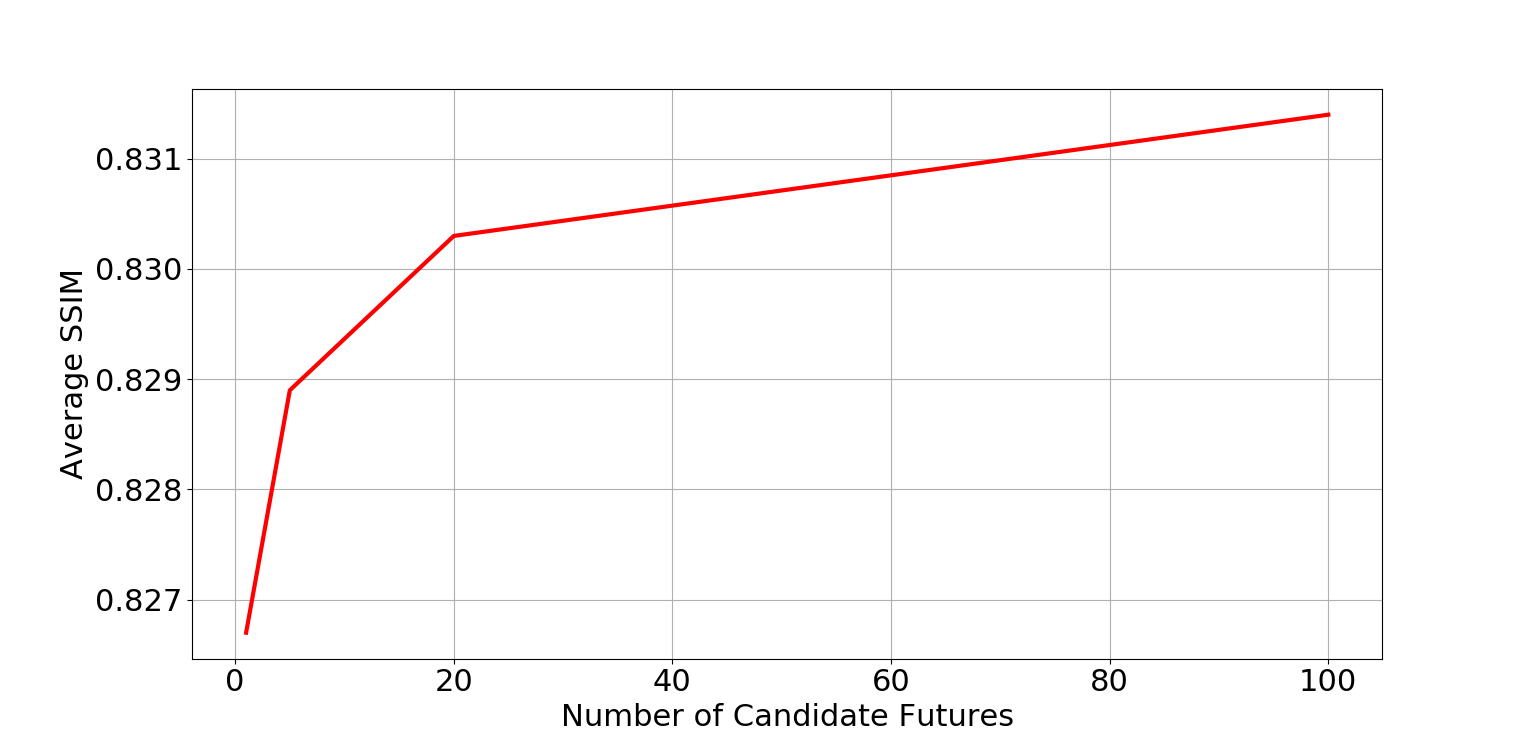}  }
    \caption{Diversity in Generated Futures: Evaluation of diversity in the generation using SSIM on: (a) SMMNIST, (b) BAIR-Push for increasing number of candidate futures, computed by comparing against the ground truth (higher the better). We used 2000 samples for training NUQ for both datasets.}
    %\label{fig:smmnist_ssim_div}
    %\vspace{-0.2cm}
\end{figure}

\begin{figure}[th]
   \centering
     \subfigure[SMMNIST-Intra]{\label{fig:smmnist_ssim_div_intra-a}\includegraphics[width=6.8cm,trim={0cm 0cm 4cm 0cm},clip]{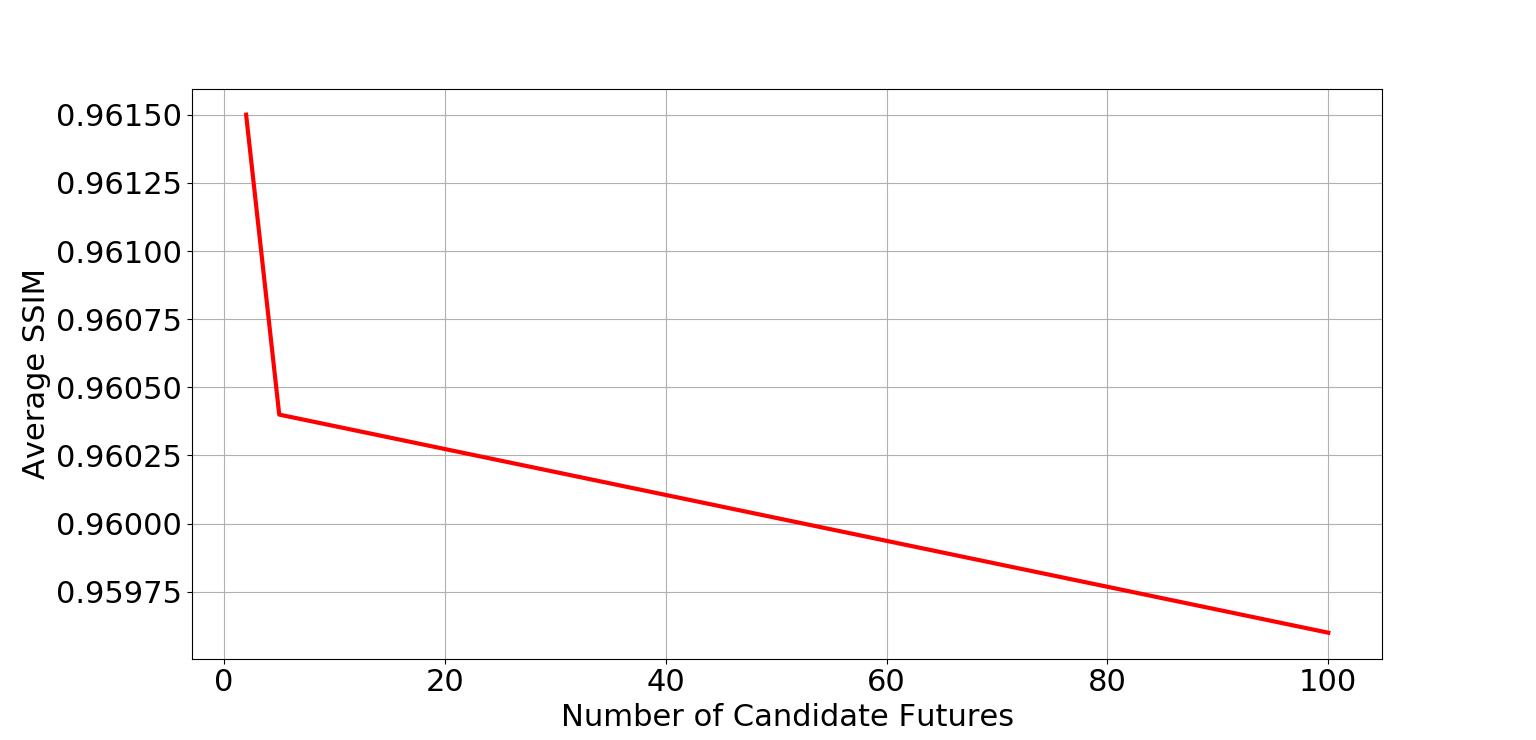} } 
      %\subfigure[BAIR-Push]{\label{fig:smmnist_ssim_div_intra-b}\includegraphics[width=6.8cm,trim={0cm 0cm 4cm 0cm},clip]{figs/supplementary_figs/Diversity/IntraSamp_PSNR.png}  }
      \subfigure[BAIR-Intra]{\label{fig:smmnist_ssim_div_intra-b}\includegraphics[width=6.8cm,trim={0cm 0cm 4cm 0cm},clip]{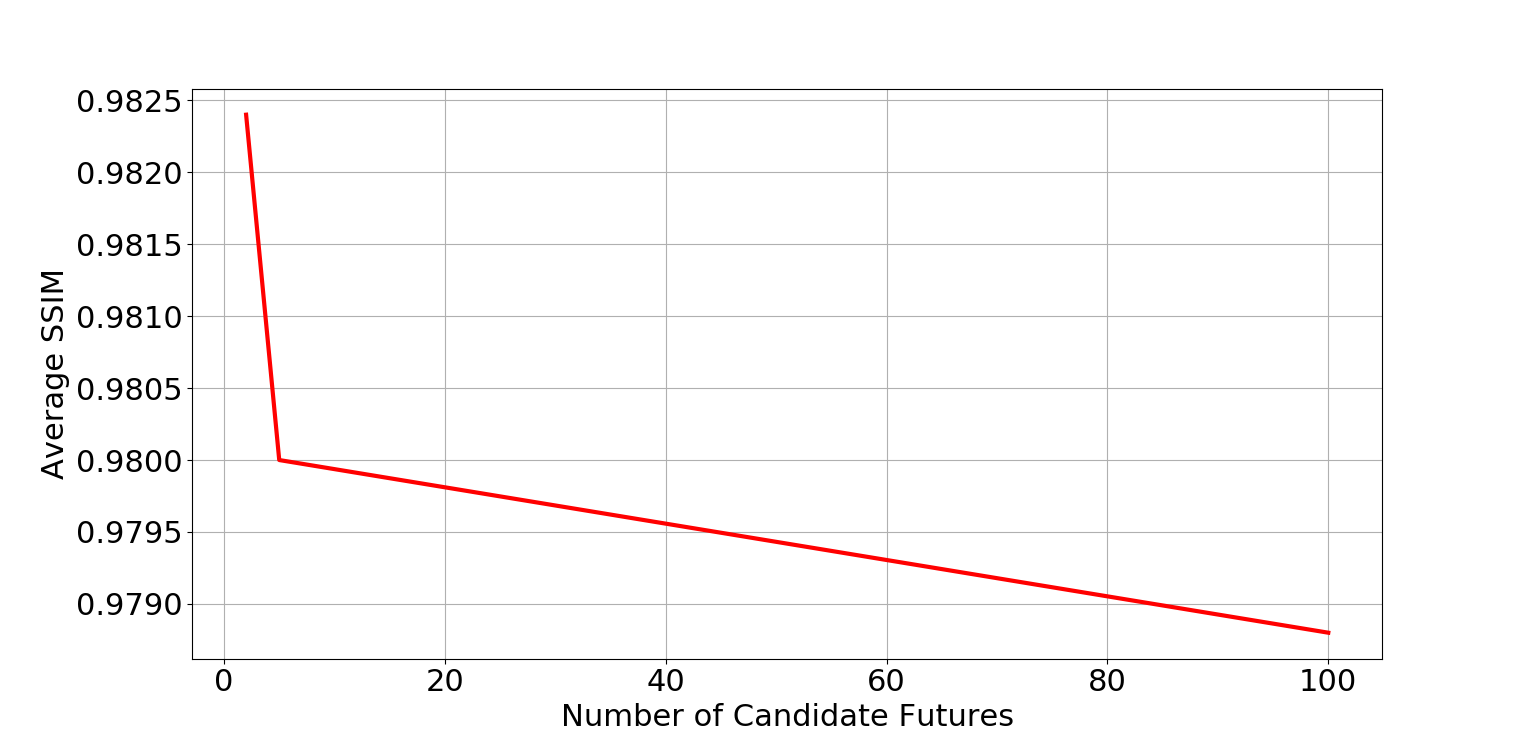} } % trim={1cm 0cm 1cm 0cm}
  \caption{Diversity in Generated Futures: Evaluation of diversity in the generation: (a,b) shows diversity in the generated futures by comparing intra-SSIM distances between all the futures, at a given time step, and computing the average (lower the better), for SMMNIST and BAIR Push respectively. For each of these datasets NUQ was trained with 2000 samples.}
    %\label{fig:smmnist_ssim_div}
    \vspace*{-0.2cm}
\end{figure}

In order to analyze the extent of diversity in the generated frames of our model, we first resort to quantitative evaluation. In Figures~\ref{fig:smmnist_ssim_div},~\ref{fig:bair_ssim_div}, we plot the average SSIM scores (over time steps) against the number of  generated future candidates per time-step for each of the three datasets. For purposes of these plots, the SSIM is computed between the generated samples and the ground-truth. The monotonically increasing curve, in these figures, suggests straightforwardly, that sampling more future per time step helps in better generation, resulting from the synthesis of more accurate samples - indicating the model's diversity. In Figures~\ref{fig:smmnist_ssim_div_intra-a},~\ref{fig:smmnist_ssim_div_intra-b}, we plot the average SSIM and PSNR scores between every pair of candidates generated in each time step, against the number of futures. These plots decrease monotonically, suggesting greater difference (i.e. diversity) between the generated frames as the number of sampled futures goes up. See the figure caption for more details.

%~\ref{fig:bair_comp_13}, ,~\ref{fig:bair_comp_90}

 \section{Qualitative Results}
 \begin{figure}[th]
    %\centering
    %\hspace{-2.5cm}
    \includegraphics[width=8.5cm]{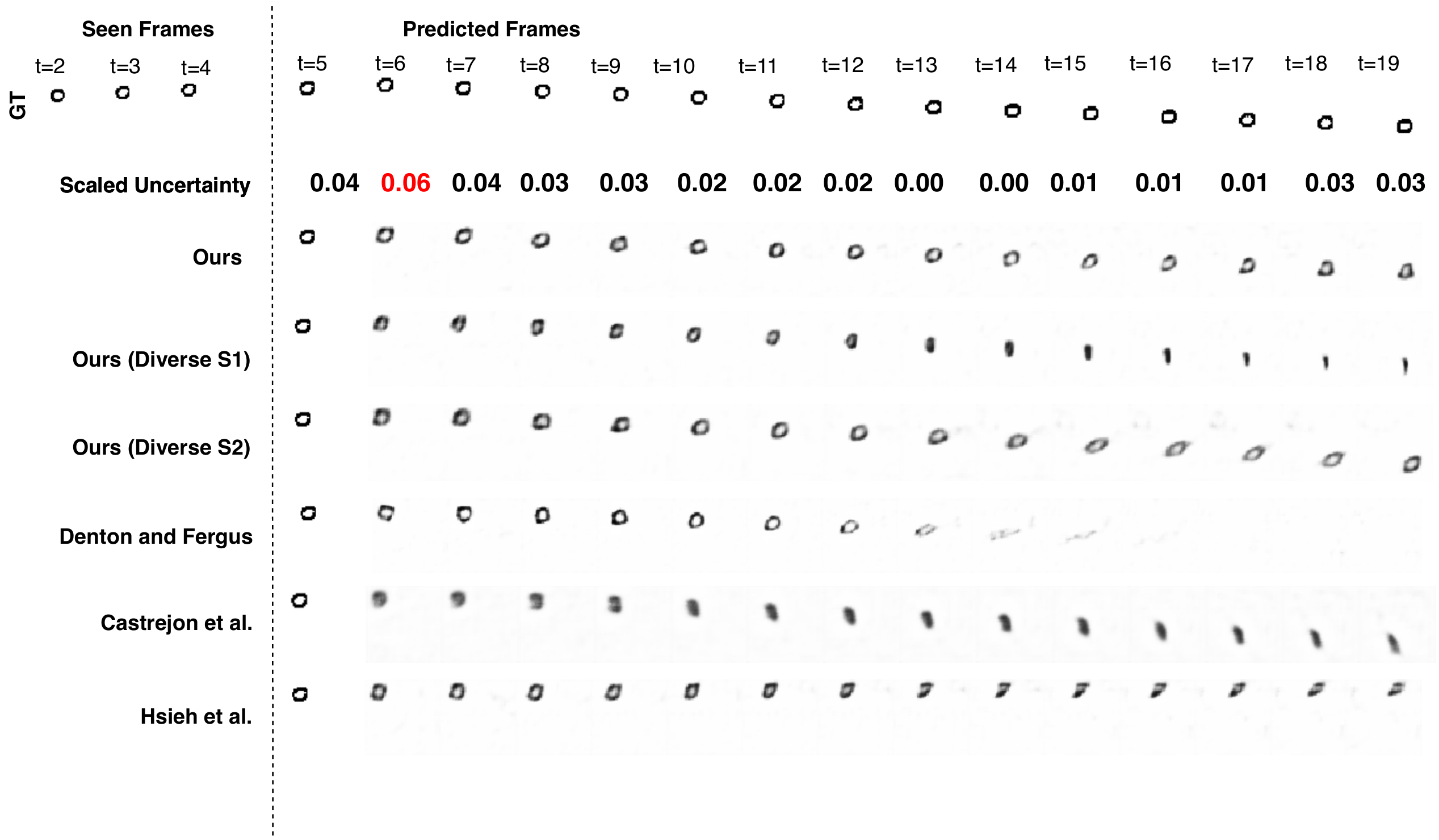} %  % [scale=0.7]
   \caption{Visualization of generations by our method versus competing baselines on the SMMNIST Dataset, trained with 2,000 training samples. Further, diverse generations by our method are also shown. Note scaled uncertainty higher than 0.05 is shown in red.}
    \label{fig:smmnist_comp_317}
    %\vspace*{-0.5cm}
\end{figure}

\begin{figure}[th]
    \centering
    \includegraphics[width=8.5cm]{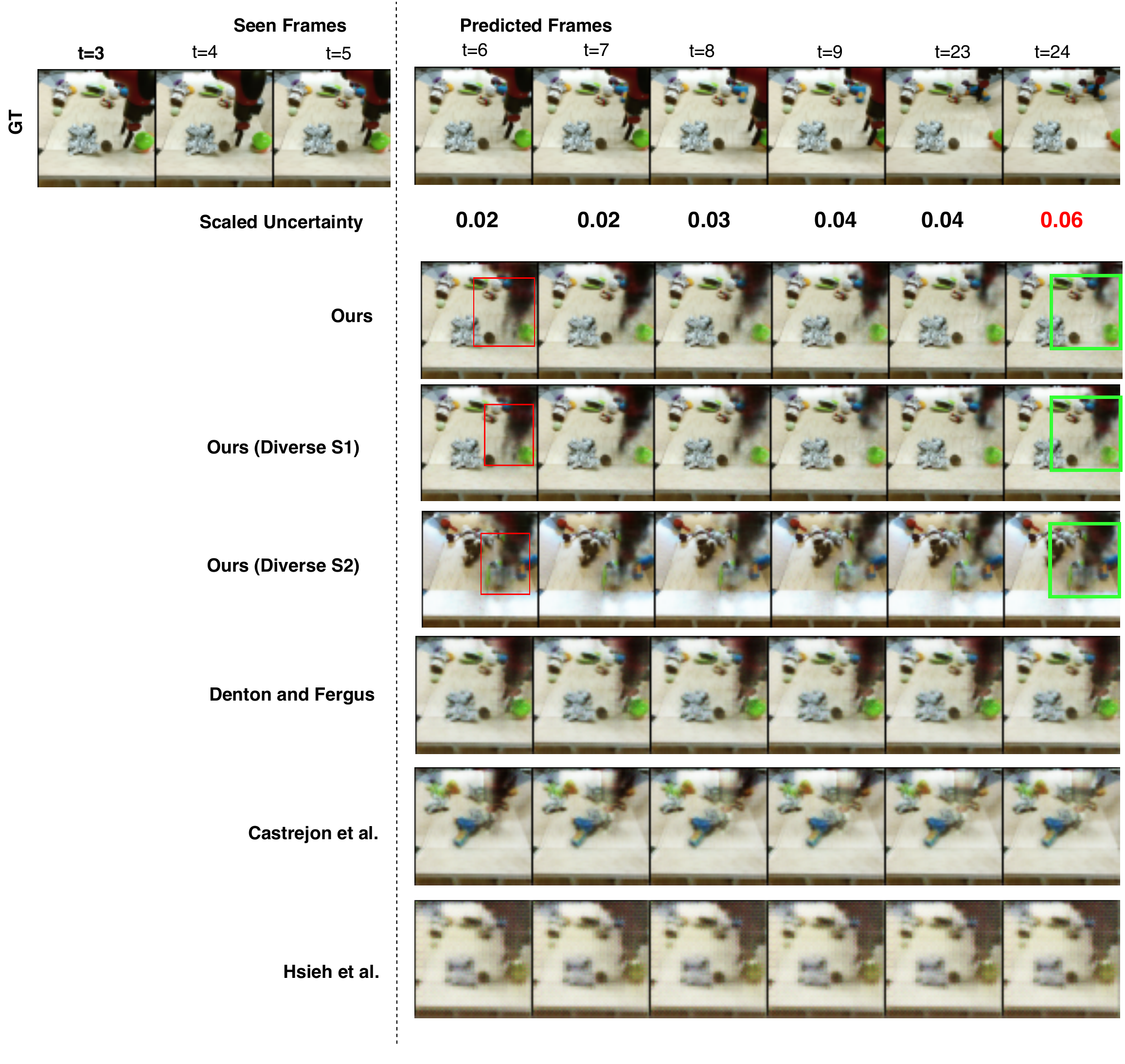} %  scale=0.75
   \caption{Visualization of generations by our method versus competing baselines on the BAIR Robot Push Dataset, trained with 2,000 training samples.  Further, diverse generations by our method are also shown. High motion regions are indicated by a red bounding box, while spatial regions exhibiting high diversity are shown by a green bounding box. Note scaled uncertainty higher than 0.05 is shown in red.}
    \label{fig:bair_comp_188}
    %\vspace*{-0.5cm}
\end{figure}

\begin{figure}[th]
    \centering
    \includegraphics[width=8.5cm]{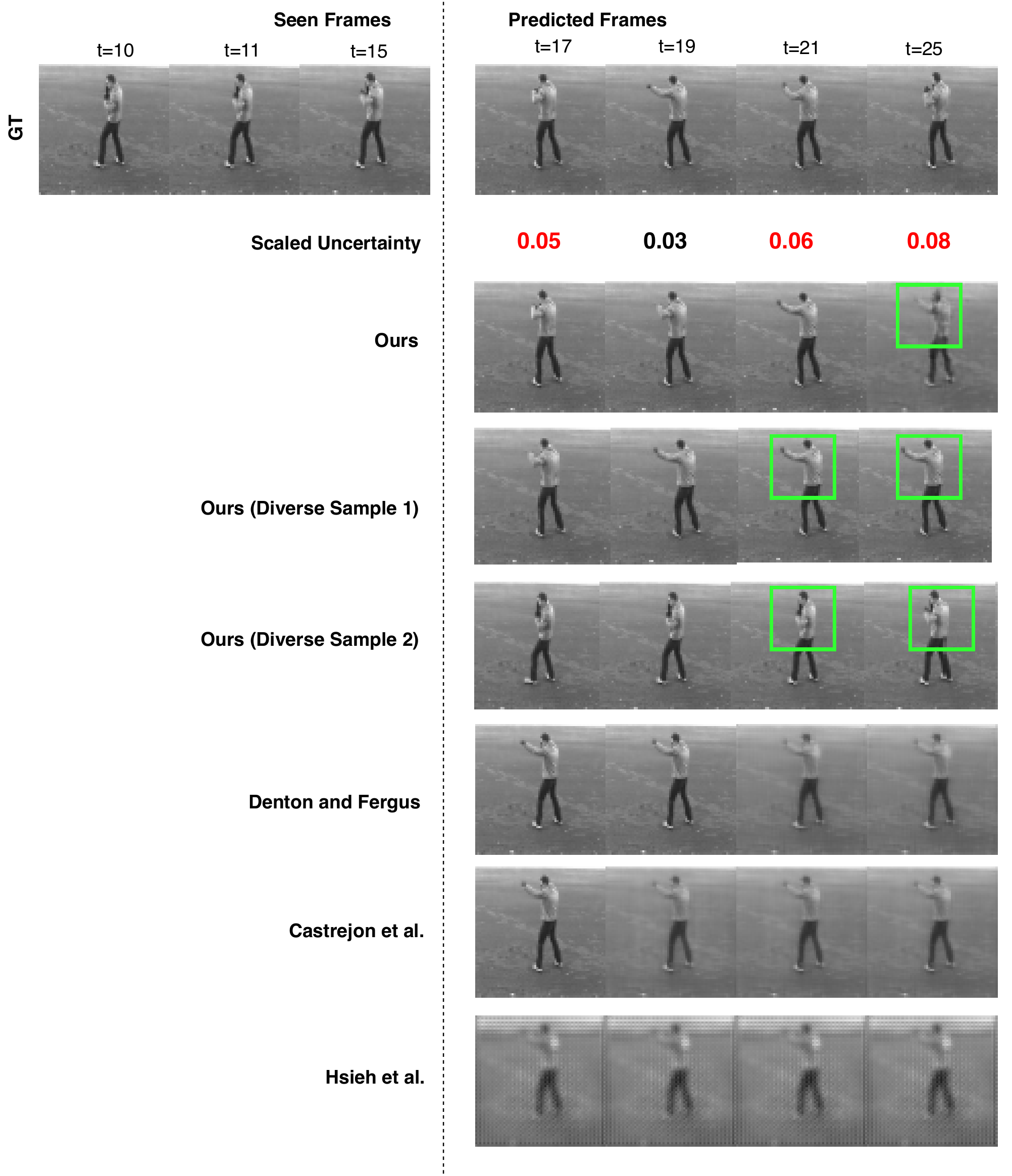} %  scale=0.8
   \caption{Visualization of generations by our method versus competing baselines on the KTH Action Dataset, trained with the full training data of 1,911 training samples.  Further, diverse generations by our method are also shown. Spatial regions exhibiting high diversity are shown by a green bounding box. Note scaled uncertainty higher than 0.05 is shown in red.}
    \label{fig:kth_comp_99}
    %\vspace*{-0.5cm}
\end{figure}
 
 We next present visualizations of frames generated using NUQ vis-\'a-vis competing baselines, on the SMMNIST, BAIR push, UCF-101, and KTH Action datasets. Also shown are diverse frame generations by NUQ for each of these datasets.

\begin{figure*}[th]
    %\centering
    %\hspace{-2.5cm}
    \includegraphics[scale=0.6]{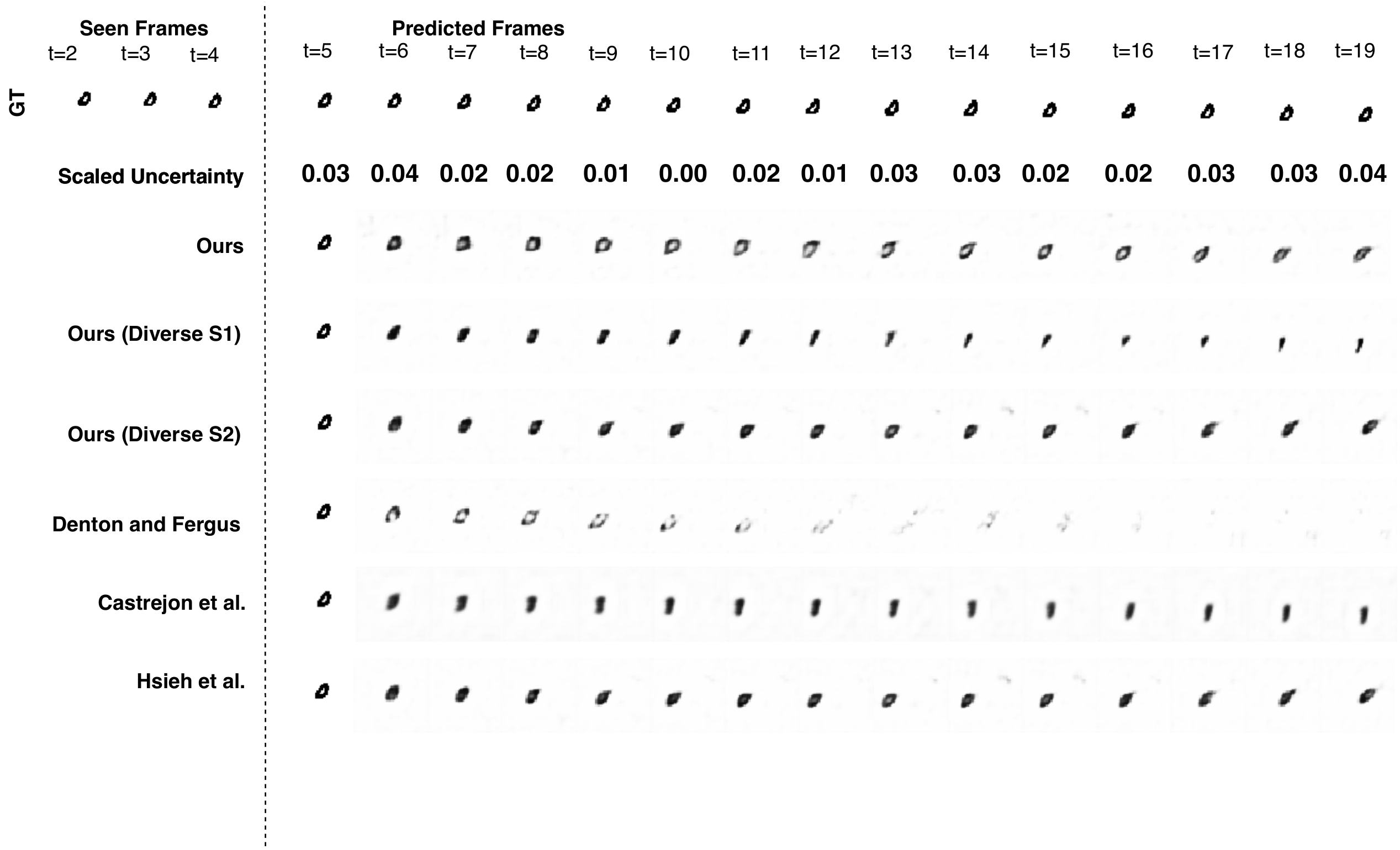} % width=6.9cm
   \caption{Visualization of generations by our method versus competing baselines on the SMMNIST Dataset, trained with 2,000 training samples. Further, diverse generations by our method are also shown. Note scaled uncertainty higher than 0.05 is shown in red.}
    \label{fig:smmnist_comp_157}
    %\vspace*{-0.5cm}
\end{figure*}

\begin{figure*}[th]
    %\centering
    %\hspace{-2.5cm}
    \includegraphics[scale=0.6]{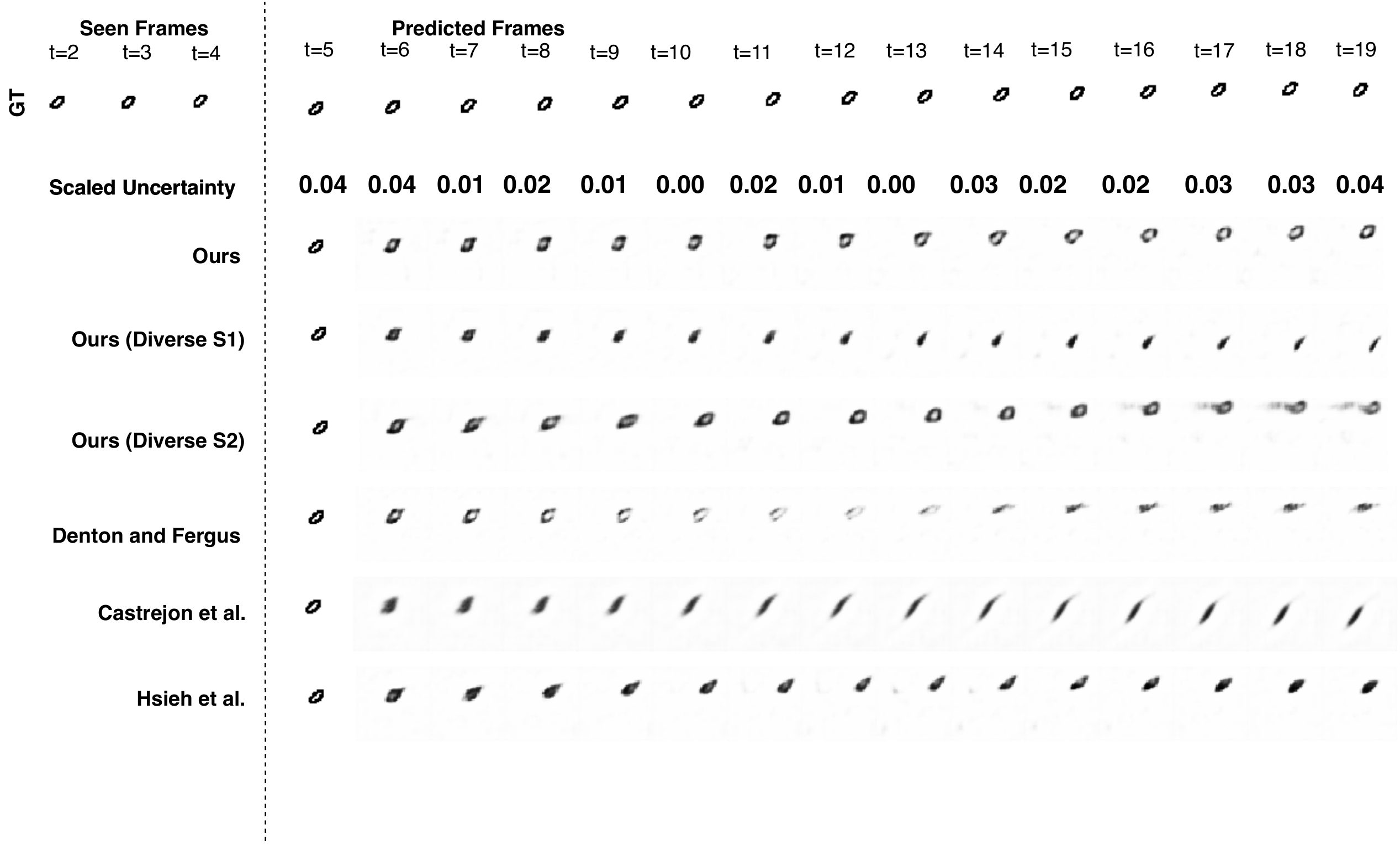} % width=6.9cm
   \caption{Visualization of generations by our method versus competing baselines on the SMMNIST Dataset, trained with 2,000 training samples. Further, diverse generations by our method are also shown. Note scaled uncertainty higher than 0.05 is shown in red.}
    \label{fig:smmnist_comp_180}
    %\vspace*{-0.5cm}
\end{figure*}

\begin{figure*}[th]
    %\centering
    %\hspace{-2.5cm}
    \includegraphics[scale=0.6]{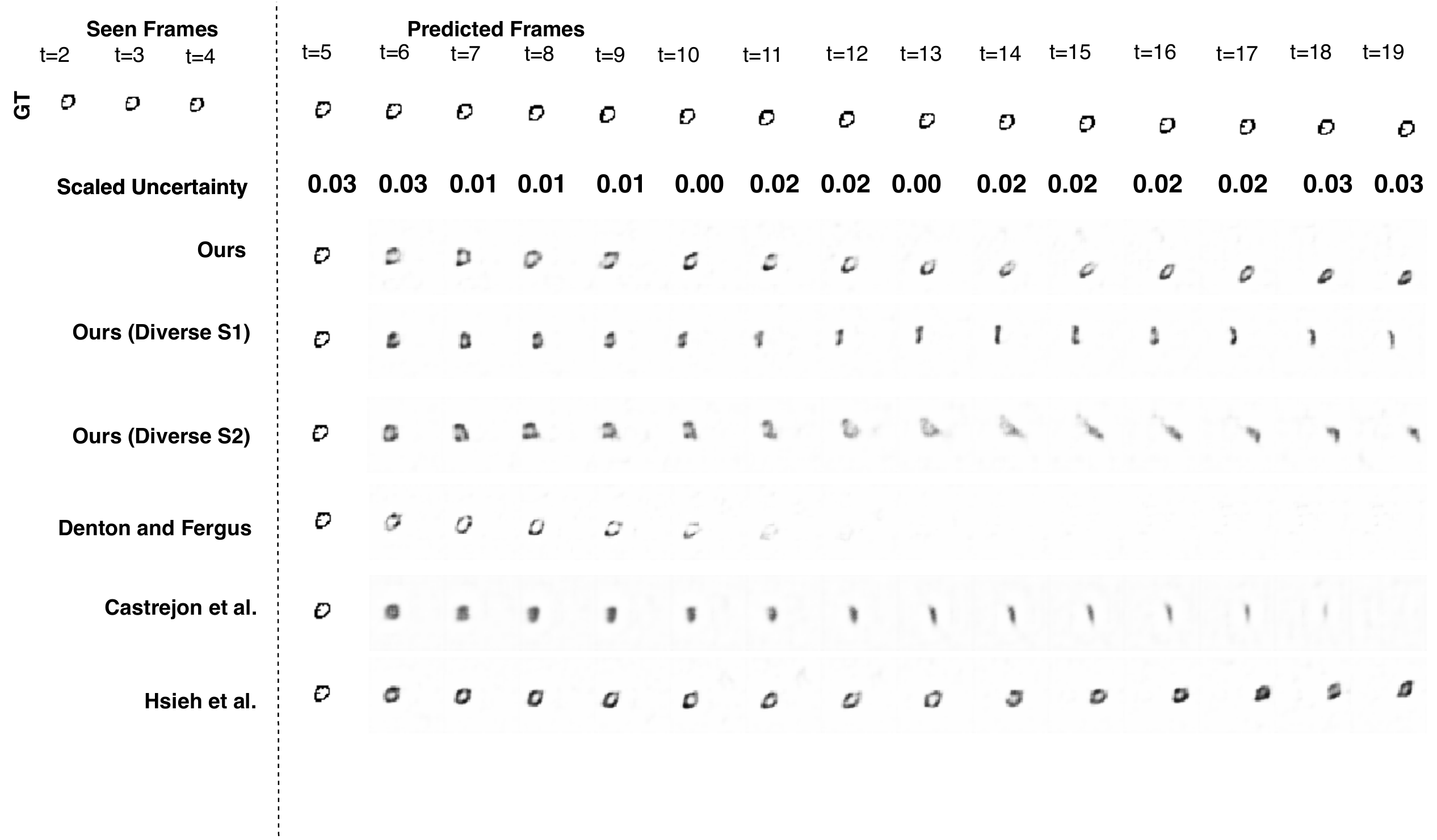} % width=6.9cm
   \caption{Visualization of generations by our method versus competing baselines on the SMMNIST Dataset, trained with 2,000 training samples. Further, diverse generations by our method are also shown. Note scaled uncertainty higher than 0.05 is shown in red.}
    \label{fig:smmnist_comp_295}
    %\vspace*{-0.5cm}
\end{figure*}

\begin{figure*}[th]
    %\centering
    %\hspace{-2.5cm}
    \includegraphics[scale=0.6]{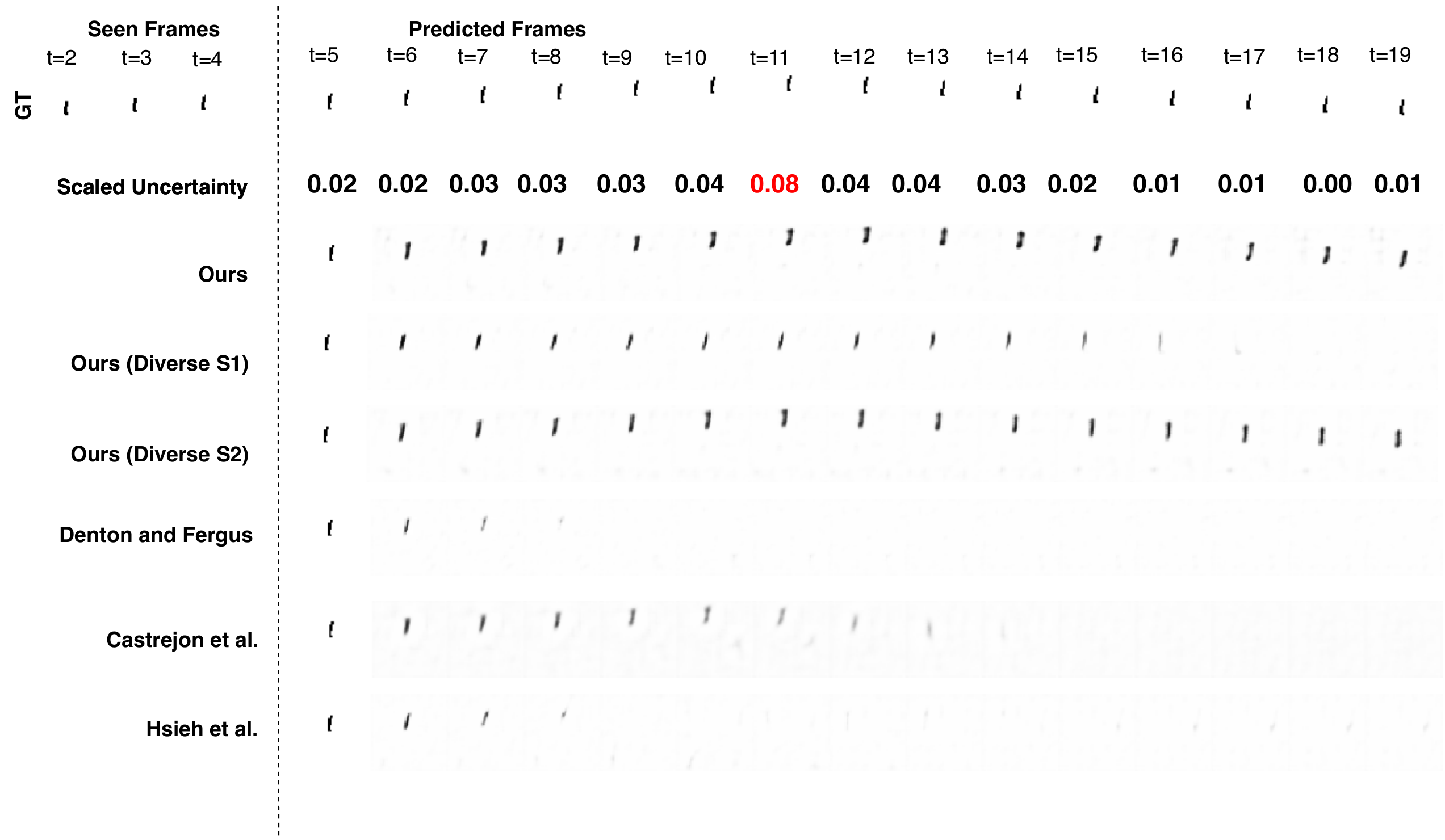} % width=6.9cm
   \caption{Visualization of generations by our method versus competing baselines on the SMMNIST Dataset, trained with 2,000 training samples. Further, diverse generations by our method are also shown. Note scaled uncertainty higher than 0.05 is shown in red.}
    \label{fig:smmnist_comp_366}
    %\vspace*{-0.5cm}
\end{figure*}

\begin{figure*}[th]
    %\centering
    %\hspace{-2.5cm}
    \includegraphics[scale=0.7]{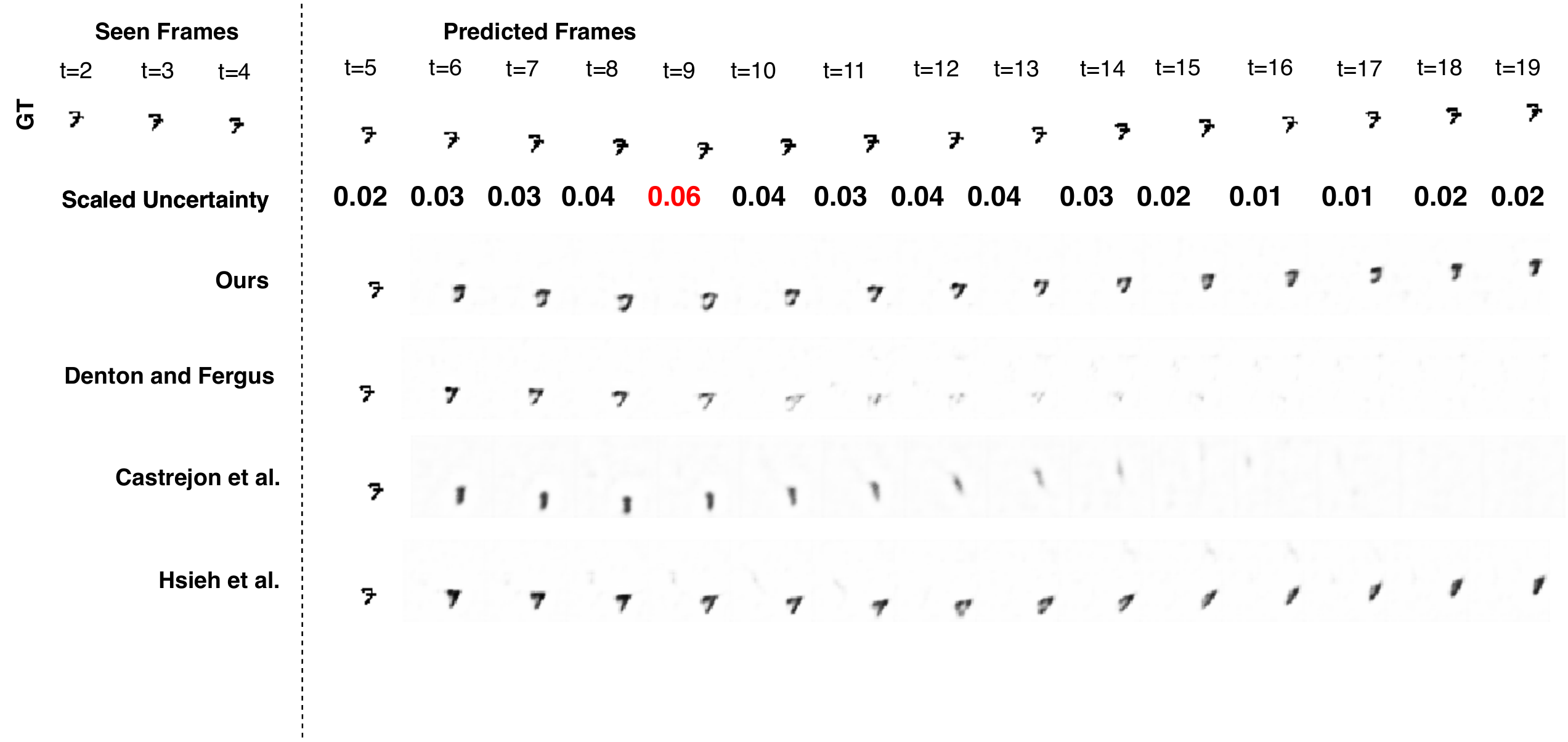} % width=6.9cm
   \caption{Visualization of generations by our method versus competing baselines on the SMMNIST Dataset, trained with 2,000 training samples. Note scaled uncertainty higher than 0.05 is shown in red.}
    \label{fig:smmnist_comp_399}
    %\vspace*{-0.5cm}
\end{figure*}

\begin{figure*}[th]
    %\centering
    %\hspace{-2.5cm}
    \includegraphics[scale=0.7]{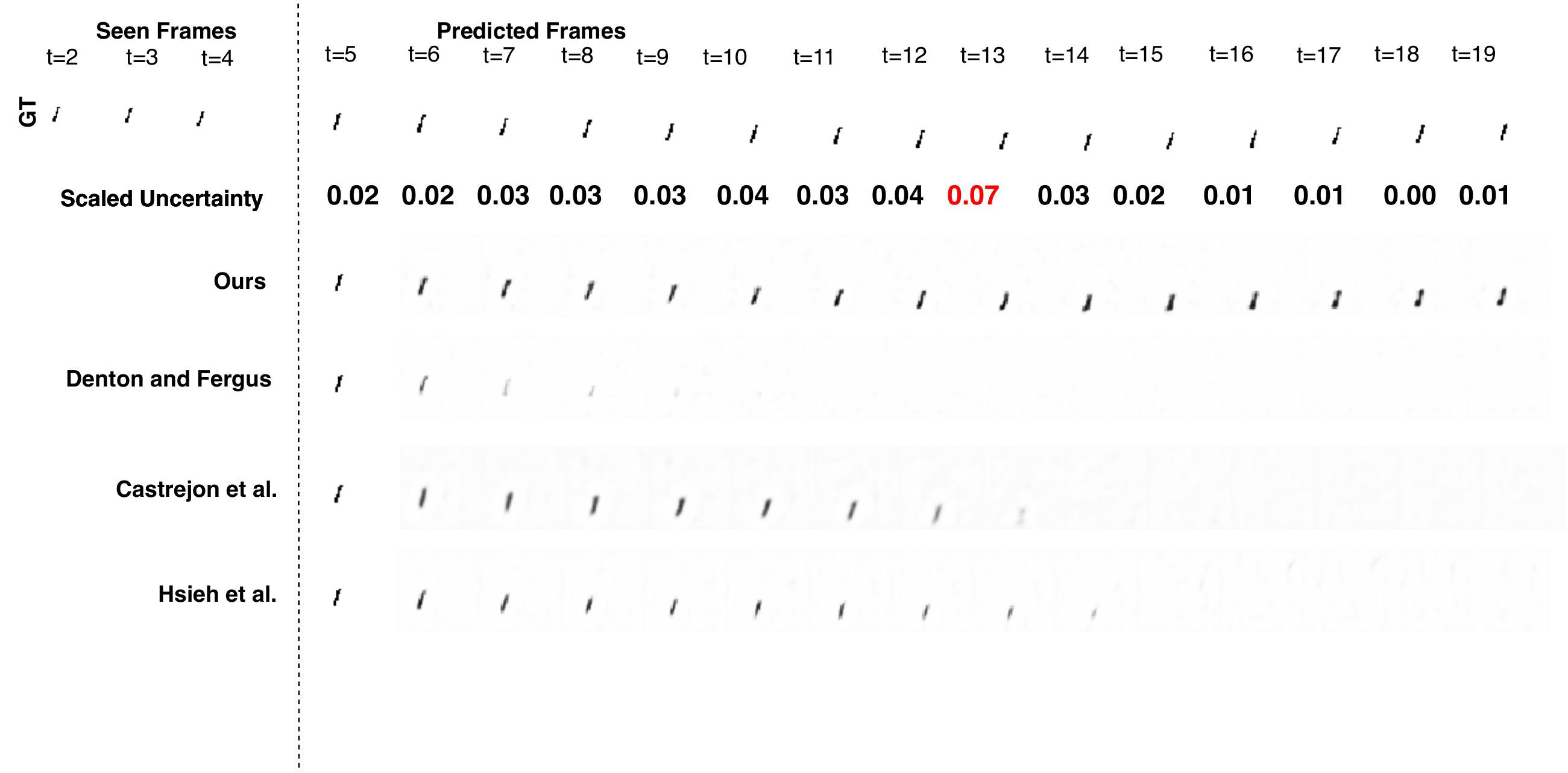} % width=6.9cm
   \caption{Visualization of generations by our method versus competing baselines on the SMMNIST Dataset, trained with 2,000 training samples. Note scaled uncertainty higher than 0.05 is shown in red.}
    \label{fig:smmnist_comp_449}
    %\vspace*{-0.5cm}
\end{figure*}

\begin{figure*}[th]
    %\centering
    %\hspace{-2.5cm}
    \includegraphics[scale=0.7]{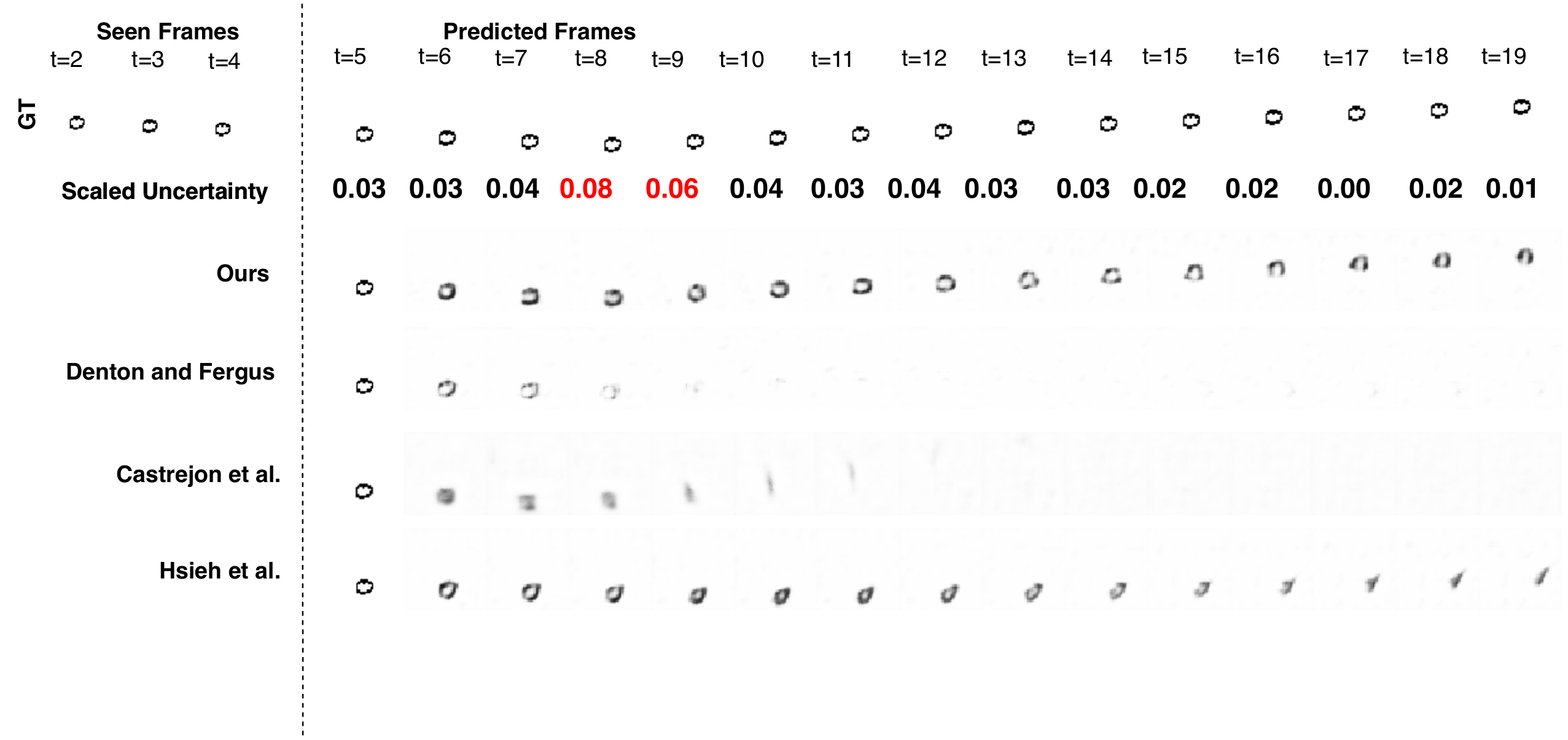} % width=6.9cm
   \caption{Visualization of generations by our method versus competing baselines on the SMMNIST Dataset, trained with 2,000 training samples. Note scaled uncertainty higher than 0.05 is shown in red.}
    \label{fig:smmnist_comp_482}
    %\vspace*{-0.5cm}
\end{figure*}

\begin{figure*}[th]
    %\centering
    %\hspace{-2.5cm}
    \includegraphics[scale=0.7]{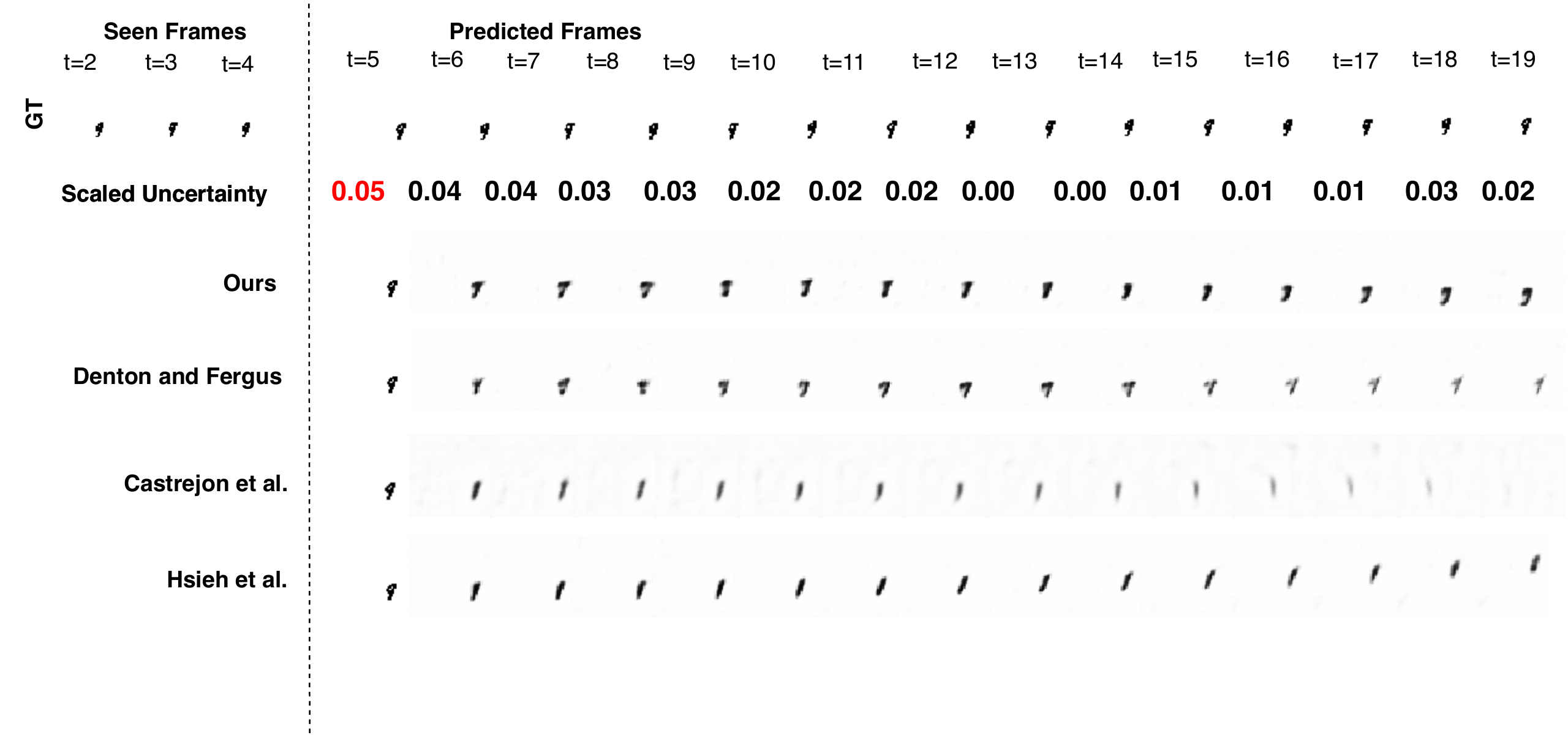} % width=6.9cm
   \caption{Visualization of generations by our method versus competing baselines on the SMMNIST Dataset, trained with 2,000 training samples. Note scaled uncertainty higher than 0.05 is shown in red.}
    \label{fig:smmnist_comp_500}
    %\vspace*{-0.5cm}
\end{figure*}

The results in Figures~\ref{fig:smmnist_comp_317},~\ref{fig:smmnist_comp_157},~\ref{fig:smmnist_comp_180},~\ref{fig:smmnist_comp_295},~\ref{fig:smmnist_comp_366},~\ref{fig:smmnist_comp_399},~\ref{fig:smmnist_comp_449},~\ref{fig:smmnist_comp_482},~\ref{fig:smmnist_comp_500} show qualitative generation results on the SMMNIST dataset, trained with 2000 samples. Besides the superior quality of the results generated by our method, we note that for some cases such as in Figures~\ref{fig:smmnist_comp_366},~\ref{fig:smmnist_comp_399},~\ref{fig:smmnist_comp_449} the prediction of the baseline method simply disappears. We surmise that this is due to their inability to learn the motion dynamics of the digit well, in uncertain environments. In particular, if the motion is not aptly learnt, then the model often gets penalized heavily for inaccurately placing a digit (via the MSE loss), since this results in a high pixel-wise error. In such a scenario, a model might prefer to not display the digit at all. However, high stochasticity in the data may not suit this well and as a result might hurt the generalization. By intelligently down-weighting the MSE, we circumvent this problem. 

\begin{figure*}[th]
    \centering
    \includegraphics[scale=0.75]{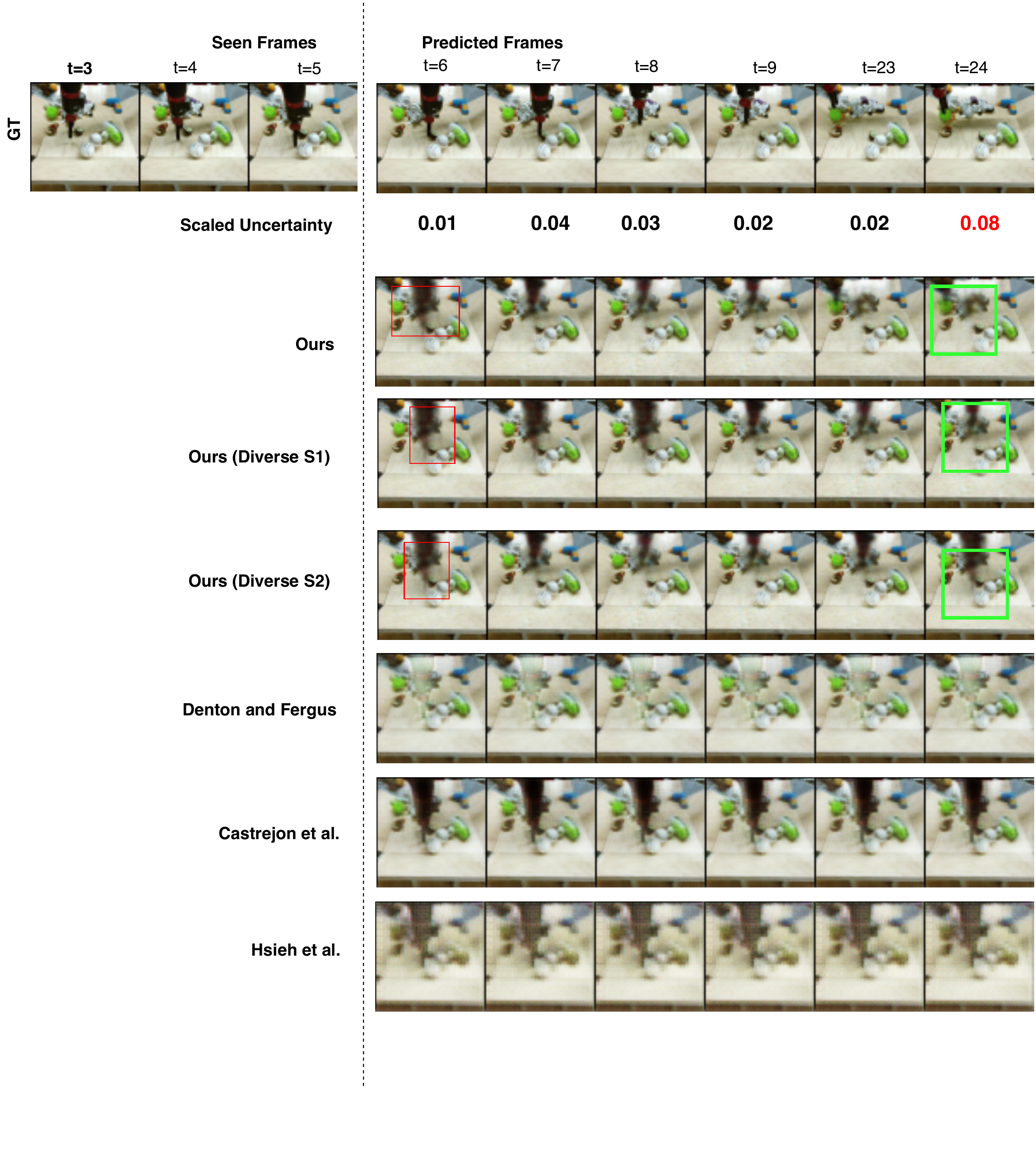} % width=6.9cm
   \caption{Visualization of generations by our method versus competing baselines on the BAIR Robot Push Dataset, trained with 2,000 training samples.  Further, diverse generations by our method are also shown. High motion regions are indicated by a red bounding box, while spatial regions exhibiting high diversity are shown by a green bounding box. Note scaled uncertainty higher than 0.05 is shown in red.}
    \label{fig:bair_comp_48}
    %\vspace*{-0.5cm}
\end{figure*}

\begin{figure*}[th]
    \centering
    \includegraphics[scale=0.75]{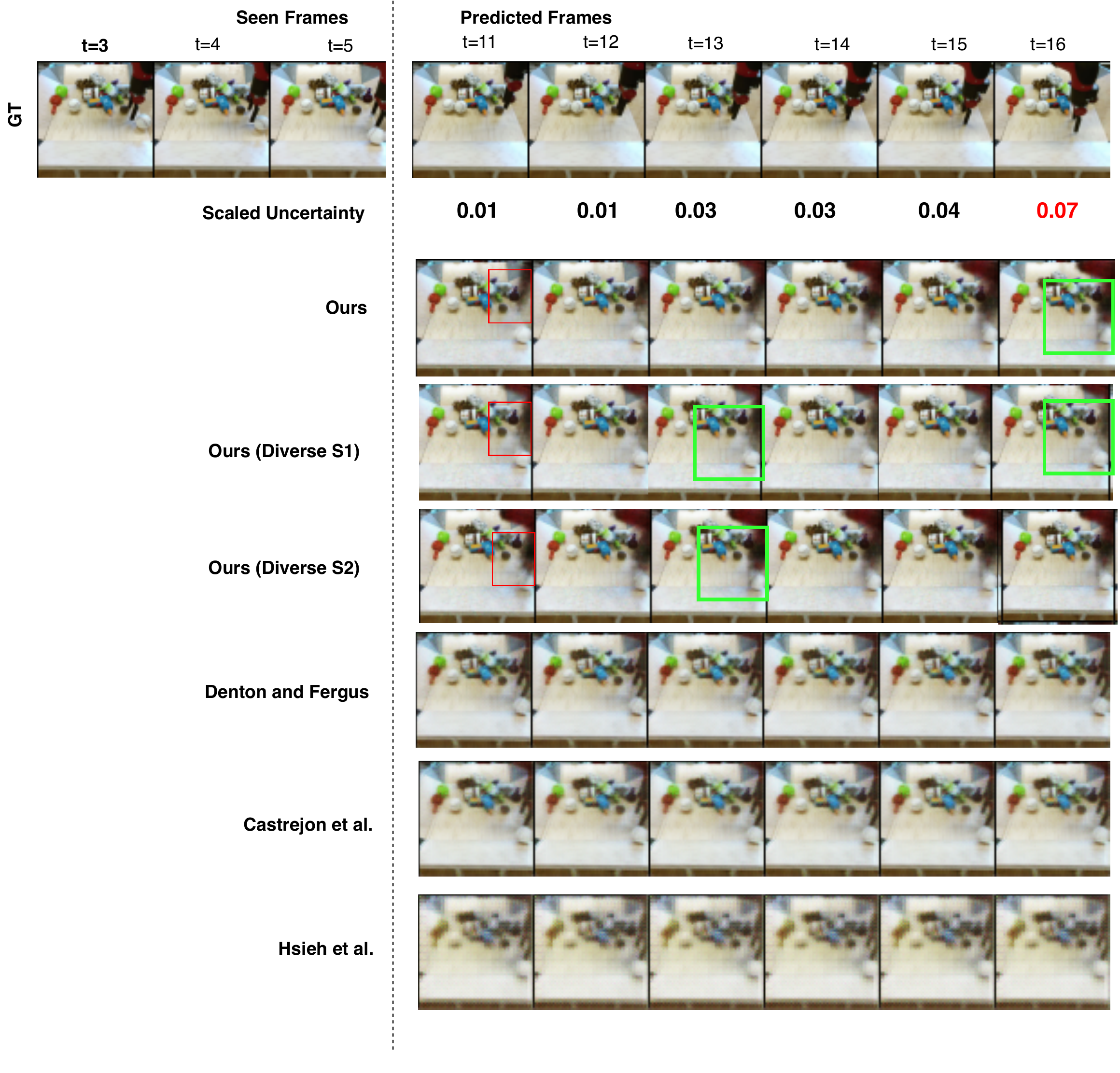} % width=6.9cm
   \caption{Visualization of generations by our method versus competing baselines on the BAIR Robot Push Dataset, trained with 2,000 training samples.  Further, diverse generations by our method are also shown. High motion regions are indicated by a red bounding box, while spatial regions exhibiting high diversity are shown by a green bounding box. Note scaled uncertainty higher than 0.05 is shown in red.}
    \label{fig:bair_comp_9}
    %\vspace*{-0.5cm}
\end{figure*}

\begin{figure*}[th]
    \centering
    \includegraphics[scale=0.75]{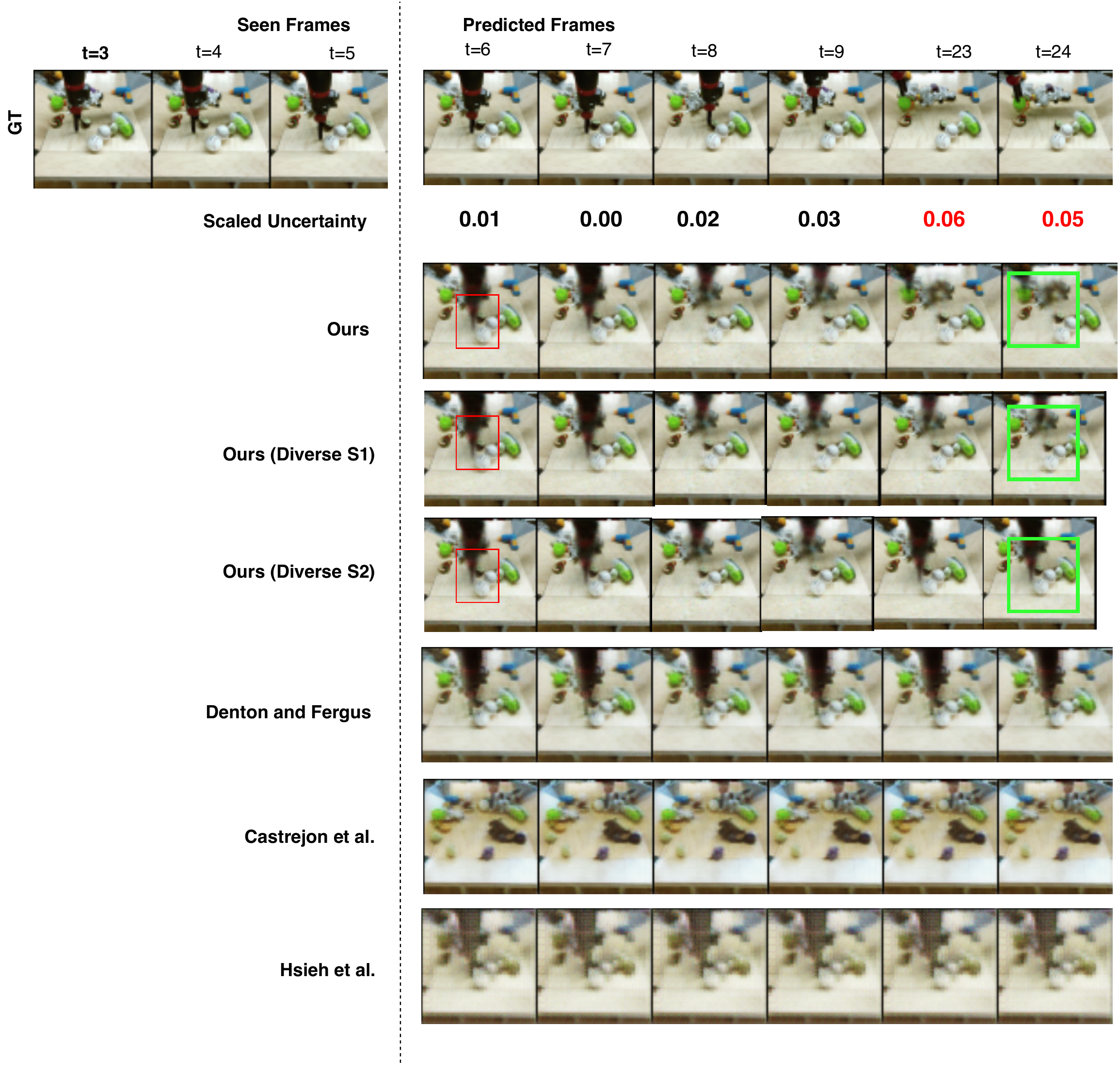} % width=6.9cm
   \caption{Visualization of generations by our method versus competing baselines on the BAIR Robot Push Dataset, trained with 2,000 training samples.  Further, diverse generations by our method are also shown. High motion regions are indicated by a red bounding box, while spatial regions exhibiting high diversity are shown by a green bounding box. Note scaled uncertainty higher than 0.05 is shown in red.}
    \label{fig:bair_comp_176}
    %\vspace*{-0.5cm}
\end{figure*}

\begin{figure*}[th]
    \centering
    \includegraphics[scale=0.75]{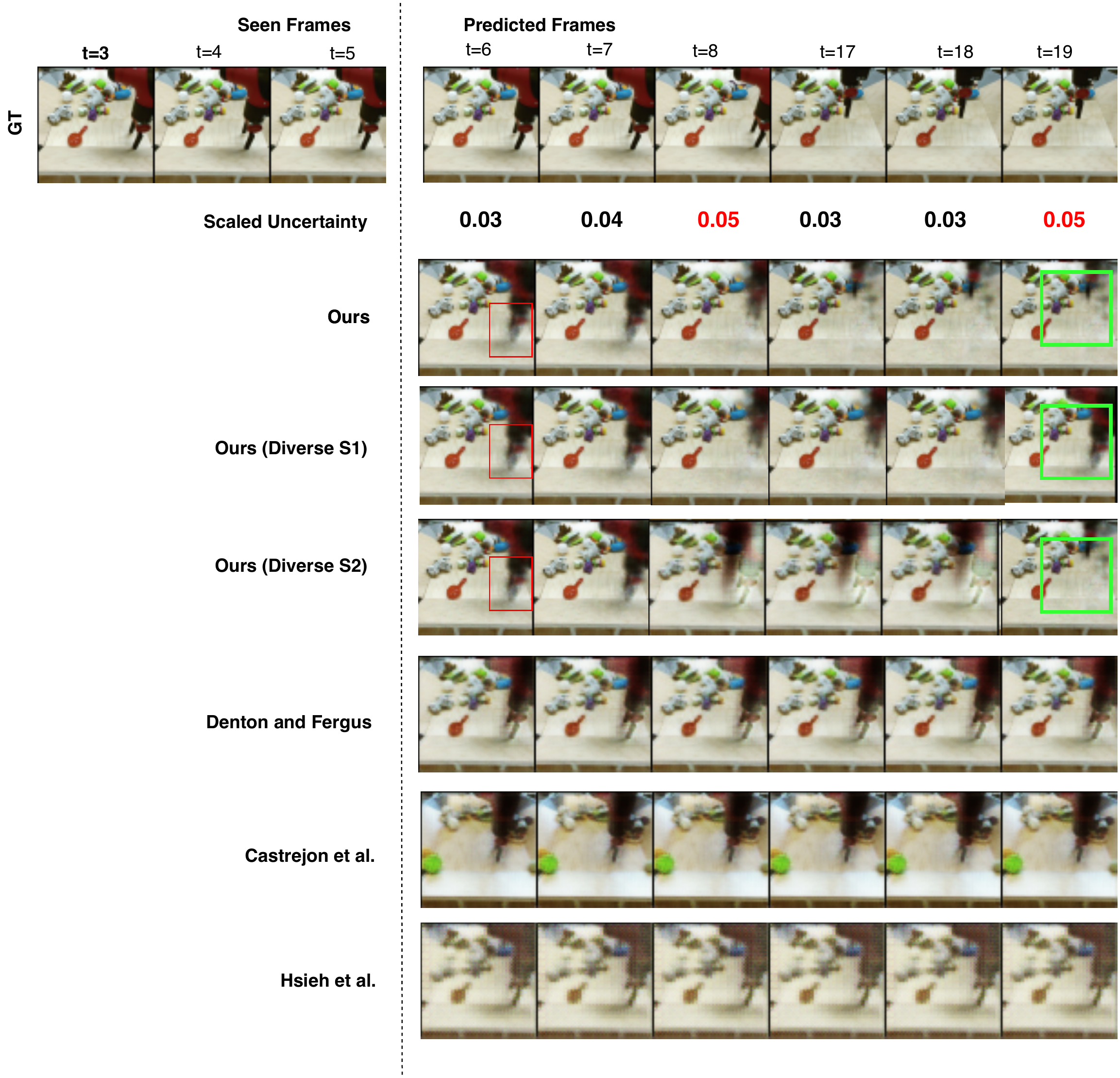} % width=6.9cm
   \caption{Visualization of generations by our method versus competing baselines on the BAIR Robot Push Dataset, trained with 2,000 training samples.  Further, diverse generations by our method are also shown. High motion regions are indicated by a red bounding box, while spatial regions exhibiting high diversity are shown by a green bounding box. Note scaled uncertainty higher than 0.05 is shown in red.}
    \label{fig:bair_comp_191}
    %\vspace*{-0.5cm}
\end{figure*}

\begin{figure*}[th]
    \centering
    \includegraphics[scale=0.75]{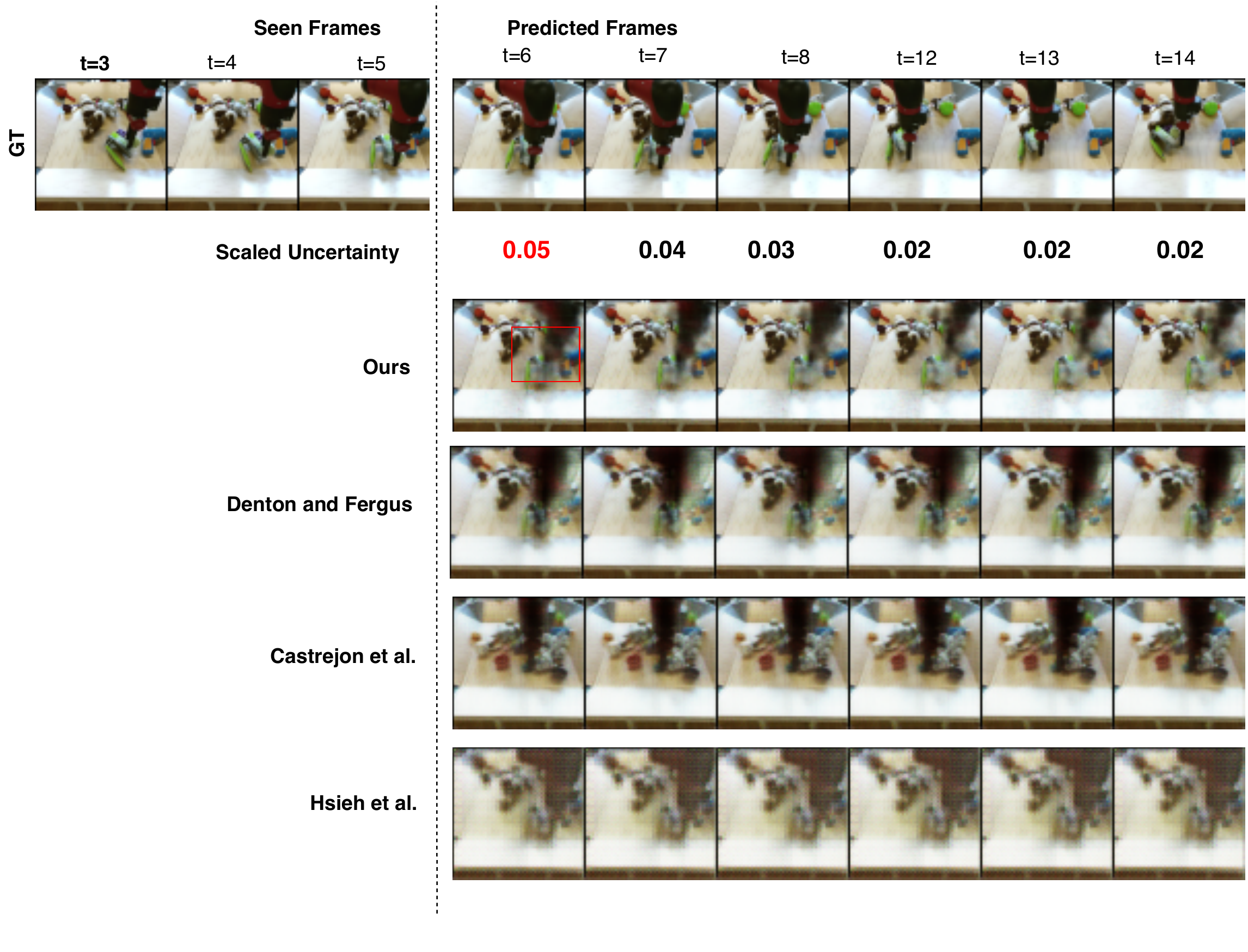} % width=6.9cm
  \caption{Visualization of generations by our method versus competing baselines on the BAIR Robot Push Dataset, trained with 2,000 training samples.  High motion regions are indicated by a red bounding box. Note scaled uncertainty higher than 0.05 is shown in red.}
    \label{fig:bair_comp_161}
    %\vspace*{-0.5cm}
\end{figure*}

\begin{figure*}[th]
    \centering
    \includegraphics[scale=0.75]{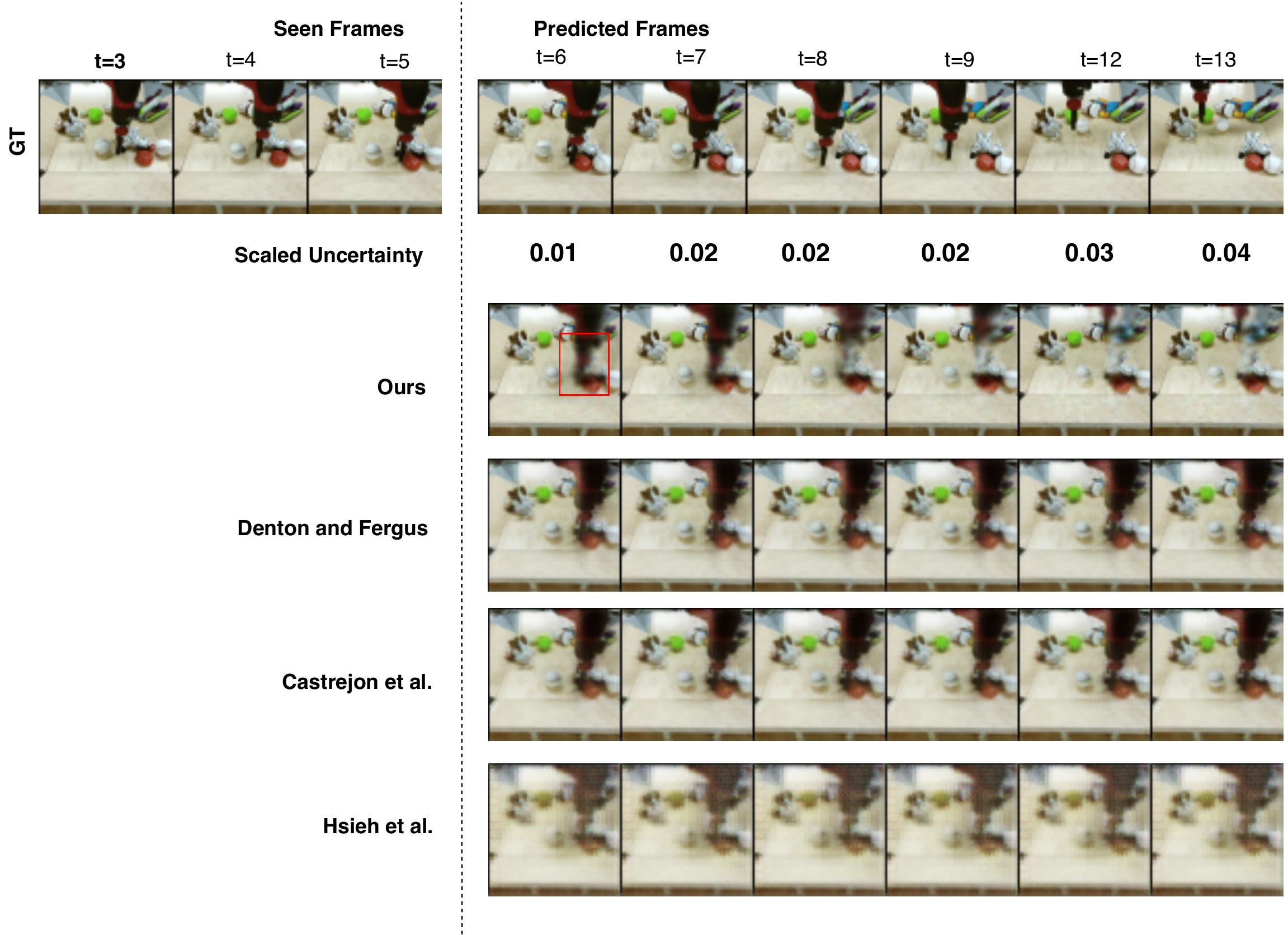} % width=6.9cm
   \caption{Visualization of generations by our method versus competing baselines on the BAIR Robot Push Dataset, trained with 2,000 training samples.  High motion regions are indicated by a red bounding box. Note scaled uncertainty higher than 0.05 is shown in red.}
    \label{fig:bair_comp_18}
    %\vspace*{-0.5cm}
\end{figure*}

\begin{figure*}[th]
    \centering
    \includegraphics[scale=0.75]{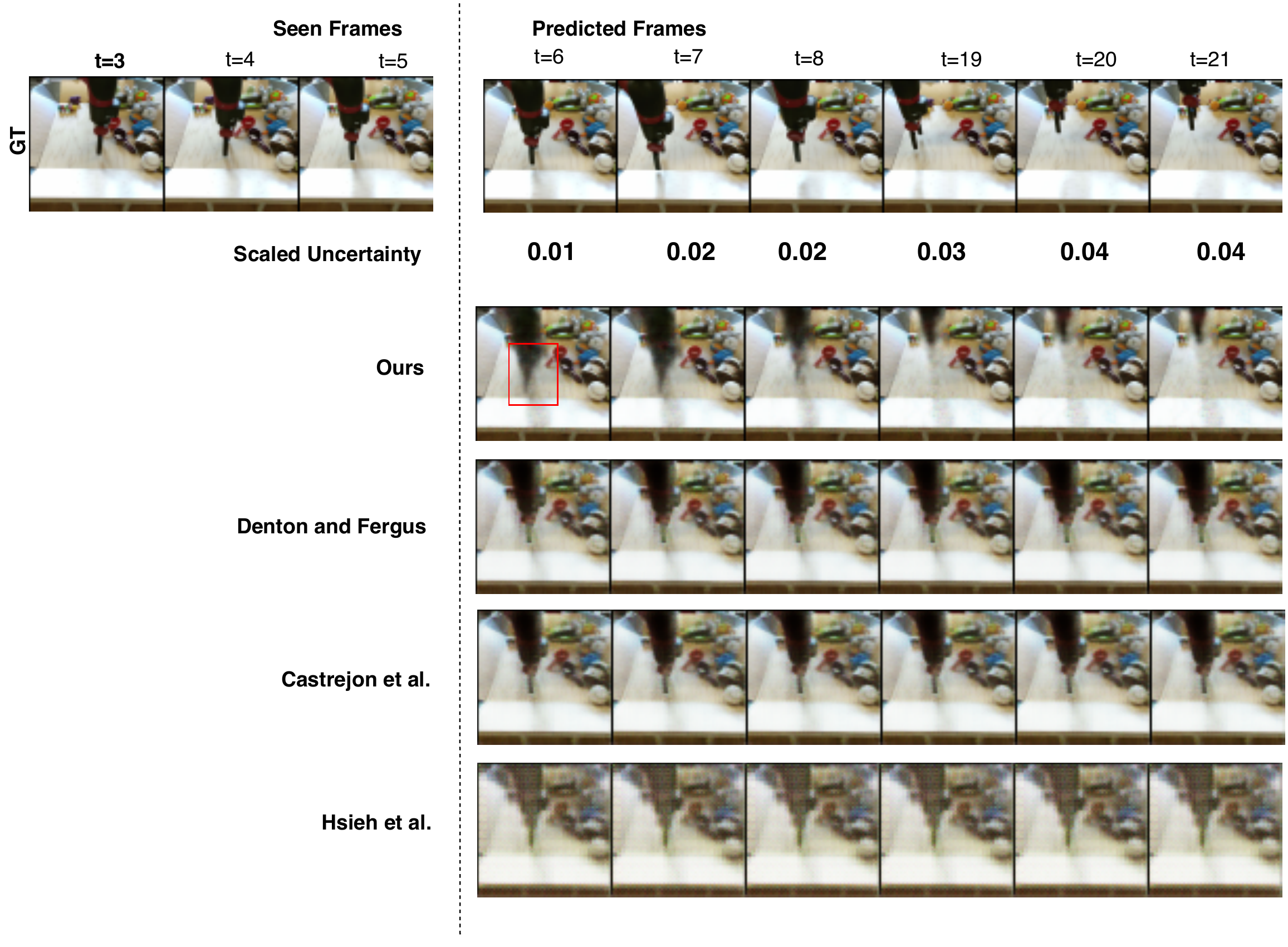} % width=6.9cm
  \caption{Visualization of generations by our method versus competing baselines on the BAIR Robot Push Dataset, trained with 2,000 training samples.  High motion regions are indicated by a red bounding box. Note scaled uncertainty higher than 0.05 is shown in red.}
    \label{fig:bair_comp_29}
    %\vspace*{-0.5cm}
\end{figure*}

\begin{figure*}[th]
    \centering
    \includegraphics[scale=0.75]{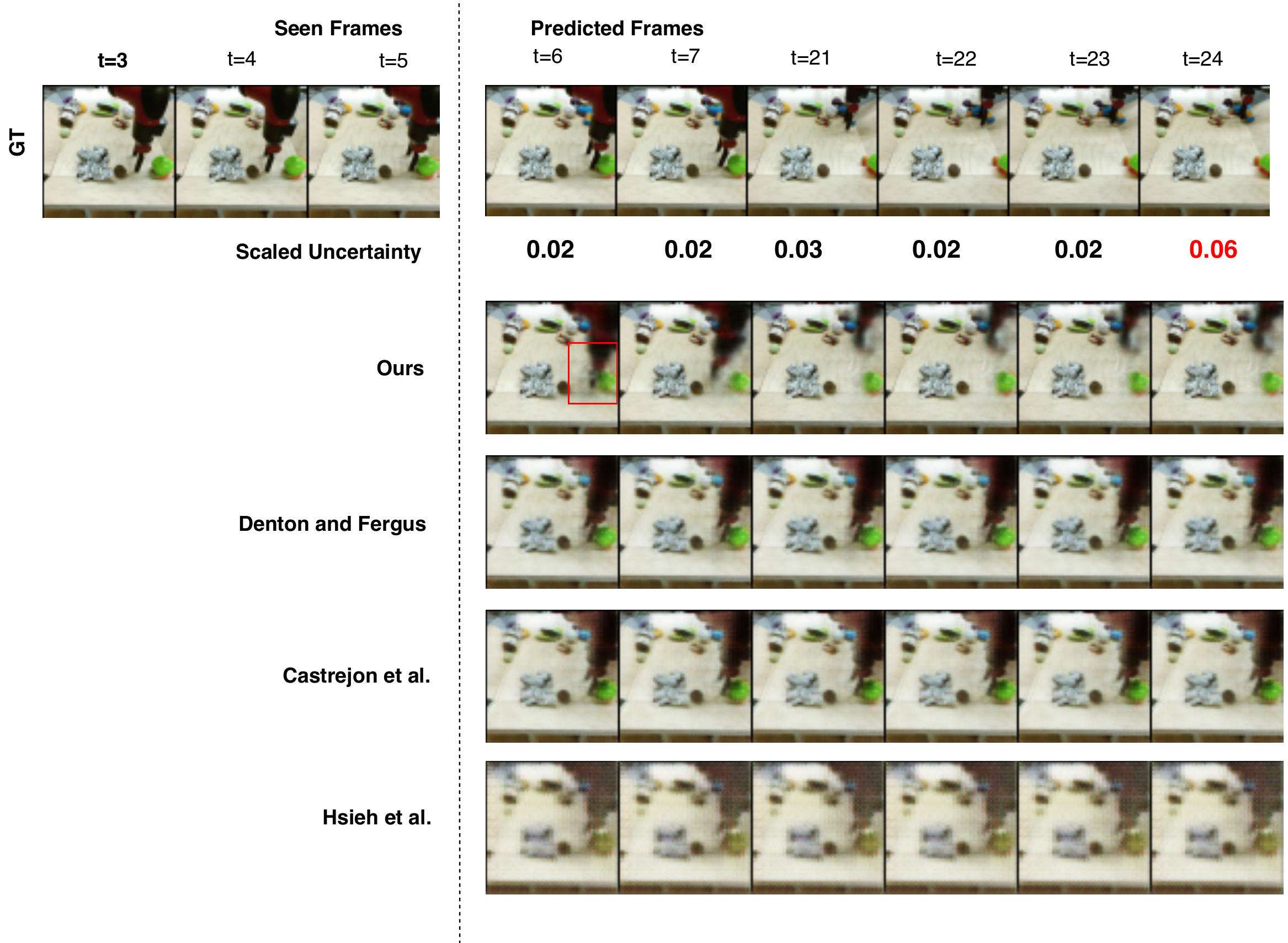} % width=6.9cm
   \caption{Visualization of generations by our method versus competing baselines on the BAIR Robot Push Dataset, trained with 2,000 training samples.  High motion regions are indicated by a red bounding box. Note scaled uncertainty higher than 0.05 is shown in red.}
    \label{fig:bair_comp_60}
    %\vspace*{-0.5cm}
\end{figure*}

\begin{figure*}[th]
    \centering
    \includegraphics[scale=0.75]{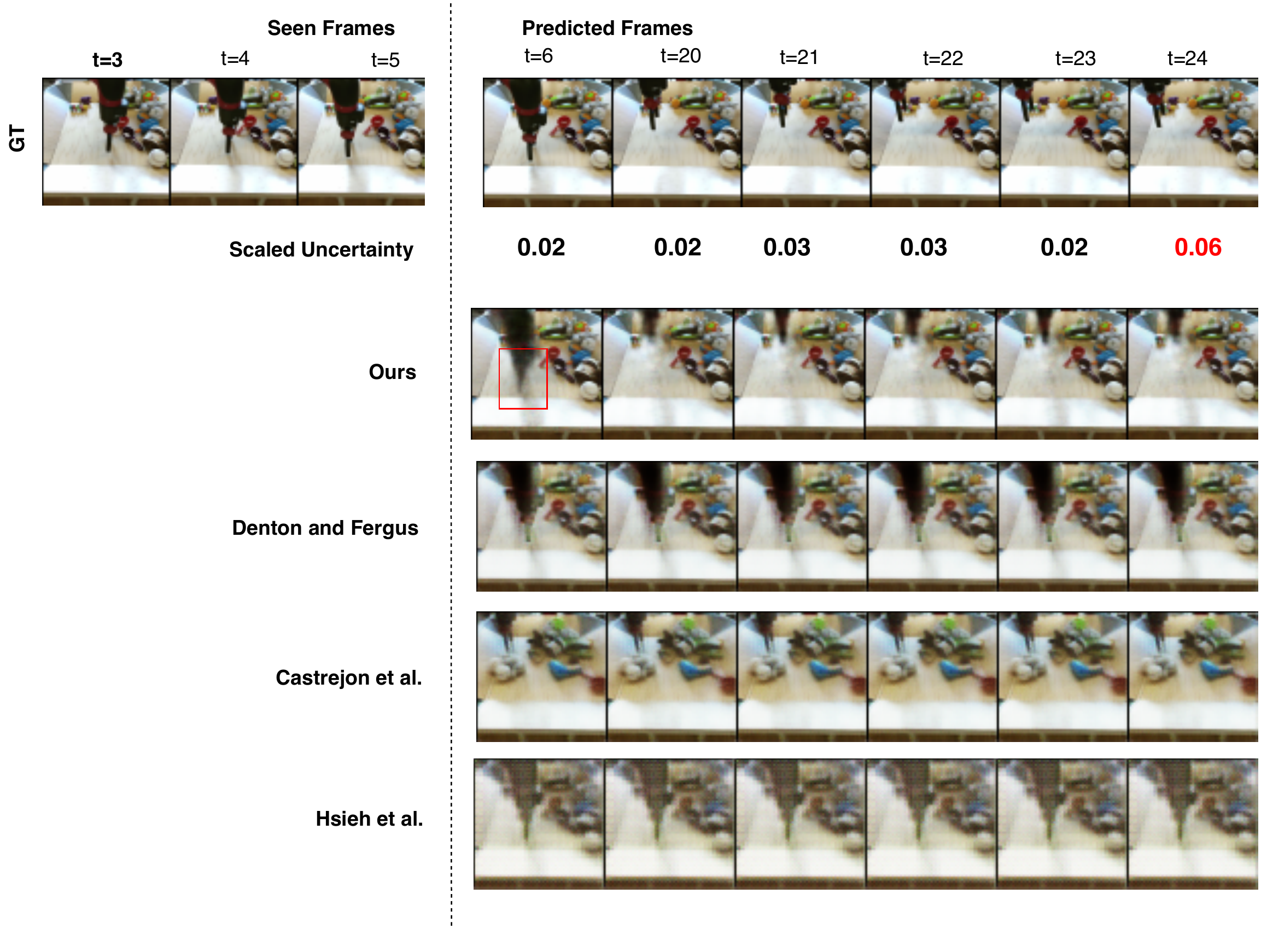} % width=6.9cm
   \caption{Visualization of generations by our method versus competing baselines on the BAIR Robot Push Dataset, trained with 2,000 training samples.  High motion regions are indicated by a red bounding box. Note scaled uncertainty higher than 0.05 is shown in red.}
    \label{fig:bair_comp_157}
    %\vspace*{-0.5cm}
\end{figure*}

We also see in Figures~\ref{fig:bair_comp_188},~\ref{fig:bair_comp_48},~\ref{fig:bair_comp_161},~\ref{fig:bair_comp_9},~\ref{fig:bair_comp_18},~\ref{fig:bair_comp_29},~\ref{fig:bair_comp_60},~\ref{fig:bair_comp_157},~\ref{fig:bair_comp_176},~\ref{fig:bair_comp_191} the performance of different competing methods versus NUQ on the BAIR push dataset, trained with 2000 samples. The figures reveal that our method captures the motion of the robot arm, reasonably well, compared to competing methods.

\begin{figure*}[th]
    \centering
    \includegraphics[scale=0.8]{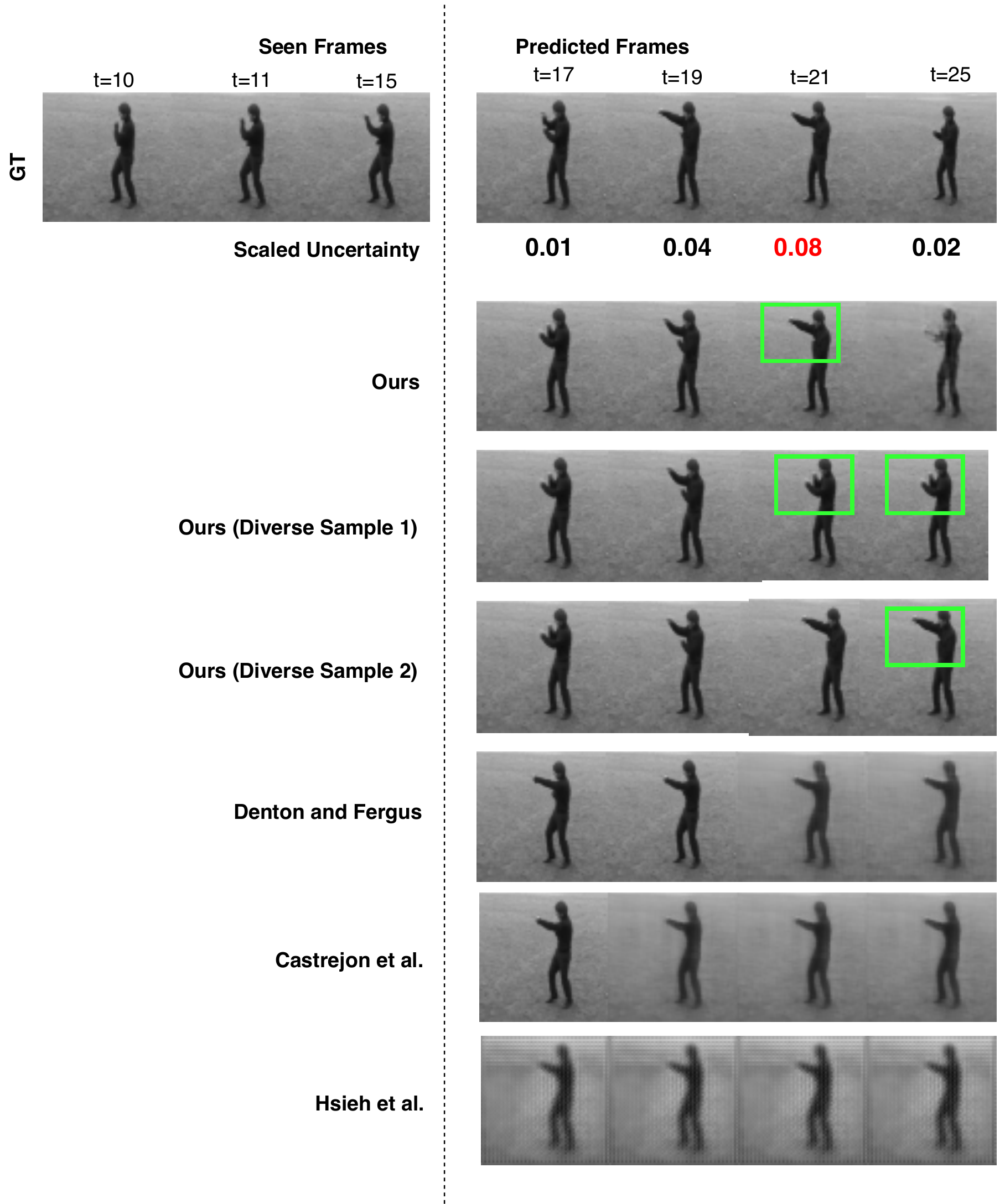} % width=6.9cm
   \caption{Visualization of generations by our method versus competing baselines on the KTH Action Dataset, trained with the full training data of 1,911 training samples.  Further, diverse generations by our method are also shown. Spatial regions exhibiting high diversity are shown by a green bounding box. Note scaled uncertainty higher than 0.05 is shown in red.}
    \label{fig:kth_comp_4}
    %\vspace*{-0.5cm}
\end{figure*}

\begin{figure*}[th]
    \centering
    \includegraphics[scale=0.8]{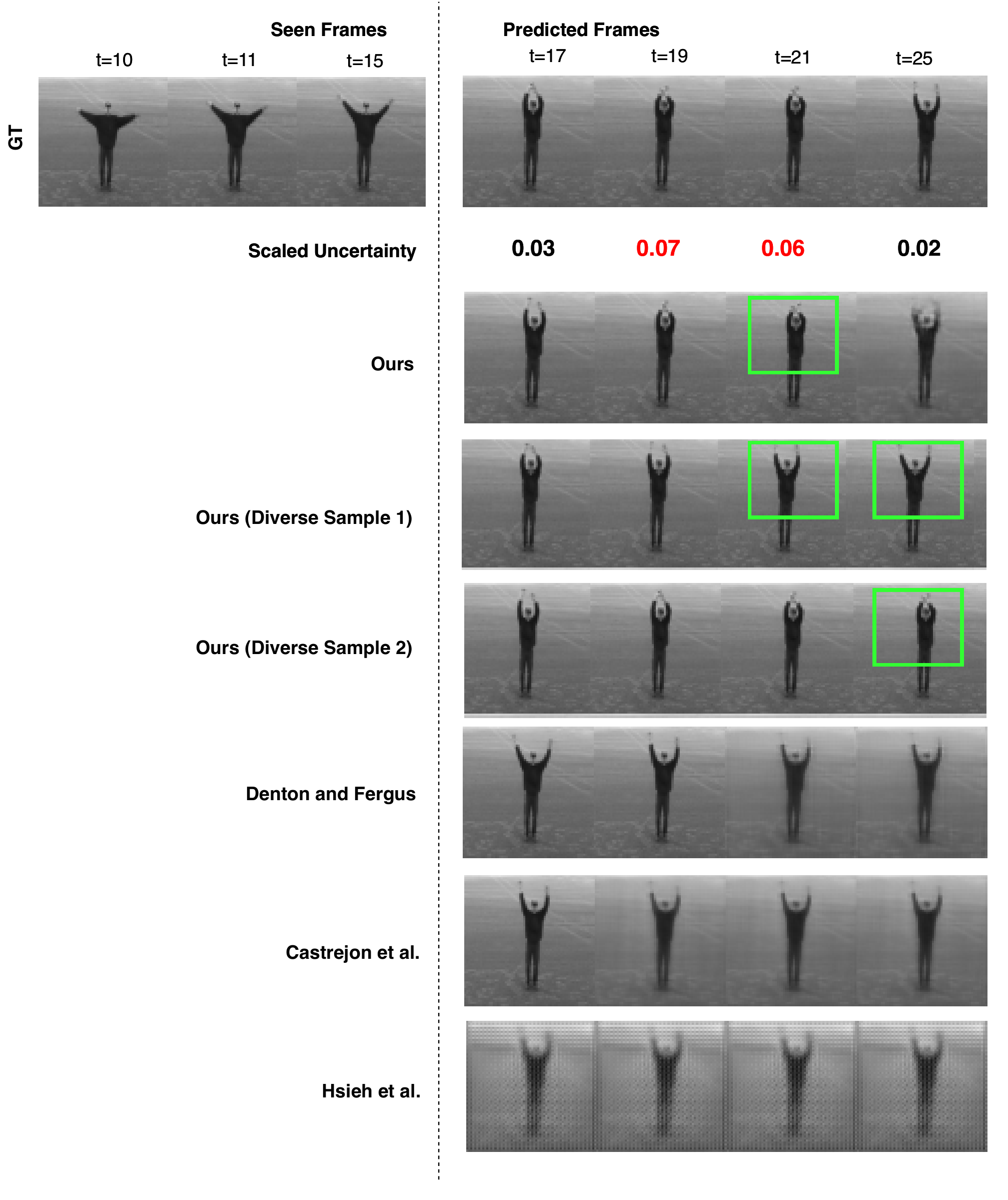} % width=6.9cm
   \caption{Visualization of generations by our method versus competing baselines on the KTH Action Dataset, trained with the full training data of 1,911 training samples.  Further, diverse generations by our method are also shown. Spatial regions exhibiting high diversity are shown by a green bounding box. Note scaled uncertainty higher than 0.05 is shown in red.}
    \label{fig:kth_comp_223}
    %\vspace*{-0.5cm}
\end{figure*}

Figures~\ref{fig:kth_comp_99},~\ref{fig:kth_comp_4},~\ref{fig:kth_comp_223} present sample generation results by our method versus competing baselines on the KTH Action dataset. From the figures, we see that while all of the methods do a reasonable job of modeling the appearance of the person, nonetheless the competing methods fail to capture the motion dynamics well.

\begin{figure*}[th]
    \centering
    \includegraphics[scale=0.8]{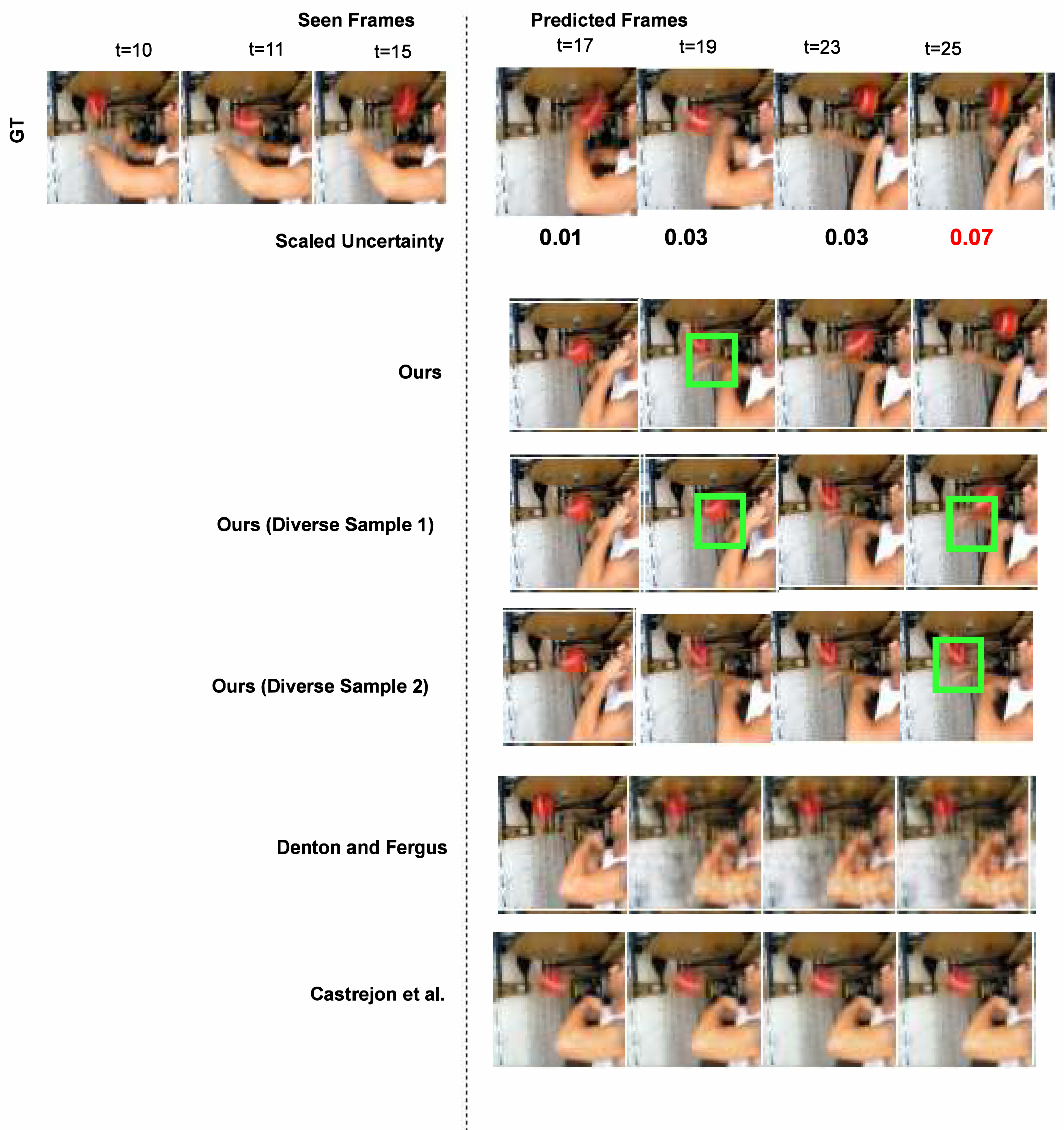} % width=6.9cm
   \caption{Visualization of generations by our method versus competing baselines on the UCF-101 Dataset, trained with the full training data of 11,425 training samples.  Further, diverse generations by our method are also shown. Spatial regions exhibiting high diversity are shown by a green bounding box. Note scaled uncertainty higher than 0.05 is shown in red.}
    \label{fig:ucf_comp_2}
    %\vspace*{-0.5cm}
\end{figure*}

\begin{figure*}[th]
    \centering
    \includegraphics[scale=0.8]{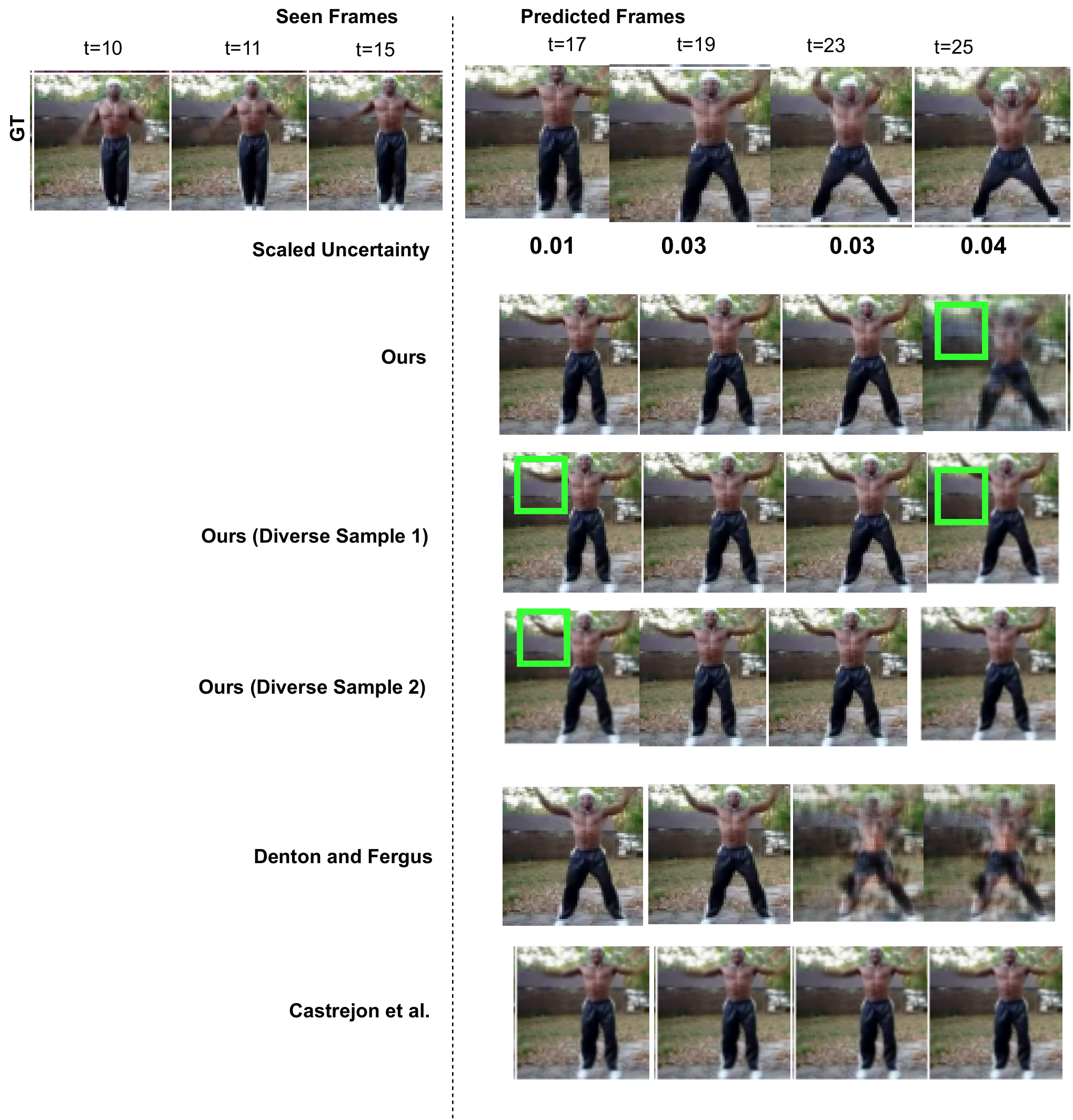} % width=6.9cm
   \caption{Visualization of generations by our method versus competing baselines on the UCF-101 Dataset, trained with the full training data of 11,425 training samples.  Further, diverse generations by our method are also shown. Spatial regions exhibiting high diversity are shown by a green bounding box. Note scaled uncertainty higher than 0.05 is shown in red.}
    \label{fig:ucf_comp_3}
\end{figure*}

Such trends extend into the UCF-101 dataset as well. The results for this dataset are shown in  Figures~\ref{fig:ucf_comp_2},~\ref{fig:ucf_comp_3}.

Moreover, in some of the aforementioned figures (such as Figures~\ref{fig:smmnist_comp_157},~\ref{fig:smmnist_comp_180},~\ref{fig:smmnist_comp_295},~\ref{fig:bair_comp_48},~\ref{fig:bair_comp_188},~\ref{fig:bair_comp_191},~\ref{fig:kth_comp_4},~\ref{fig:kth_comp_223},~\ref{fig:ucf_comp_2},~\ref{fig:ucf_comp_3} diverse sample generations by NUQ is also shown. 
 
% \begin{figure*}[th]
%     %\centering
%     %\hspace{-2.5cm}
%     \includegraphics[scale=0.6]{figs/suppl_cvpr_figs/838_Comp_SMMNIST.pdf} % width=6.9cm
%   \caption{Visualization of generations by our method versus competing baselines on the SMMNIST Dataset, trained with 2,000 training samples. Note scaled uncertainty higher than 0.05 is shown in red.}
%     \label{fig:smmnist_comp_838}
%     %\vspace*{-0.5cm}
% \end{figure*}

\end{document}

%% file: math_commands.tex
\usepackage{amsmath,amsfonts,bm}

\def\eqref#1{Eq.~\ref{#1}}
\def\1{\bm{1}}

\def\vmu{{\bm{\mu}}}

\def\vh{{\bm{h}}}

\def\vw{{\bm{w}}}
\def\vx{{\bm{x}}}
\def\vy{{\bm{y}}}
\def\vz{{\bm{z}}}

\def\mSigma{{\bm{\Sigma}}}

\DeclareMathAlphabet{\mathsfit}{\encodingdefault}{\sfdefault}{m}{sl}
\SetMathAlphabet{\mathsfit}{bold}{\encodingdefault}{\sfdefault}{bx}{n}

%% file: definitions.tex
\newcommand{\s}{s}

\newcommand{\ts}{\tilde{\s}}

\newcommand{\disc}{D}

\newcommand{\loss}{\mathcal{L}}

\newcommand{\dataset}{\mathcal{X}}

\newcommand{\prob}[1]{p_{#1}}
\newcommand{\p}{p}
\newcommand{\q}{q}
\newcommand{\f}{f}
\newcommand{\g}{g}

\newcommand{\expect}[1]{\mathbb{E}_{#1}}

\newcommand{\reals}[1]{\mathbb{R}^{#1}}

\newcommand{\enorm}[1]{\left\|{#1}\right\|}

\newcommand{\seq}[1]{\left\langle{#1}\right\rangle}

\newcommand{\normal}{\mathcal{N}}

\DeclareMathOperator{\kl}{KL}

\DeclareMathOperator{\uniform}{Uniform}